\documentclass[journal,10pt,english,final]{IEEEtran}
\IEEEoverridecommandlockouts

\usepackage[T1]{fontenc} % optional
\usepackage[utf8]{inputenc}
\usepackage{amsmath}

\usepackage{bm} % optional

\usepackage{babel}
\usepackage{graphicx}
\usepackage{array}
\usepackage{pbox}
\usepackage{multirow}

\usepackage{url}

\usepackage{amsfonts}
\usepackage{amssymb}
\usepackage{amsthm}
\usepackage{balance}
\usepackage{placeins}

\usepackage{cite}

\usepackage{abstract}
\usepackage{subcaption}
\usepackage{gensymb}

\DeclareMathOperator*{\argmin}{arg\,min}
\theoremstyle{definition}
\newtheorem{problem}{Problem}[section]

\numberwithin{equation}{section}
\numberwithin{figure}{section}

\usepackage[ruled,vlined]{algorithm2e}	
\renewcommand{\vec}[1]{\mathbf{#1}}

\providecommand{\tabularnewline}{\\}
\usepackage{float}
\restylefloat{table}

\usepackage{booktabs}% http://ctan.org/pkg/booktabs
\usepackage{colortbl}% http://ctan.org/pkg/colortbl

\usepackage{xcolor}% http://ctan.org/pkg/xcolor
\definecolor{white}{rgb}{1,1,1}
\definecolor{grey}{rgb}{0.9,0.9,0.9}
\definecolor{dark_grey}{rgb}{0.8,0.8,0.8}

\usepackage{wrapfig}

\colorlet{tableheadcolor}{gray!25} % Table header colour = 25% gray
\colorlet{tablerowcolor}{gray!10} % Table row separator colour = 10% gray
\usepackage{etoolbox}
\usepackage{environ}
\usepackage{ifthen}
\newboolean{showshort}
\newboolean{showfull}

\setboolean{showshort}{false}
\setboolean{showfull}{true}
\ifthenelse{\boolean{showshort}}%
{%
  \NewEnviron{short}
    {%
     \color{black}
     \BODY{}
     \color{black}
    }%
   \newcommand{\shortc}[1]{\textcolor{black}{#1}}  
}%
    {\NewEnviron{short}
    {%
    }% 
   \newcommand{\shortc}[1]{}   
}%

\ifthenelse{\boolean{showfull}}%
{%
  \NewEnviron{full}
    {%
     \color{black}
     \BODY{}
     \color{black}
    }%
    \newcommand{\fullc}[1]{\textcolor{black}{#1}}
}%
    {\NewEnviron{full}
    {%
    }%
    \newcommand{\fullc}[1]{}
}%

\begin{document}

\title{A Survey of Motion Planning and Control Techniques for Self-driving Urban Vehicles}

\author{Brian Paden$^{\ast,1}$, Michal Čáp$^{\ast,1,2}$, Sze Zheng Yong$^{1}$, Dmitry Yershov$^{1}$, and Emilio Frazzoli$^{1}$
\thanks{$^{\ast}$ The first two authors contributed equally to this work.}
\thanks{$^{1}$ The authors are with the Laboratory for Information and Decision Systems,
        Massachusetts Institute of Technology, Cambridge MA, USA.
        {\tt\small email: bapaden@mit.edu, mcap@mit.edu, szyong@mit.edu, yershov@mit.edu, frazzoli@mit.edu}}
\thanks{$^{2}$ Michal Čáp is also affiliated with Dept. of Computer Science, Faculty of Electrical Engineering, CTU in Prague, Czech Republic.}
}

%\maketitle

%\onecolumn{
%%\begin{@twocolumnfalse}
%    \tableofcontents}
%    \end{@twocolumnfalse}
%     \twocolumn  
\begin{full}
 \twocolumn[
\begin{@twocolumnfalse}
\maketitle \vspace{-0.3cm}

\begin{onecolabstract}
Self-driving vehicles are a maturing technology with the potential to reshape mobility by enhancing the safety, accessibility, efficiency, and convenience of automotive transportation. 
Safety-critical tasks that must be executed by a self-driving vehicle include planning of motions through a dynamic environment shared with other vehicles and pedestrians, and their robust executions via feedback control. 
The objective of this paper is to survey the current state of the art on planning and control algorithms with particular regard to the urban setting. 
A selection of proposed techniques is reviewed along with a discussion of their effectiveness.  
The surveyed approaches differ in the vehicle mobility model used, in assumptions on the structure of the environment, and in computational requirements. 
The side-by-side comparison presented in this survey helps to gain insight into the strengths and limitations of the reviewed approaches and assists with system level design choices.
\end{onecolabstract}
\hrulefill 
\vspace{-0.3cm}
{\small \tableofcontents}
\vspace{0.1cm}
\hrulefill
\end{@twocolumnfalse}]
\saythanks
\end{full}

\begin{short}
\maketitle
\begin{abstract}
Self-driving vehicles are a maturing technology with the potential to reshape mobility by enhancing the safety, accessibility, efficiency, and convenience of automotive transportation. 
Safety-critical tasks that must be executed by a self-driving vehicle include planning of motions through a dynamic environment shared with other vehicles and pedestrians, and their robust executions via feedback control. 
The objective of this paper is to survey the current state of the art on planning and control algorithms with particular regard to the urban setting. 
A selection of proposed techniques is reviewed along with a discussion of their effectiveness.  
The surveyed approaches differ in the vehicle mobility model used, in assumptions on the structure of the environment, and in computational requirements. 
The side by side comparison presented in this survey helps to gain insight into the strengths and limitations of the reviewed approaches and assists with system level design choices.
\end{abstract}
\end{short}

\section{Introduction}

The last three decades have seen steadily increasing research efforts, both in academia and in industry, towards developing driverless vehicle technology. 
These developments have been fueled by recent advances in sensing and computing technology together with the potential transformative impact on automotive transportation and the perceived societal benefit: 
%
%These developments have been fueled by advances in sensing and computing technology, and by the high expectations over the potential transformative impact on automotive transportation and resulting benefits to society such as drastically decreased traffic fatalities, improved quality of life for disabled, and large gains in economic productivity:
%
In 2014 there were 32,675 traffic related fatalities, 2.3 million injuries, and 6.1 million reported collisions~\cite{nhtsa2014}. 
Of these, an estimated 94\% are attributed to driver error  with 31\% involving legally intoxicated drivers, and 10\% from distracted drivers~\cite{nhtsa2015}. 
Autonomous vehicles have the potential to dramatically reduce the contribution of driver error and negligence as the cause of vehicle collisions.
They will also provide a means of personal mobility to people who are %the portion of the population 
unable to drive due to physical or visual disability.
Finally, for the 86\% of the US work force that commutes by car, on average 25 minutes (one way) each day~\cite{mckenzie2011}, autonomous vehicles would facilitate more productive use of the transit time, or simply reduce the measurable ill effects of driving stress~\cite{hennessy1999}.

% History; major events and milestones 
Considering the potential impacts of this new technology, it is not surprising that %the vision of 
self-driving cars have had a long history.
The idea has been around as early as in %since at least 
the 1920s, but it was not until the 1980s that driverless cars seemed like a real possibility. 
Pioneering work led by Ernst Dickmanns (e.g.,~\cite{dickmanns1988dynamic}) in the 1980s paved the way for the development of autonomous vehicles.
At that time a massive research effort, the PROMETHEUS project, was funded to develop an autonomous vehicle. 
A notable demonstration in 1994 resulting from the work was a 1,600 km drive by the VaMP driverless car, of which 95\% was driven autonomously~\cite{dickmanns1997vehicles}. 
At a similar time, the CMU NAVLAB was making advances in the area and in 1995 demonstrated further progress with a 5,000 km drive across the US of which 98\% was driven autonomously~\cite{noHands}.
The next major milestone in driverless vehicle technology was the first DARPA Grand Challenge in 2004. 
The objective was for a driverless car to navigate a 150-mile off-road course as quickly as possible. 
This was a major challenge in comparison to previous demonstrations in that there was to be no human intervention during the race.
Although prior works demonstrated nearly autonomous driving, eliminating human intervention at critical moments proved to be a major challenge. 
None of the 15 vehicles entered into the event completed the race. In 2005 a similar event was held; this time 5 of 23 teams reached the finish line~\cite{buehler20072005}. 
Later, in 2007, the DARPA Urban Challenge was held, in which %. In this event 
vehicles were required to drive autonomously in a simulated urban setting. Six teams finished the event demonstrating that fully autonomous urban driving is possible~\cite{buehler2009darpa}. 
Numerous events and major autonomous vehicle system tests have been carried out since the DARPA challenges. 
Notable examples include the Intelligent Vehicle Future Challenges from 2009 to 2013 \cite{xin2014china}, Hyundai Autonomous Challenge in 2010 \cite{cerri2011computer}, the VisLab Intercontinental Autonomous Challenge in 2010 \cite{broggi2012vislab}, the Public Road Urban Driverless-Car Test  in 2013 \cite{broggi2015proud}, and the autonomous drive of the Bertha-Benz historic route \cite{ziegler2014making}.
Simultaneously, %focused 
research has continued at an accelerated pace in both the academic setting as well as in industry. The Google self-driving car \cite{googleSite} and Tesla's Autopilot system \cite{teslaSite} are two examples of commercial efforts receiving considerable media attention.  
%

%
%Scope of survey
%

The extent to which a car is automated can vary from fully human operated to fully autonomous.
The SAE J3016 standard~\cite{sae2014taxonomy} introduces a scale from 0 to 5 for grading vehicle automation. 
In this standard, the level 0 represents a vehicle where all driving tasks are the responsibility of a human driver.
Level 1 includes basic driving assistance such as adaptive cruise control, anti-lock braking systems and electronic stability control \cite{rajamani2011vehicle}. 
Level 2 includes advanced assistance such as hazard-minimizing longitudinal/lateral control~\cite{gerdes2001unified} or emergency braking~\cite{brannstrom2010model, vahidi2003research}, often based upon set-based formal control theoretic methods to compute `worst-case' sets of provably collision free (safe) states \cite{hafner2013cooperative,colombo2012efficient,kowshik2011provable}.
At level 3 the system monitors the environment and can drive with full autonomy under certain conditions, but the human operator is still required to take control if the driving task leaves the autonomous system's operational envelope.
A vehicle with level 4 automation is capable of fully autonomous driving in certain conditions and will safely control the vehicle if the operator fails to take control upon request to intervene.  
Level 5 systems are fully autonomous in all driving modes. 

The availability of on-board computation and wireless communication technology allows cars to exchange information with other cars and with the road infrastructure giving rise to a closely related area of research on connected intelligent vehicles~\cite{eskandarian2012handbook}. 
This research aims to improve the safety and performance of road transport through information sharing and coordination between individual vehicles.
For instance, connected vehicle technology has a potential to improve throughput at intersections~\cite{miculescu2014polling} or prevent formation of traffic shock waves~\cite{zhou2015parsimonious}.

%An emerging area of research which will contribute to vehicle automation at several levels is multi-vehicle coordination and connected vehicle systems \cite{yang2004vehicle, biswas2006vehicle, miculescu2014polling, kamal2015vehicle}. 
%
%A detailed text covering many aspects of intelligent vehicles systems can be found in \cite{eskandarian2012handbook}.
%
To limit the scope of this survey, we focus on aspects of decision making, motion planning, and control for self-driving cars, in particular, for systems falling into the  automation level of 3 and above.
For the same reason, the broad field of perception for autonomous driving is omitted and instead the reader is referred to a number of comprehensive surveys and major recent contributions on the subject~\cite{geronimo2009survey,ros2012visual,geiger2011stereoscan,liu2015classification}.  
% 
%There will be minimal discussion of perception, an important aspect of vehicle automation, and only a list of select recent works are referenced:
%
%Object identification and labeling using machine learning techniques is surveyed in \cite{geronimo2009survey},
%
%a survey of simultaneous localization and mapping techniques for driverless vehicles is presented in \cite{ros2012visual}, 
%
%recent advances in efficient stereo vision for autonomous vehicles are introduced in \cite{geiger2011stereoscan}, 
%
%and vehicle behavioral estimation of nearby vehicles using Hidden Markon Models is addressed in \cite{liu2015classification}.

The decision making in contemporary autonomous driving systems is typically hierarchically structured into route planning, behavioral decision making, local motion planning and feedback control. 
The partitioning of these levels are, however, rather blurred with different variations of this scheme occurring in the literature. 
%
%Many techniques have been proposed for each of these subtasks, each potentially solving a %slightly different formulation of the problem. 
%
This paper provides a survey of proposed methods to address these core problems of autonomous driving. 
Particular emphasis is placed on methods for local motion planning and control. 

\par
%Paper contents
%
The remainder of the paper is structured as follows: 
In Section \ref{section_decision_making}, a high level overview of the hierarchy of decision making processes and some of the methods for their design are presented. 
Section \ref{section_models} reviews models used to approximate the mobility of cars in urban settings for the purposes of motion planning and feedback control.
Section \ref{section_mp} surveys the rich literature on motion planning and discusses its applicability for self-driving cars. 
Similarly,  Section \ref{section_control} discusses the problems of path and trajectory stabilization and specific feedback control methods for driverless cars. 
Lastly,  Section \ref{section_conclusion} concludes with remarks on the state of the art and potential areas for future research.

\FloatBarrier

\section{Overview of the Decision-Making Hierarchy used in Driverless Cars} \label{section_decision_making}

In this section we describe the decision making architecture of a typical self-driving car and comment on the responsibilities of each component. 
Driverless cars are essentially autonomous decision-making systems that process a stream of observations from on-board sensors such as radars, LIDARs, cameras, GPS/INS units, and odometry.
These observations, together with prior knowledge about the road network, rules of the road, vehicle dynamics, and sensor models, are used to automatically select values for controlled variables governing the vehicle's motion.
Intelligent vehicle research aims at automating as much of the driving task as possible.
The commonly adopted approach to this problem is to partition and organize perception and decision-making tasks into a hierarchical structure.
%
%The first division is between perception and decision making. 
%
The prior information and collected observation data are used by the perception system to provide an estimate of the state of the vehicle and its surrounding environment;
the estimates are then used by the decision-making system to control the vehicle so that the driving objectives are accomplished. 

\begin{figure}[t]%[!tp] 
%\begin{minipage}[c]{1\textwidth}
\begin{center}
\includegraphics[width=0.7\columnwidth]{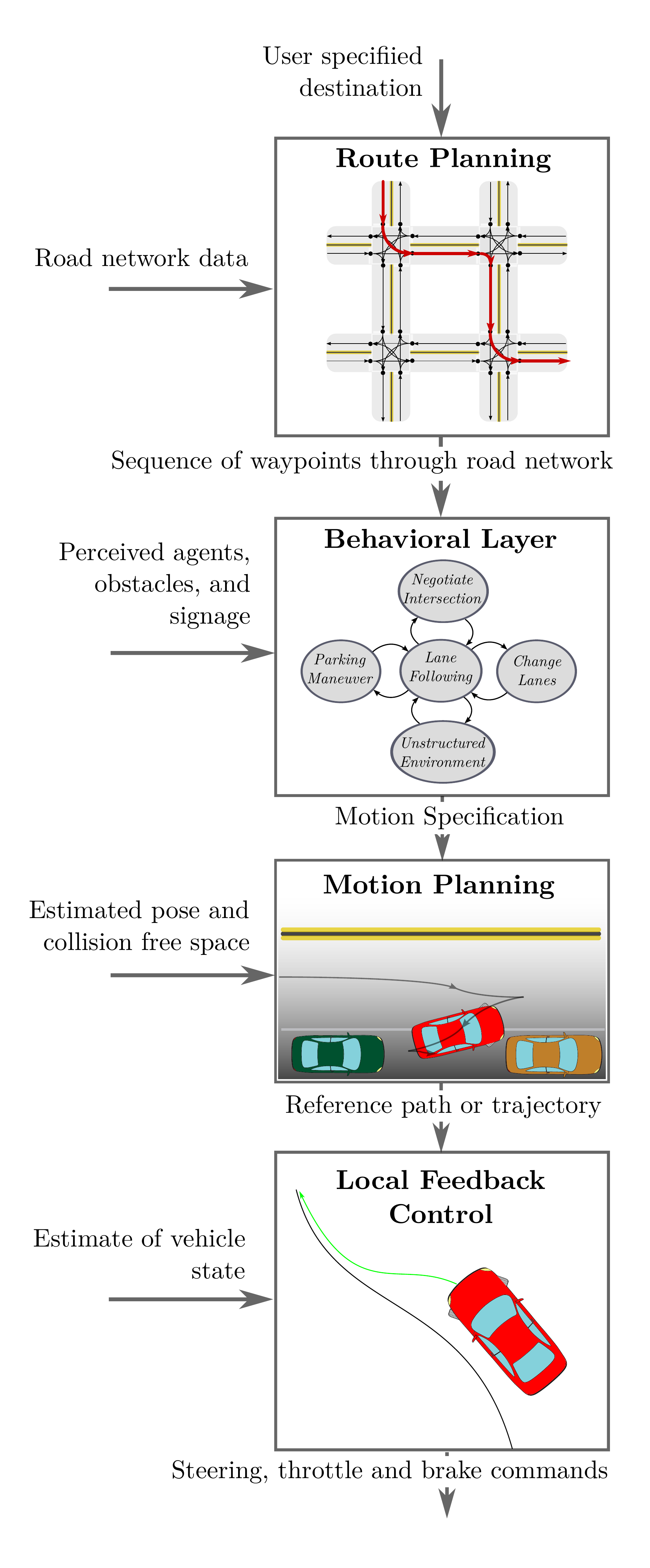}
\caption{Illustration of the hierarchy of decision-making processes. A destination is passed to a route planner that generates a route through the road network. A behavioral layer reasons about the environment and generates a motion specification to progress along the selected route. A motion planner then solves for a feasible motion accomplishing the specification. A feedback control adjusts actuation variables to correct errors in executing the reference path. \label{fig:decision_doodle}}
\end{center}
%\end{minipage} 
\end{figure}

The decision making system of a typical self-driving car is hierarchically decomposed into four components (cf. Figure \ref{fig:decision_doodle}):
At the highest level a route is planned through the road network. 
This is followed by a behavioral layer, which decides on a local driving task that progresses the car towards the destination and abides by rules of the road. 
A motion planning module then selects a continuous path through the environment to accomplish a local navigational task. 
A control system then reactively corrects errors in the execution of the planned motion.
In the remainder of the section we discuss the responsibilities of each of these components in more detail.

\subsection{Route Planning}

At the highest level, a vehicle's decision-making system must select a route through the road network from its current position to the requested destination. 
By representing the road network as a directed graph with edge weights corresponding to the cost of traversing a road segment, such a route can be formulated as the problem of finding a minimum-cost path on a road network graph. 
The graphs representing road networks can however contain millions of edges making classical shortest path algorithms such as Dijkstra~\cite{dijkstra1959note} or A*~\cite{nilsson1969mobile} impractical. 
The problem of efficient route planning in transportation networks has attracted significant interest in the transportation science community leading to the invention of a family of algorithms that after a one-time pre-processing step return an optimal route on a continent-scale network in  milliseconds~\cite{goldberg2005computing,geisberger2012exact}. 
For a comprehensive survey and comparison of practical algorithms that can be used to efficiently plan routes for both human-driven and self-driving vehicles, see~\cite{bast2015route}.

\subsection{Behavioral Decision Making}
After a route plan has been found, the autonomous vehicle must be able to navigate the selected route and interact with other traffic participants according to driving conventions and  rules of the road.
Given a sequence of road segments specifying the selected route, the behavioral layer is responsible for selecting an appropriate driving behavior at any point of time based on the perceived behavior of other traffic participants, road conditions, and signals from infrastructure. For example, when the vehicle is reaching the stop line before an intersection, the behavioral layer will command the vehicle to come to a stop, observe the behavior of other vehicles, bikes, and pedestrians at the intersection, and let the vehicle proceed once it is its turn to go.

Driving manuals dictate qualitative actions for specific driving contexts. 
Since both driving contexts and the behaviors available in each context can be modeled as finite sets, a natural approach to automating this decision making is to model each behavior as a state in a finite state machine with transitions governed by the perceived driving context such as relative position with respect to the planned route and nearby vehicles. 
In fact, finite state machines coupled with different heuristics specific to considered driving scenarios were adopted as a mechanism for behavior control by most teams in the DARPA Urban Challenge~\cite{buehler2009darpa}. 

Real-world driving, especially in an urban setting, is however characterized by uncertainty over the intentions of other traffic participants. The problem of intention prediction and estimation of future trajectories of other vehicles, bikes and pedestrians has also been studied. Among the proposed solution techniques are machine learning based techniques, e.g., Gaussian mixture models~\cite{havlak2014discrete}, Gaussian process regression~\cite{tran2013modelling}, %partially observable Markov decision process (POMDP)~\cite{bandyopadhyay2013intention} 
the learning techniques reportedly used in Google's self-driving system for intention prediction~\cite{madrigal2014}, and model-based approaches for directly estimating intentions from sensor measurements \cite{verma2011semiautonomous,yong2014generalized}.

This uncertainty in the behavior of other traffic participants is commonly considered in the behavioral layer for decision making using probabilistic planning formalisms, such as Markov Decision Processes (MDPs) and their generalizations. 
For example, \cite{brechtel2011probabilistic} formulates the behavioral decision-making
problem in the MDP framework. 
Several works~\cite{ulbrich2013probabilistic,brechtel2014probabilistic,galceran2015multipolicy,bandyopadhyay2013intention}  model unobserved driving scenarios and pedestrian intentions explicitly %considering unobserved intentions of other vehicles 
using a partially-observable Markov decision process (POMDP) framework and propose specific approximate solution strategies. %Another common approach for decision making in the presence of other traffic participants is based on scheduling algorithms that are built upon set-based formal control theoretic methods to compute `worse-case' sets of provably collision free (safe) states \cite{colombo2012efficient,kowshik2011provable}.

%An alternative approach models the decision making as a multi-critera decision-making problem to account for the multiple objectives of urban driving such as maintaining safe distances from obstacles and reaching goal configurations in a reasonable time~\cite{furda2011enabling}.
%%
%Techniques that incorporate behavioral aspects directly into trajectory planning process also exist. 
%%
%For instance,~\cite{reyes2013incremental} encodes the behavioral specification in linear temporal logic to constrain the motion planning algorithm. 

\subsection{Motion Planning}

When the behavioral layer decides on the driving behavior to be performed in the current context, which could be, e.g., cruise-in-lane, change-lane, or turn-right, the selected behavior has to be translated into a path or trajectory that can be tracked by the low-level feedback controller. 
The resulting path or trajectory must be dynamically feasible for the vehicle, comfortable for the passenger, and avoid collisions with obstacles detected by the on-board sensors. 
The task of finding such a path or trajectory is a responsibility of the motion planning system\fullc{.}\shortc{, which is discussed in greater detail in Section \ref{section_mp}.}
 
\begin{full} 
The task of motion planning for an autonomous vehicle corresponds to solving the standard motion planning problem as discussed in the robotics literature. 
Exact solutions to the motion planning problem are in most cases computationally intractable. 
Thus, numerical approximation methods are typically used in practice. 
Among the most popular numerical approaches are variational methods that pose the problem as non-linear optimization in a function space, graph-search approaches that construct graphical discretization of the vehicle's state space and search for a shortest path using graph search methods, and incremental tree-based approaches that incrementally construct a tree of reachable states from the initial state of the vehicle and then select the  best branch of such a tree. 
The motion planning methods relevant for autonomous driving are discussed in greater detail in Section~\ref{section_mp}.
\end{full}

\subsection{Vehicle Control}

In order to execute the reference path or trajectory from the motion planning system a feedback controller is used to select appropriate actuator inputs to carry out the planned motion and correct tracking errors.
The tracking errors generated during the execution of a planned motion are due in part to the inaccuracies of the vehicle model.
Thus, a great deal of emphasis is placed on the robustness and stability of the closed loop system. 
%
%A major focus in the automatic controls literature is on obtaining control laws that assign the actuator commands according to simple static functions of the state of the system or a measured output so that the closed loop system achieves some form of stability. 
%
%Thus, the computational load of running the feedback control process is typically minimal.
%
%The exception is model-predictive control which formulates the control problem in the optimization framework and uses mathematical programming techniques to determine the control signal at each time step~\cite{allgower2012nonlinear,raffo2009predictive,yoon2009model,falcone2008linear,falcone2007predictive,kim2014model}. 

Many effective feedback controllers have been proposed for executing the reference motions provided by the motion planning system. 
%
%These are usually based on a kinematic model of the vehicle and involve choosing a steering angle or heading rate so that the vehicle tends towards a reference path. 
%
A survey of related techniques are discussed in detail in Section \ref{section_control}.

\FloatBarrier

\section{Modeling for Planning and Control } \label{section_models}
In this section we will survey the most commonly used models of mobility of car-like vehicles.  Such models are widely used in control and motion planning algorithms to approximate a vehicle's behavior in response to control actions in relevant operating conditions.  
A high-fidelity model may accurately reflect the response of the vehicle, but the added detail may complicate the planning and control problems.
This presents a trade-off between the accuracy of the selected model and the difficulty of the decision problems. 
This section provides an overview of general modeling concepts and a survey of models used for motion planning and control.

Modeling begins with the notion of the vehicle \textit{configuration}, representing its pose or position in the world.   
For example, configuration can be expressed as the planar coordinate of a point on the car together with the car's heading. 
This is a \textit{coordinate system} for the configuration space of the car. 
This coordinate system describes planar rigid-body motions (represented by the Special Euclidean group in two dimensions, $\mathrm{SE}(2)$) and is a commonly used configuration space \cite{lavalle2006planning,de1998feedback,murray1993nonholonomic}. 
Vehicle motion must then be planned and regulated to accomplish driving tasks and while respecting the constraints introduced by the selected model.

\subsection{The Kinematic Single-Track Model}
In the most basic model of practical use, the car consists of two wheels connected by a rigid link and is restricted to move in a plane \cite{fraichard1998path,egerstedt1997path,coulter1992implementation,de1998feedback,murray1993nonholonomic}. 
It is assumed that the wheels do not slip at their contact point with the ground, but can rotate freely about their axes of rotation. 
The front wheel has an added degree of freedom where it is allowed to rotate about an axis normal to the plane of motion. This is to model steering.
These two modeling features reflect the experience most passengers have where the car is unable to make lateral displacement without simultaneously moving forward. 
More formally, the limitation on maneuverability is referred to as a \textit{nonholonomic} constraint \cite{goldstein1965classical,lavalle2006planning}.
The nonholonomic constraint is expressed as a differential constraint on the motion of the car. 
This expression varies depending on the choice of coordinate system. 
Variations of this model have been referred to as the car-like robot, bicycle model, kinematic model, or single track model. 

\begin{figure}[ht!]
\centering{}\includegraphics[width=8cm]{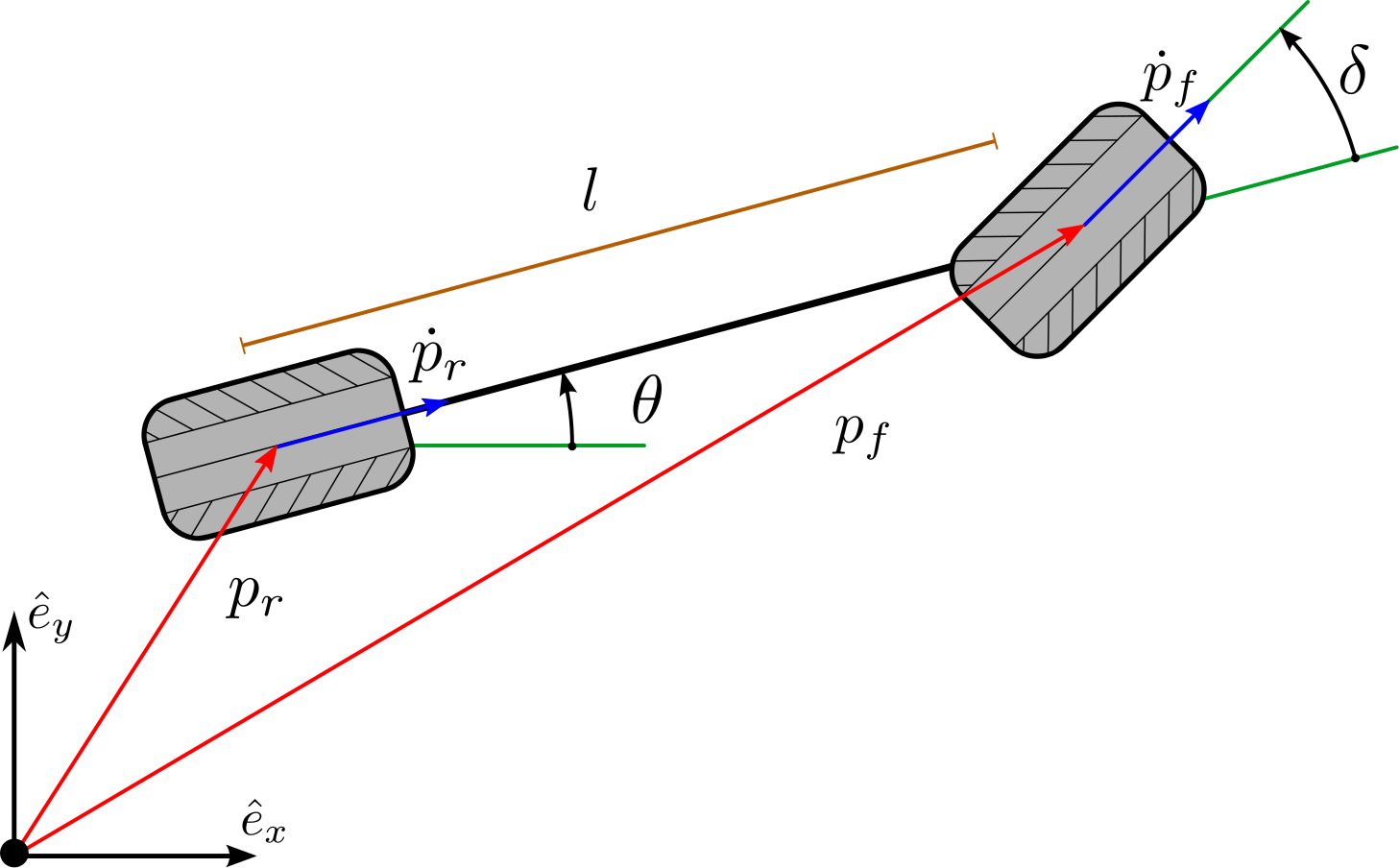}\caption{Kinematics of the single track model. $p_{r}$ and $p_{f}$ are the ground contact points of the rear and front tire respectively. 
$\theta$ is the vehicle heading. 
Time derivatives of $p_{r}$ and $p_{f}$ are restricted by the nonholonomic constraint to the direction indicated by the blue arrows.
$\delta$ is the steering angle of the front wheel.
\label{fig:Kinematics-of-the}}
\end{figure}

The following is a derivation of the differential constraint in several popular coordinate systems for the configuration.
In reference to Figure \ref{fig:Kinematics-of-the}, the vectors $p_{r}$ and $p_{f}$ denote the location of the rear and front wheels in a stationary or inertial coordinate system with basis vectors $(\hat{e}_{x},\hat{e}_{y},\hat{e}_{z})$.
The heading $\theta$ is an angle describing the direction that the vehicle is facing. 
This is defined as the angle between vectors $\hat{e}_{x}$ and $p_{f}-p_{r}$. 

Differential constraints will be derived for the coordinate systems consisting of the angle $\theta$, together with the motion of one of the points $p_{r}$ as in \cite{kuwata2009real}, and $p_{f}$ as in \cite{ventures2006stanley}.      

The motion of the points $p_{r}$ and $p_{f}$ must be collinear with the wheel orientation to satisfy the no-slip assumption.
Expressed as an equation, this constraint on the rear wheel is
\begin{equation}
\left(\dot{p}_{r}\cdot\hat{e}_{y}\right)\cos(\theta)-\left(\dot{p}_{r}\cdot\hat{e}_{x}\right)\sin(\theta)=0,\label{eq:rear_wheel_nonholonomic}
\end{equation}
and for the front wheel:
\begin{equation}
\left(\dot{p}_{f}\cdot\hat{e}_{y}\right)\cos(\theta+\delta)-\left(\dot{p}_{f}\cdot\hat{e}_{x}\right)\sin(\theta+\delta)=0.\label{eq:front_wheel_nonholonomic}
\end{equation}
This expression is usually rewritten in terms of the component-wise motion of each point along the basis vectors. The motion of the rear wheel along the $\hat{e}_x$-direction is $x_{r}:=p_{r}\cdot\hat{e}_x$. Similarly, for $\hat{e}_y$-direction, $y_{r}:=p_{r}\cdot\hat{e}_{y}$. The forward speed is $v_{r}:=\dot{p}_{r}\cdot(p_{f}-p_{r})/\Vert(p_{f}-p_{r})\Vert$, which is the magnitude of $\dot{p}_{r}$  with the correct sign to indicate forward or reverse driving. In terms of the scalar quantities $x_r$, $y_r$, and $\theta$, the differential constraint is
\begin{equation}
\begin{array}{lll}
\dot{x}_{r} & = & v_{r}\cos(\theta),\\
\dot{y}_{r} & = & v_{r}\sin(\theta),\\
\dot{\theta} & = & \frac{v_{r}}{l}\tan(\delta).
\end{array}\label{eq:rear_wheel_dynamics}
\end{equation}
Alternatively, the differential constraint can be written in terms the motion of $p_f$, 
\begin{equation}
\begin{array}{lll}
\dot{x}_{f} & = & v_{f}\cos(\theta+\delta),\\
\dot{y}_{f} & = & v_{f}\sin(\theta+\delta),\\
\dot{\theta} & = & \frac{v_{f}}{l}\sin(\delta),
\end{array}\label{eq:front_wheel_dynamics}
\end{equation}
where the front wheel forward speed $v_{f}$ is now used.
The front wheel speed, $v_{f}$, is related to the rear wheel speed by 
\begin{equation}
\frac{v_{r}}{v_{f}}=\cos(\delta).\label{eq:front_rear_vel}
\end{equation}

The planning and control problems for this model involve selecting the steering angle $\delta$ within the mechanical limits of the vehicle $\delta\in[\delta_{min},\delta_{max}]$, and forward speed $ v_r $ within an acceptable range, $v_r\in[v_{min},v_{max}].$

A simplification that is sometimes utilized, e.g. \cite{samson1992path}, is to select the heading rate $\omega$ instead of steering angle $\delta$. 
These quantities are related by
\begin{equation}
\delta=\arctan\left(\frac{l\omega}{v_{r}}\right),
\end{equation}
simplifying the heading dynamics to 
\begin{equation}
\dot{\theta}=\omega,\quad\omega\in\left[\frac{v_{r}}{l}\tan\left(\delta_{min}\right),\frac{v_{r}}{l}\tan\left(\delta_{max}\right)\right].\label{eq:unicycle}
\end{equation}
In this situation, the model is sometimes referred to as the unicycle model since it can be derived by considering the motion of a single wheel.  

An important variation of this model is the case when $v_r$ is fixed. This is sometimes referred to as the Dubins car, after Lester Dubins who derived the minimum time motion between to points with prescribed tangents \cite{dubins1957oncurves}.
Another notable variation is the Reeds-Shepp car for which minimum length paths are known when $v_r$ takes a single forward and reverse speed \cite{reeds1990optimal}. 
These two models have proven to be of some importance to motion planning and will be discussed further in Section \ref{section_mp}.

The kinematic models are suitable for planning paths at low speeds (e.g. parking maneuvers and urban driving) where inertial effects are small in comparison to the limitations on mobility imposed by the no-slip assumption.
A major drawback of this model is that it permits instantaneous steering angle changes which can be problematic if the motion planning module generates solutions with such instantaneous changes. 

Continuity of the steering angle can be imposed by augmenting \eqref{eq:front_wheel_dynamics}, where the steering angle integrates a commanded rate as in  \cite{murray1993nonholonomic}. 
Equation \eqref{eq:front_wheel_dynamics} becomes 
\begin{equation} 
\begin{array}{lll}
\dot{x}_{f} & = & v_{f}\cos(\theta+\delta),\\
\dot{y}_{f} & = & v_{f}\sin(\theta+\delta),\\
\dot{\theta} & = & \frac{v_{f}}{l}\sin(\delta),\\
\dot{\delta} & = & v_{\delta}.
\end{array}\label{eq:Steering_rate_control}
\end{equation}
In addition to the limit on the steering angle, the steering rate can now be limited: $v_{\delta}\in\left[\dot{\delta}_{min},\dot{\delta}_{max}\right]$.
The same problem can arise with the car's speed $v_r$ and can be resolved in the same way.
The drawback to this technique is the increased dimension of the model which can complicate motion planning and control problems.

\begin{short}
While the kinematic bicycle model and simple variations are very useful for motion planning and control, models considering wheel slip \cite{bakker1987tyre}, inertia \cite{rajamani2011vehicle, velenis2005minimum, peters2011differential, rucco2012computing}, and chassis dynamics \cite{velenis2005minimum} can better utilize the vehicle's capabilities for executing agile maneuvers. 
These effects become significant when planning motions with high acceleration and jerk.
\end{short}
\begin{full}
The choice of coordinate system is not limited to using one of the wheel locations as a position coordinate.  
For models derived using principles from classical mechanics it can be convenient to use the center of mass as the position coordinate as in \cite{velenis2005minimum,rucco2012computing}, or the center of oscillation as in \cite{hwan2013optimal,peters2011differential}.

\subsection{Inertial Effects}\label{inertia}

When the acceleration of the vehicle is sufficiently large, the no-slip assumption between the tire and ground becomes invalid.
In this case a more accurate model for the vehicle is as a rigid body satisfying basic momentum principles. 
That is, the acceleration is proportional to the force generated by the ground on the tires.
Taking $p_{c}$ to be the vehicles center of mass, and a coordinate of the configuration (cf. Figure \ref{fig:slipModel}), the motion of the vehicle is governed by 
\begin{equation}
\begin{array}{rcl}
m\ddot{p}_{c} & = & F_{f}+F_{r},\\
I_{zz}\ddot{\theta} & = & (p_{c}-p_{f})\times F_{f}+(p_{c}-p_{r})\times F_{r}, 
\end{array}\label{eq:newtons}
\end{equation}
where $F_{r}$ and $F_{f}$ are the forces applied to the vehicle by the ground through the ground-tire interaction, $m$ is the vehicles total mass, and $I_{zz}$ is the polar moment
of inertia in the $\hat{e}_{z}$ direction about the center of mass.
In the following derivations we tacitly neglect the motion of $p_{c}$
in the $\hat{e}_{z}$ direction with the assumptions that the road
is level, the suspension is rigid and vehicle remains on the road.

The expressions for $F_r$ and $F_f$ vary depending on modeling assumptions \cite{rajamani2011vehicle, velenis2005minimum, peters2011differential, rucco2012computing}, but in any case the expression can be tedious to derive. 
Equations \eqref{eq:front_rear_kinematics}-\eqref{eq:pacejka} therefore provide a detailed derivation as a reference.

The force between the ground and tires is modeled as being dependent on the rate that the tire slips on the ground. 
Although the center of mass serves as a coordinate for the configuration, the velocity of each wheel relative to the ground is needed to determine this relative speed. 
The kinematic relations between these three points are 
\begin{equation}
\begin{array}{rcl}
p_{r} & = & p_{c}+\left(\begin{array}{c}
-l_{r}\cos(\theta)\\
-l_{r}\sin(\theta)\\
0
\end{array}\right),\\
\dot{p}_{r} & = & \dot{p}_{c}+\left(\begin{array}{c}
0\\
0\\
\dot{\theta}
\end{array}\right)\times\left(\begin{array}{c}
-l_{r}\cos(\theta)\\
-l_{r}\sin(\theta)\\
0
\end{array}\right),\\
p_{f} & = & p_{c}+\left(\begin{array}{c}
l_{f}\cos(\theta)\\
l_{f}\sin(\theta)\\
0
\end{array}\right),\\
\dot{p}_{f} & = & \dot{p}_{c}+\left(\begin{array}{c}
0\\
0\\
\dot{\theta}
\end{array}\right)\times\left(\begin{array}{c}
l_{f}\cos(\theta)\\
l_{f}\sin(\theta)\\
0
\end{array}\right).
\end{array}\label{eq:front_rear_kinematics}
\end{equation}
These kinematic relations are used to determine the velocities of the point on each tire in contact with the ground, $s_{r}$ and $s_{f}$. 
The velocity of these points are referred to as the tire slip velocity. 
In general, $s_{r}$ and $s_{f}$ differ from $\dot{p}_{r}$ and $\dot{p}_{f}$ through the angular velocity of the wheel. 
The kinematic relation is 
\begin{equation}
\begin{array}{rcl}
s_{r} & = & \dot{p}_{r}+\omega_{r}\times R,\\
s_{f} & = & \dot{p}_{f}+\omega_{f}\times R.
\end{array}\label{eq:slip}
\end{equation}
The angular velocities of the wheels are given by 
\begin{equation}
\omega_{r}=\begin{pmatrix}\Omega_{r}\sin(\theta)\\
-\Omega_{r}\cos(\theta)\\
0
\end{pmatrix},\quad\omega_{f}=\begin{pmatrix}\Omega_{f}\sin(\theta+\delta)\\
-\Omega_{f}\cos(\theta+\delta)\\
0
\end{pmatrix},\label{eq:omega_wheels}
\end{equation}
and $R=\begin{pmatrix}0, & 0, & -r\end{pmatrix}^{T}$. 
The wheel radius is the scalar quantity $r$, and $\Omega_{\{r,f\}}$ are the angular speeds of each wheel relative to the car. 
This is illustrated for the rear wheel in Figure \ref{fig:slip}. 

\begin{figure}[b!]
\centering{}\includegraphics[width=8cm]{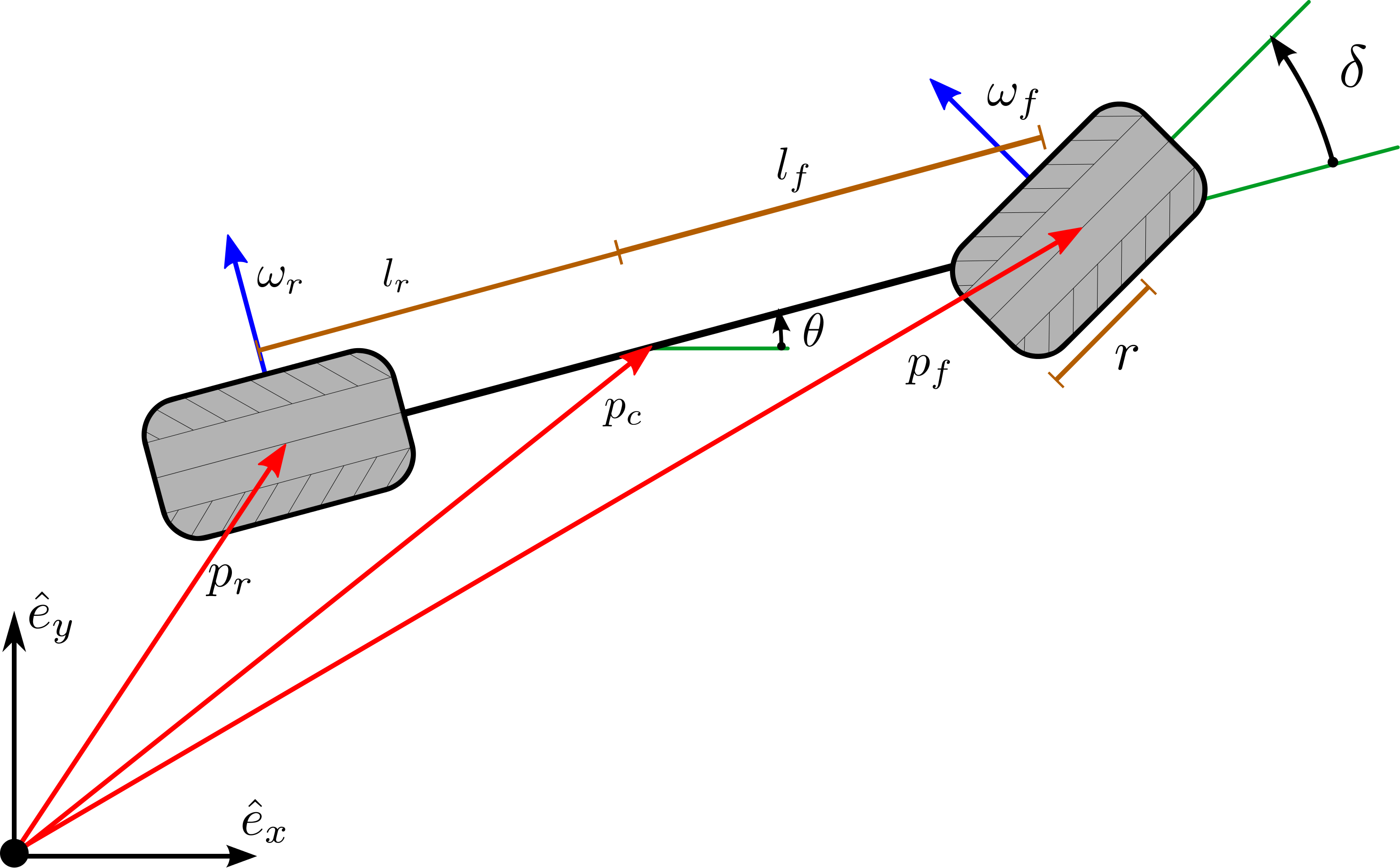}\caption{Illustration of single track model kinematics without the no-slip assumption. $\omega_{ \{ r,f \} }$ are relative angular velocities of the wheels with respect to the vehicle. \label{fig:slipModel}}
\end{figure}
\begin{figure}[bt!]
\centering{}\includegraphics[width=6cm]{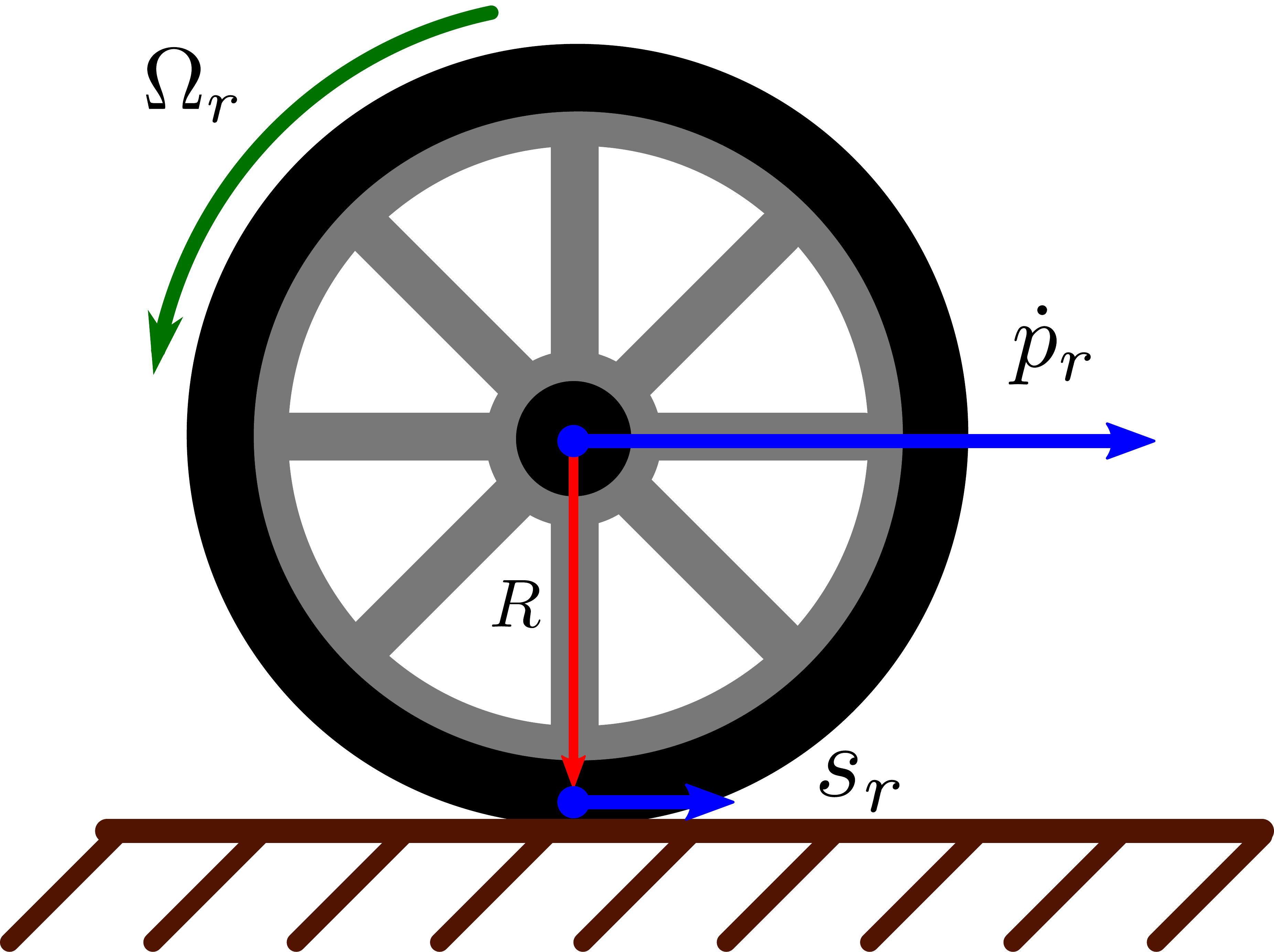}\caption{Illustration of the rear wheel kinematics \textit{in two dimensions}
showing the wheel slip, $s_{r}$, in relation to the rear wheel velocity,
$\dot{p}_{r}$, and angular speed, $\Omega_{r}$. In general,
$s_{r}$ and $\dot{p}_{r}$ are not collinear and may have nonzero
components normal to the plane depicted. \label{fig:slip}}
\end{figure}

Under static conditions, or when the height of the center of mass
can be approximated as $p_{c}\cdot\hat{e}_{z}\approx0$, the component
of the force normal to the ground, $F_{\{r,f\}}\cdot\hat{e}_{z}$
can be computed from a static force-torque balance as 
\begin{equation}
F_{f}\cdot\hat{e}_{z}=\frac{l_{r}mg}{l_{f}+l_{r}},\quad F_{r}\cdot\hat{e}_{z}=\frac{l_{f}mg}{l_{f}+l_{r}}.\label{eq:normal_force}
\end{equation}
The normal force is then used to compute the traction force on each
tire together with the slip and a friction coefficient model, $\mu$,
for the tire behavior. The traction force on the rear tire is given
component-wise by

\begin{equation}
\begin{array}{ccc}
F_{r}\cdot\hat{e}_{x} & = & -\frac{\left(F_{r}\cdot\hat{e}_{z}\right)\mu\left(\frac{\left\Vert s_{r}\right\Vert }{\Omega_{r}r}\right)s_{r}}{\left\Vert s_{r}\right\Vert }\cdot\hat{e}_{x},\\
F_{r}\cdot\hat{e}_{y} & = & -\frac{\left(F_{r}\cdot\hat{e}_{z}\right)\mu\left(\frac{\left\Vert s_{r}\right\Vert }{\Omega_{r}r}\right)s_{r}}{\left\Vert s_{r}\right\Vert }\cdot\hat{e}_{y}.
\end{array}\label{eq:Force_formula}
\end{equation}
The same expression describes the front tire with the $r$-subscript
replaced by an $f$-subscript. The formula above models the traction
force as being anti-parallel to the slip with magnitude proportional
to the normal force with a nonlinear dependence on the slip ratio
(the magnitude of the slip normalized by $\Omega_{r}r$ for the rear
and $\Omega_{f}r$ for the front). Combining \eqref{eq:front_rear_kinematics}-\eqref{eq:pacejka}
yields expressions for the net force on each wheel of the car in terms
of the control variables, generalized coordinates, and their velocities.
Equation \eqref{eq:Force_formula}, together with the following model for $\mu$,
\begin{equation}
\mu\left(\frac{\left\Vert s_{r}\right\Vert }{\Omega_{r}r}\right)=D\sin\left(C\arctan\left(B\frac{\left\Vert s_{r}\right\Vert }{\Omega_{r}r}\right)\right),\label{eq:pacejka}
\end{equation}
are a frequently used model for tire interaction with the
ground. Equation \eqref{eq:pacejka} is a simplified version of the well known model due to Pacejka \cite{bakker1987tyre}.

The rotational symmetry of \eqref{eq:Force_formula} together with the
peak in \eqref{eq:pacejka} lead to a maximum norm force that the
tire can exert in any direction. This peak is referred to as the friction circle depicted
in Figure \ref{fig:Friction-parameter-as}. 
\begin{figure}[ht!]
\centering{}\includegraphics[width=8.5cm,trim=3mm 35mm 3mm 40mm]{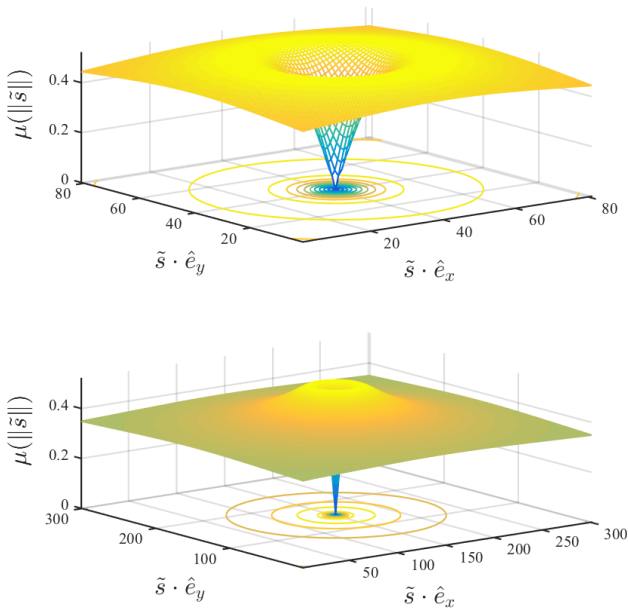}\caption{A zoomed in view of the wheel slip to traction force map at each tire (top) and a zoomed out view emphasizing the peak that defines the friction circle (bottom); cf. Equation \eqref{eq:pacejka}.  \label{fig:Friction-parameter-as}}
\end{figure}

The models discussed in this section appear frequently in the literature on motion planning and control for driverless cars. They are suitable for the motion planning and control tasks discussed in this survey. However, lower level control tasks such as electronic stability control and active suspension systems typically use more sophisticated models for the chassis, steering and, drive-train. 
\end{full}

\SetKwFunction{samplepoints}{sample\textrm{-}points}
\SetKwFunction{neighbors}{neighbors}
\SetKwFunction{colfree}{col\textrm{-}free}
\SetKwFunction{path}{path}
\SetKwFunction{connect}{connect}
\SetKwFunction{steer}{steer}
\SetKwFunction{select}{select}
\SetKwFunction{extend}{extend}

\section{Motion Planning \label{section_mp}}
The motion planning layer is responsible for computing a safe, comfortable, and dynamically feasible trajectory from the vehicle's current configuration to the goal configuration provided by the behavioral layer of the decision making hierarchy. Depending on context, the goal configuration may differ. For example, the goal location may be the center point of the current lane a number of meters ahead in the direction of travel, the center of the stop line at the next intersection, or the next desired parking spot.
The motion planning component accepts information about static and dynamic obstacles around the vehicle and generates a collision-free trajectory that satisfies dynamic and kinematic constraints on the motion of the vehicle. Oftentimes, the motion planner also minimizes a given objective function. 
In addition to travel time, the objective function may penalize hazardous motions or motions that cause passenger discomfort.
In a typical setup, the output of the motion planner is then passed to the local feedback control layer. In turn, feedback controllers generate an input signal to regulate the vehicle to follow this given motion plan. %output trajectory.

A motion plan for the vehicle can take the form of a path or a trajectory. %be represented as a path or as a trajectory. 
Within the path planning framework, 
%the dynamics of the vehicle are not modeled explicitly and only static obstacles are considered. 
the solution \empty{path} is represented as a function $\sigma(\alpha):\ [0,1] \rightarrow \mathcal{X}$, where $\mathcal{X}$ is the configuration space of the vehicle. 
Note that such a solution does not prescribe how this path should be followed and one can either choose a velocity profile for the path or delegate this task to lower layers of the decision hierarchy. 
Within the trajectory planning framework, the control execution time is explicitly considered. 
This consideration allows for direct modeling of vehicle dynamics and dynamic obstacles. In this case, the solution \empty{trajectory} is represented as a time-parametrized function $\pi(t):\:[0,T]\rightarrow \mathcal{X}$, where $T$ is the planning horizon. Unlike a path, the trajectory prescribes how the configuration of the vehicle evolves over time.

In the following two sections, we provide a formal problem definition of the path planning and trajectory planning problems and review the main complexity and algorithmic results for both formulations.

\subsection{Path Planning}
The path planning problem is to find a path $\sigma(\alpha):\ [0,1]\rightarrow \mathcal{X}$ in the configuration space $\mathcal{X}$ of the vehicle (or more generally, a robot)  that starts at the initial configuration and reaches the goal region while satisfying given global and local constraints.
Depending on whether the quality of the solution path is considered, the terms  \emph{feasible} and \emph{optimal} are used to describe this path. % we use the terms \emph{feasible} path planning and \emph{optimal} path planning. 
Feasible path planning refers to the problem of determining a path that satisfies some given problem constraints without focusing on the quality of the solution; whereas optimal path planning refers to the problem of finding a path that optimizes some quality criterion subject to given constraints.

%The quality criterion is modeled using an cost functional that assigns a real-valued cost to each possible path. 
%\subsubsection*{Formal definition}
The optimal path planning problem can be formally stated as follows. Let $\mathcal{X}$ be the configuration space of the vehicle and let 
$\Sigma(\mathcal{X})$ denote the set of all continuous functions $[0,1] 
\rightarrow \mathcal{X}$. The initial configuration of the vehicle is 
$\mathbf{x}_{\mathrm{init}} \in \mathcal{X}$. The path is required to end in a 
goal region $X_\mathrm{goal} \subseteq \mathcal{X}$. The set of all allowed 
configurations of the vehicle is called the free configuration space and 
denoted $\mathcal{X}_\mathrm{free}$. Typically, the free configurations are 
those that do not result in collision with obstacles, but the 
free-configuration set can also represent other holonomic constraints on the 
path. 
The differential constraints on the path are represented by a predicate 
$D(\mathbf{x}, \mathbf{x}', \mathbf{x}'',\ldots)$ and can be used to enforce 
some degree of smoothness of the path for the vehicle, such as the bound on the 
path curvature and/or the rate of curvature. For example, in the case of 
$\mathcal{X} \subseteq \mathbb{R}^2$, the differential constraint may enforce 
the maximum curvature $\kappa$ of the path  using Frenet-Serret formula as 
follows:
$$ D(\mathbf{x}, \mathbf{x}', \mathbf{x}'',\ldots) \: \Leftrightarrow \: \frac{ 
\Vert \mathbf{x}' \times \mathbf{x}'' \Vert }{\Vert \mathbf{x}' \Vert^3} \leq 
\kappa.$$
Further, let $J(\sigma): \: \Sigma(\mathcal{X}) \rightarrow \mathbb{R}$ be the 
cost functional. Then, the optimal version of the path planning problem can be 
generally stated as follows.

\begin{problem}[Optimal path planning]
Given a 5-tuple 
$(\mathcal{X}_\mathrm{free},\mathbf{x}_{\mathrm{init}},X_\mathrm{goal},D,J)$ 
find $\sigma^{*}=$
$$
\begin{array}{rll}
\underset{\sigma \in \Sigma(\mathcal{X})}{\argmin} \; & J(\sigma) \\  
\text{ subj. to } & \sigma(0) = \mathbf{x}_{\mathrm{init}} \text{ and } \sigma(1) \in X_\mathrm{goal} \\
& \sigma(\alpha) \in \mathcal{X}_\mathrm{free}  & \forall \alpha \in [0,1]  \\
& D(\sigma(\alpha), \sigma'(\alpha), \sigma''(\alpha), \ldots) & \forall \alpha 
\in [0,1]. \\
\end{array}
$$
\label{pr:optimal_path_planning}
\end{problem}

The problem of feasible and optimal path planning has been studied extensively in the past few decades. The complexity of this problem is well understood, and many practical algorithms have been developed.

\begin{short}
The problem of finding an optimal path subject to holonomic and differential constraints as formulated in Problem~\ref{pr:optimal_path_planning} is known to be PSPACE-hard~\cite{reif1979complexity}. 
This means that it is at least as hard as solving any NP-complete problem and thus, assuming $\mathrm{P} \neq \mathrm{NP}$, there is no efficient (polynomial-time) algorithm that is able to solve all instances of the problem. 
Research attention has since been directed toward studying approximate methods, or approaches to subsets of the general motion planning problem.

In particular, a shortest path for a holonomic vehicle in a 2-D environment with polygonal obstacles can be obtained using visibility graph approach in $O(n^{2})$~\cite{overmars1988new}. 
Also, a shortest paths for a car-like vehicle in the absence of obstacles can be constructed analytically: Dubins \cite{dubins1957oncurves} has shown that the shortest path having curvature bounded by $\kappa$ between given two points $p_1,p_2$ and with prescribed tangents $\theta_1,\theta_2$ is a curve consisting of at most three segments, each one being either a circular arc segment or a straight line. 
Later, Reeds and Shepp \cite{reeds1990optimal} extended the method for a car that can move both forwards and backwards.
\end{short}

\begin{full}
\subsubsection*{Complexity}
%A significant body of literature is devoted to studying the complexity of motion planning problems demonstrating that these problems are difficult solve.
A significant body of literature is devoted to studying the complexity of motion planning problems. 
The following is a brief survey of some of the major results regarding the computational complexity of these problems. 

The problem of finding an optimal path subject to holonomic and differential constraints as formulated in Problem~\ref{pr:optimal_path_planning} is known to be PSPACE-hard~\cite{reif1979complexity}. 
This means that it is at least as hard as solving any NP-complete problem and thus, assuming $\mathrm{P} \neq \mathrm{NP}$, there is no efficient (polynomial-time) algorithm able to solve all instances of the problem. 
Research attention has since been directed toward studying approximate methods, or approaches to subsets of the general motion planning problem.
 
Initial research focused primarily on feasible (i.e., non-optimal) path planning for a holonomic vehicle model in polygonal/polyhedral environments. 
That is, the obstacles are assumed to be polygons/polyhedra and there are no differential constraints on the resulting path. 
In 1970, Reif~\cite{reif1979complexity} found that an obstacle-free path for a holonomic vehicle, whose footprint can be described as a single polyhedron, can be found in polynomial time in both 2-D and 3-D environments. 
%
%\bpmargin{Later, Schwartz and Sharir demonstrated that feasible path planning for a vehicle consisting of a \emph{fixed} number of freely linked and independently controlled polyhedra in a 3-D environment with polyhedral obstacles is polynomially solvable -- however, the provided algorithm is polynomial in the size of the description of the environment, but exponential in the number of degrees of 
%freedom~\cite{schwartz1983onthepiano}.}{not cars} 
%
%On the other hand, when the vehicle 
%consists of $n$ freely-linked bodies moving in a 2-D/3-D environment and $n$ is 
%not fixed, then the problem of finding a collision-free path for such a robot 
%becomes PSPACE-hard~\cite{reif1979complexity,hopcroft1984movement}. 
%
Canny~\cite{canny1988complexity} has shown that the problem of feasible path planning in a free space represented using polynomials is in PSPACE, which rendered the decision version of feasible path planning without differential constraints as a PSPACE-complete problem.

For the optimal planning formulation, where the objective is to find the \emph{shortest} obstacle-free path. 
It has been long known that a shortest path for a holonomic vehicle in a 2-D environment with polygonal obstacles can be found in polynomial time~\cite{lozano1979analgorithm,storer1994shortest}. 
More precisely, it can be computed in time $O(n^{2})$, where $n$ is the number of vertices of the polygonal obstacles~\cite{overmars1988new}.
This can be solved by constructing and searching the so-called visibility graph~\cite{berg2000computational}. 
In contrast, Lazard, Reif and Wang~\cite{lazard1998complexity} established that the problem of finding a shortest curvature-bounded path in a 2-D plane amidst polygonal obstacles (i.e., a path for a car-like robot) is NP-hard, which suggests that there is no known polynomial time algorithm for finding a shortest path for a car-like robot among polygonal obstacles. 
A related result is that, that the existence of a curvature constrained path in polygonal environment can be 
decided in EXPTIME~\cite{fortune1991planning}.

A special case where a solution can be efficiently computed is the shortest curvature bounded path in an obstacle free environment.
Dubins \cite{dubins1957oncurves} has shown that the shortest path having curvature bounded by $\kappa$ between given two points $p_1,p_2$ and with prescribed tangents $\theta_1,\theta_2$ is a curve consisting of at most three segments, each one being either a circular arc segment or a straight line. 
Reeds and Shepp \cite{reeds1990optimal} extended the method for a car that can move both forwards and backwards.
Another notable case due to Agraval et al.~\cite{agarwal2002curvature} is an $O(n^2\log n)$ algorithm for finding a shortest path with bounded curvature inside a convex polygon.
Similarly, Boissonnat and Lazard~\cite{boissonnat1996polynomial} proposed a polynomial time algorithm for finding an exact curvature-bounded path in environments where obstacles have bounded-curvature boundary.
\end{full}

%\subsubsection*{Summary of Solution Techniques}
% 
Since for most problems of interest in autonomous driving, exact algorithms with practical computational complexity are unavailable~\cite{lazard1998complexity}, one has to resort to more general, numerical solution methods. 
These methods generally do not find an exact solution, but attempt to find a satisfactory solution or a sequence of feasible solutions that converge to the optimal solution.
The utility and performance of these approaches are typically quantified by the class of problems for which they are applicable as well as their guarantees for converging to an optimal solution.     
The numerical methods for path planning can be broadly divided in three main categories: 

\noindent\emph{ 
Variational methods} represent the path as a function parametrized by a finite-dimensional vector and the optimal path is sought by optimizing over the vector parameter using non-linear continuous optimization techniques. 
These methods are attractive for their rapid convergence to \emph{locally} optimal solutions; however, they typically lack the ability to find globally optimal solutions unless an appropriate initial guess in provided.
For a detailed discussion on variational methods, see Section~\ref{sec:variational-methods}.

\noindent\emph{ Graph-search methods} 
discretize the configuration space of the vehicle as a graph, where the vertices represent a finite collection of vehicle configurations and the edges represent  transitions between vertices.  
The desired path is found by performing a search for a minimum-cost path in such a graph. Graph search methods are not prone to getting stuck in local minima, however, they are limited to optimize only over a finite set of paths, namely those that can be constructed from the atomic motion primitives in the graph.  
%
%Graph search methods are effective in unstructured driving scenarios.
%
%Effective implementations are called resolution complete when the cost of paths in the graph converge to the optimal cost for the problem as the size of the graph tends to infinity.
%
%One drawback is that searching large graphs can become computationally prohibitive and suboptimality is difficult to bound in terms of the size of the graph.
%
%Additionally, the graph is sequentially constructed and then searched. 
%
%So if the graph sizeis too small there may be no feasible solution within the graph, let alone one who's cost is sufficiently close to the optimal cost.
%
%In such a circumstance an entirely new, larger graph may need to be constructed. 
%
For a detailed discussion about graph search methods, see Section~\ref{sec:state-space-discretization}.

\noindent\emph{ Incremental search methods} sample the configuration space and incrementally build a reachability graph (oftentimes a tree) that maintains a discrete set of reachable configurations and feasible transitions between them. 
Once the graph is large enough so that at least one node is in the goal region, the desired path is obtained by tracing the edges that lead to that node from the start configuration. %tracing the branch ending in the goal region backwards to the start configuration. 
In contrast to more basic graph search methods, sampling-based methods incrementally increase the size of the graph until a satisfactory solution is found within the graph. 
For a detailed discussion about incremental search methods, see Section~\ref{sec:sampling-based-methods}.

Clearly, it is possible to exploit the advantages of each of these methods by combining them. 
For example, one can use a coarse graph search to obtain an initial guess for the variational method as reported in \cite{lamiraux04kinodynamic} and \cite{boyer06trajectory}. 
A comparison of key properties of select path planning methods is given in Table~\ref{tab:motion_planning_comparison}. In the remainder of this section, we will discuss the path planning algorithms and their properties in detail.

\begin{table*}
\begin{tabular}{|>{\raggedright}p{3.6cm}|>{\centering}p{3cm}|>{\centering}p{2.25cm}|>{\centering}p{2.1cm}|>{\centering}p{3.3cm}|>{\centering}p{1.25cm}|}%\rowcolor{dark_grey}
\hline 
\cellcolor{dark_grey} & \cellcolor{dark_grey} {\small Model assumptions} & \cellcolor{dark_grey} {\small{}Completeness} & \cellcolor{dark_grey} {\small{}Optimality} & \cellcolor{dark_grey} {\small{}Time Complexity} & \cellcolor{dark_grey} {\small{}Anytime}\tabularnewline
\hline 
\hline %\cellcolor{grey}
\multicolumn{6}{|c|}{{\small{} \cellcolor{grey} Geometric Methods}}\tabularnewline
\hline 
{\small{}\multirow{2}{*}{Visibility graph~\cite{nilsson1969mobile}}} & {\small{}2-D polyg. conf. space,\\no diff. constraints} & {\small{}\multirow{2}{*}{Yes}} & {\small{}\multirow{2}{*}{Yes\,$^a$}} & {\small{} \multirow{2}{*}{$O(n^{2})$\cite{overmars1988new}\,$^b$} } & {\small{}\multirow{2}{*}{No}}\tabularnewline
\hline 
{\small{}Cyl. algebr. decomp.~\cite{canny1987new}} & {\small{}No diff. constraints} & {\small{}Yes} & {\small{}No} & {\small{}Exp. in dimension. \cite{canny1988complexity}} & {\small{}No}\tabularnewline
\hline %\rowcolor{grey}
\multicolumn{6}{|c|}{{\small{} \cellcolor{grey} Variational Methods}}\tabularnewline
\hline 
{\small{}Variational methods (Sec~\ref{sec:variational-methods})} & {\small{} Lipschitz-continuous Jacobian} & {\small{}\multirow{2}{*}{No}} & {\small{}\multirow{2}{*}{Locally optimal}} & {\small{} \multirow{2}{*}{$O(1/\epsilon)$ \cite{bertsekas1999nonlinear}\,$^{k,l}$}} & {\small{}\multirow{2}{*}{Yes}}\tabularnewline
\hline %\rowcolor{grey}
\multicolumn{6}{|c|}{{\small{} \cellcolor{grey} Graph-search Methods}}\tabularnewline
\hline 
{\small{}Road lane graph + Dijkstra~(Sec~\ref{sec:road_lane_graph})} & {\small{}\multirow{2}{*}{Arbitrary}} & {\small{}\multirow{2}{*}{No\,$^c$}} & {\small{}\multirow{2}{*}{No\,$^d$}} & {\small{}\multirow{2}{*}{$O(n + m\log m)$ \cite{fredman1987fibonacci}\,$^{e,f}$}} & {\small{}\multirow{2}{*}{No}}\tabularnewline
\hline 
{\small{}Lattice/tree of motion prim. + Dijkstra~(Sec~\ref{sec:sampling-based-methods})} & {\small{}\multirow{2}{*}{Arbitrary}} & {\small{}\multirow{2}{*}{No\,$^c$}} & {\small{}\multirow{2}{*}{No\,$^d$}} & {\small{}\multirow{2}{*}{$O(n + m\log m)$ \cite{fredman1987fibonacci}\,$^{e,f}$}} & {\small{}\multirow{2}{*}{No}}\tabularnewline
\hline 
{\small{}\multirow{2}{*}{PRM~\cite{kavraki1998analysis}~$^g$ + Dijkstra}} & {\small{}Exact steering procedure available} & {\small{}Probabilistically complete} \cite{karaman2011sampling}\,$^\ast$ & {\small{}Asymptotically optimal}\,$^\ast$ \cite{karaman2011sampling} & {\small{}\multirow{2}{*}{$O(n^{2})$ \cite{karaman2011sampling}\,$^{h,f,\ast}$}} & {\small{}\multirow{2}{*}{No}}\tabularnewline
\hline 
{\small{}\multirow{3}{*}{PRM{*}~\cite{karaman2011sampling,schmerling2015optimal} + Dijkstra}} & {\small{}\multirow{3}{*}{\begin{tabular}{@{}c@{}}Exact steering\\ procedure available\end{tabular}}} & {\small{}Probabilistically complete \cite{karaman2011sampling,schmerling2015optimal,karaman2013rrtsnonholonom}}\,$^{\ast,\dagger}$ & {\small{}Asymptotically optimal \cite{karaman2011sampling,schmerling2015optimal,karaman2013rrtsnonholonom}}\,$^{\ast,\dagger}$ & {\small{}\multirow{3}{*}{\begin{tabular}{@{}c@{}}$O(n\,\log n)$~\cite{karaman2011sampling},\\ \cite{karaman2013rrtsnonholonom}\,$^{h,f,\ast,\dagger}$\end{tabular}}} & {\small{}\multirow{3}{*}{No}}\tabularnewline
\hline 
{\small{}\multirow{2}{*}{RRG~\cite{karaman2011sampling} + Dijkstra}} & {\small{}Exact steering procedure available} & {\small{}Probabilistically complete \cite{karaman2011sampling}\,$^{\ast}$ } & {\small{}Asymptotically optimal \cite{karaman2011sampling}}\,$^{\ast}$ & {\small{}\multirow{2}{*}{$O(n\,\log n)$~\cite{karaman2011sampling}\,$^{h,f,\ast}$}} & {\small{}\multirow{2}{*}{Yes}}\tabularnewline
\hline %\rowcolor{grey}
\multicolumn{6}{|c|}{{\small{} \cellcolor{grey} Incremental Search}}\tabularnewline
\hline 
{\small{}\multirow{2}{*}{RRT~\cite{lavalle2001randomized}}} & {\small{}\multirow{2}{*}{Arbitrary}} & {\small{}Probabilistically complete~\cite{lavalle2001randomized}}\,$^{i,\ast}$ & {\small{}\multirow{2}{*}{Suboptimal\,\cite{karaman2011sampling}\,$^\ast$}} & \multirow{2}{*}{$O(n\,\log n)$~\cite{karaman2011sampling}\,$^{h,f,\ast}$} & {\small{}\multirow{2}{*}{Yes}}\tabularnewline
\hline 
{\small{}\multirow{3}{*}{RRT{*}~\cite{karaman2011sampling}}} & {\small{}\multirow{3}{*}{\begin{tabular}{@{}c@{}}Exact steering\\ procedure available\end{tabular}}} & {\small{}Probabilistically complete~\cite{karaman2011sampling,karaman2013rrtsnonholonom}}\,$^{\ast,\dagger}$ & {\small{}Asymptotically optimal~\cite{karaman2011sampling,karaman2013rrtsnonholonom}}\,$^{\ast,\dagger}$ &  {\small{}\multirow{3}{*}{\begin{tabular}{@{}c@{}}$O(n\,\log n)$~\cite{karaman2011sampling},\\ \cite{karaman2013rrtsnonholonom}\,$^{h,f,\ast,\dagger}$\end{tabular}}} & {\small{}\multirow{3}{*}{Yes}}\tabularnewline
\hline 
{\small{}\multirow{2}{*}{SST{*}~\cite{li2015sparse}}} & {\small{}Lipschitz-continuous dynamics} & {\small{}Probabilistically complete~\cite{li2015sparse}}\,$^\dagger$ & {\small{}Asymptotically optimal~\cite{li2015sparse}\,$^\dagger$} & \multirow{2}{*}{N/A\,$^j$} & {\small{}\multirow{2}{*}{Yes}}\tabularnewline
\hline 
\end{tabular}
\caption{Comparison of path planning methods. \textbf{Legend:} 
$a$: for the shortest path problem; 
$b$: $n$ is the number of points defining obstacles; 
$c$: complete only w.r.t. the set of paths induced by the given graph; 
$d$: optimal only w.r.t. the set of paths induced by the given graph; 
$e$: $n$ and $m$ are the number of edges and vertices in the graph respectively; 
$f$: assuming $O(1)$ collision checking; 
$g$: batch version with fixed-radius connection strategy;
$h$: $n$ is the number of samples/algorithm iterations;
$i$: for certain variants;
$j$: not explicitly analyzed;
$k$: $\epsilon$ is the required distance from the optimal cost;
$l$: faster rates possible with additional assumptions;
$\ast$: shown for systems without differential constraints;
$\dagger$: shown for some class of nonholonomic systems.
} 

\label{tab:motion_planning_comparison}
\end{table*}

\subsection{Trajectory Planning}\label{section_trajectory_planning}

The motion planning problems in dynamic environments or with dynamic constraints may be more suitably formulated in the trajectory planning framework, in which the solution of the problem is a trajectory, i.e. a time-parametrized function $\pi(t):\:[0,T]\rightarrow 
\mathcal{X}$ prescribing the evolution of the configuration of the vehicle in time.

%\subsubsection*{Formal definition}
Let $\Pi(\mathcal{X},T)$ denote the set of all continuous functions $[0,T] 
\rightarrow \mathcal{X}$ and $x_{\mathrm{init}} \in \mathcal{X}$.  be the initial configuration of the vehicle. The goal region is $X_\mathrm{goal} \subseteq \mathcal{X}$. 
The set of all allowed configurations at time $t \in [0,T]$ is denoted as $\mathcal{X}_\mathrm{free}(t)$ 
and used to encode holonomic constraints such as the requirement on the path to avoid collisions with static and, possibly, dynamic obstacles. The differential constraints on the trajectory are represented by a predicate 
$D(\mathbf{x}, \mathbf{x}', \mathbf{x}'',\ldots)$ and can be used to enforce dynamic constraints on the trajectory. 
Further, let $J(\pi): \: \Pi(\mathcal{X},T) \rightarrow \mathbb{R}$ be the cost 
functional. Under these assumptions, the optimal version of the trajectory planning problem can be 
very generally stated as:
\begin{problem}[Optimal trajectory planning]
Given a 6-tuple $(\mathcal{X}_\mathrm{free},\mathbf{x}_{\mathrm{init}},X_\mathrm{goal},D,J,T)$ find 
$\pi^{*}=$
$$
\begin{array}{rll}
\underset{\pi \in \Pi(\mathcal{X},T)}{\argmin} \; & J(\pi) \\  
\text{ subj. to } & \pi(0) = \mathbf{x}_{\mathrm{init}} \text{ and } \pi(T) \in X_\mathrm{goal} & \\
& \pi(t) \in \mathcal{X}_\mathrm{free} & \forall t \in [0,T]  \\
& D(\pi(t), \pi'(t), \pi''(t), \ldots) & \forall t \in [0,T]. \\
\end{array}
$$
\label{pr:optimal_trajectory_planning}
\end{problem}
\begin{short}
Since trajectory planning in a dynamic environment is a generalization of path planning in static environments, the problem remains PSPACE-hard. 
Moreover, trajectory planning in dynamic environments has been shown to be harder than path planning in the sense that some variants of the problem that are tractable in static environments become intractable when an analogous problem is considered in a dynamic environment~\cite{reif1994motion,paden2016surveyext}.
\end{short}
\begin{full}
\subsubsection*{Complexity}
Since trajectory planning in a dynamic environment is a generalization of path 
planning in static environments, the problem remains PSPACE-hard. Moreover, 
trajectory planning in dynamic environments has been shown to be harder than 
path planning in the sense that some variants of the problem that are
tractable in static environments become intractable when an analogical problem 
is considered in a dynamic environment. In particular, recall
that a shortest path for a point robot in a static 2-D polygonal environment 
can be found efficiently in polynomial time and contrast it with the result of Canny and Reif~\cite{canny1987new} establishing that finding 
velocity-bounded collision-free trajectory for a holonomic point robot amidst 
moving polygonal obstacles\footnote{In fact, the authors considered an even more 
constrained \emph{2-D asteroid avoidance problem}, where the task is to find a 
collision-free trajectory for a point robot with a bounded velocity in a 2-D 
plane with convex polygonal obstacles moving with a fixed linear speed.} is 
NP-hard. Similarly, while path 
planning for a robot with a fixed number of degrees of freedom in 3-D polyhedral 
environments is tractable, Reif and Sharir~\cite{reif1994motion} established that trajectory planning for robot with 2 degrees of freedom among translating and rotating 3-D polyhedral obstacles is PSPACE-hard. 
\end{full}

%\subsubsection*{Tractable Fragments}
%\mynote{TODO}

%\subsubsection*{Summary of Solution Techniques}
Tractable exact algorithms are not available for non-trivial trajectory planning problems occurring in autonomous driving, making the numerical methods a popular choice for the task. 
Trajectory planning problems can be numerically solved using some variational methods directly in the time domain or by converting the trajectory planning problem to path planning in a configuration space with an added time-dimension~\cite{fraichard1998trajectory}.
\shortc{A solution to such a path planning problem is then found using a path planning algorithm that can handle differential constraints and converted back to the trajectory form.}

\begin{full}
The conversion from a trajectory planning problem $(\mathcal{X}^T_\mathrm{free},\mathbf{x}^T_{\mathrm{init}}, 
X^T_\mathrm{goal},D^T,J^T,T)$ to a path planning problem 
$(\mathcal{X}^P_\mathrm{free},\mathbf{x}^P_{\mathrm{init}}, 
X^P_\mathrm{goal},D^P,J^P)$ is usually done as follows. The 
configuration space where the path planning takes place is defined as   
$\mathcal{X}^P :=\mathcal{X}^T \times [0,T]$. For any $\mathbf{y} \in 
\mathcal{X}^P$, let $t(\mathbf{y}) \in [0,T]$ denote the time component and 
$c(\mathbf{y}) \in \mathcal{X}^T$ denote the "configuration" component of the 
point $\mathbf{y}$. A path $\sigma(\alpha):\ [0,1] \rightarrow \mathcal{X}^P$ 
can be converted to a trajectory $\pi(t):\ [0,T] \rightarrow \mathcal{X}^T$ 
if the time component at start and end point of the path is 
constrained as 
$$
\begin{array}{c}
t(\sigma(0)) = 0 \\
t(\sigma(1)) = T \\
\end{array}
$$ 
and the path is monotonically-increasing, which can be enforced by  a 
differential constraint 
$$
t(\sigma'(\alpha)) > 0 \quad \quad \forall \alpha \in [0,1].
$$ 
Further, the free configuration space, initial configuration, goal region and 
differential constraints is mapped to their path planning counterparts as 
follows: 
$$
\begin{array}{rcl}
\mathcal{X}^P_\mathrm{free} &=& \{(x,t) : x \in \mathcal{X}^T_\mathrm{free}(t) \wedge t \in  [0,T] \} \\
\mathbf{x}^P_\mathrm{init} &=& (\mathbf{x}_\mathrm{init}^T,0) \\
X^P_\mathrm{goal} &=& \{ (x,T) : x \in X^T_\mathrm{goal} \}\\
D^P(\mathbf{y},\mathbf{y}',\mathbf{y}'',\ldots)&=& D^T(c(\mathbf{y}), 
\frac{c(\mathbf{y}')}{t(\mathbf{y}')}, \frac{c(\mathbf{y}'')}{t(\mathbf{y}'')}, 
\ldots).
\end{array}
$$
A solution to such a path planning problem is then found using a path planning algorithm that can handle differential constraints and converted back to the trajectory form.
\end{full}

\subsection{Variational Methods} \label{sec:variational-methods}
We will first address the trajectory planning problem in the framework of non-linear continuous optimization. In this context, the problem is often referred to as trajectory optimization.
Within this subsection we will adopt the trajectory planning formulation with the understanding that doing so does not affect generality since path planning can be formulated as trajectory optimization over the unit time interval.
%
%An effective method for obtaining optimal vehicle trajectories is by expressing the problem as a nonlinear program over the input or trajectory function spaces.
%
To leverage existing nonlinear optimization methods, it is necessary to project the infinite-dimensional function space of trajectories to a finite-dimensional vector space. In addition, most nonlinear programming techniques require the trajectory optimization problem, as formulated in Problem~\ref{pr:optimal_trajectory_planning}, to be converted into the following form
$$
\begin{array}{rll}
\underset{\pi \in \Pi(\mathcal{X},T)}{\argmin} \; & J(\pi) \\  
\text{ subj. to } & \pi(0) = \mathbf{x}_{\mathrm{init}} \text{ and } \pi(T) \in X_\mathrm{goal} & \\
& f(\pi(t),\pi'(t),\ldots)=0 & \forall t \in [0,T]  \\
& g(\pi(t),\pi'(t),\ldots) \leq 0 & \forall t \in [0,T],  \\
\end{array}
$$
where the holonomic and differential constraints are represented as a system of equality and inequality constraints.

In some applications the constrained optimization problem is relaxed to an unconstrained one using penalty or barrier functions.
In both cases, the constraints are replaced by an augmented cost functional.
With the penalty method, the cost functional takes the form 
\begin{multline*}
  \tilde J(\pi) = J(\pi) + 
  \frac{1}{\varepsilon} \int^{T}_0 \Big[ \|f(\pi, \pi', \ldots) \|^2 + \\
  \|\max(0, g(\pi, \pi', \ldots)) \|^2 \Big] \; {\rm d} t \;.
\end{multline*}
Similarly, barrier functions can be used in place of inequality constraints.
The augmented cost functional in this case takes the form
\begin{equation*}
  \tilde J(\pi) = J(\pi) + 
  \varepsilon \int^{T}_0 h(\pi(t)) {\rm d} t \;,
\end{equation*}
where the barrier function satisfies $g(\pi)<0\Rightarrow h(\pi)<\infty$, $g(\pi)\geq0\Rightarrow h(\pi)=\infty$, and $\lim_{g(\pi)\rightarrow0}\left\{ h(\pi)\right\} =\infty$.
The intuition behind both of the augmented cost functionals is that, by making $\varepsilon$ small, minima in cost will be close to minima of the original cost functional. 
An advantage of barrier functions is that local minima remain feasible, but must be initialized with a feasible solution to have finite augmented cost.
Penalty methods on the other hand can be initialized with any trajectory and optimized to a local minima. 
However, local minima may violate the problem constraints.
A variational formulation using barrier functions is proposed in \cite{rucco2012computing} where a change of coordinates is used to convert the constraint that the vehicle remain on the road into a linear constraint. 
A logarithmic barrier is used with a Newton-like method in a similar fashion to interior point methods.
The approach effectively computes minimum time trajectories for a detailed vehicle model over a segment of roadway. 

Next, two subclasses of variational methods are discussed: Direct and indirect methods.

\subsubsection*{Direct Methods}

A general principle behind direct variational methods is to restrict the approximate solution to a finite-dimensional subspace of $\Pi(\mathcal{X}, T)$.
To this end, it is usually assumed that
\begin{equation*}
\pi(t) \approx \tilde \pi(t) = \sum\limits^N_{i=1} \pi_i \phi_i(t), 
\end{equation*}
where $\pi_i$ is a coefficient from $\mathbb{R}$, and $\phi_i(t)$ are \emph{basis} functions of the chosen subspace.
%
%The particular choice of basis functions is defined by the numerical approximation method. 
%
A number of numerical approximation schemes have proven useful for representing the trajectory optimization problem as a nonlinear program. 
We mention here the two most common schemes: Numerical integrators with collocation and pseudospectral methods.
\paragraph*{1) Numerical Integrators with Collocation} 
With collocation, it is required that the approximate trajectory satisfies the constraints in a set of discrete points $\{t_j\}^M_{j=1}$. 
This requirement results in two systems of discrete constraints: A system of nonlinear equations which approximates the system dynamics
  \begin{equation*}
  f(\tilde \pi(t_j), \tilde \pi'(t_j)) = 0 \quad \forall j = 1,\ldots,M
  \end{equation*}
  and a system of nonlinear inequalities which approximates the state constraints placed on the trajectory
  \begin{equation*}
  g(\tilde \pi(t_j), \tilde \pi'(t_j)) \le 0 \quad \forall j = 1,\ldots,M \; .
  \end{equation*}
Numerical integration techniques are used to approximate the trajectory between the collocation points. 
For example, a piecewise linear basis  
%Polynomials play an important role in approximation theory, and using polynomials for interpolating the trajectory is a natural choice in numerical optimal control. 
%
%Piecewise linear basis functions are defined as follows: First, discretize the interval $[0\; T]$ into $N-1$ segments, and let $0 = t_1 < t_2 < \ldots < t_N = T$ be segment boundaries. 
%
%Second, we define 
  \begin{equation*}
  \phi_i(t) = \left\{ 
    \begin{array}{ll}
      (t - t_{i-1}) / (t_i - t_{i-1}) & \text{ if } t \in [t_{i-1}\, , \;\; t_i] \\
      (t_{i+1} - t) / (t_{i+1} - t_i) & \text{ if } t \in [t_i\, , \;\; t_{i+1}] \\
      0 & \text{otherwise}
    \end{array}
  \right. \; 
  \end{equation*}
together with collocation gives rise to the Euler integration method. 
Higher order polynomials result in the Runge-Kutta family of integration methods.
Formulating the nonlinear program with collocation and Euler's method or one of the Runge-Kutta methods is more straightforward than some other methods making it a popular choice.
An experimental system which successfully uses Euler's method for numerical approximation of the trajectory is presented in \cite{ziegler2014making}.
%

%When combined with collocation method, which we discuss later in the paper, this choice yields first-order Euler integrator.
% 
%Choosing higher-order polynomials, for example, piecewise quadratic functions, gives raise to more accurate integrators, such as trapezoid rule, Simpson rule, and Runge-Kutta scheme. 
%
In contrast to Euler's method, the Adams approximation, is investigated in \cite{kasac11conjugate} for optimizing the trajectory for a detailed vehicle model and is shown to provide improved numerical accuracy and convergence rates. 

%\paragraph{Pseudospectral Methods} Spectral methods use orthogonal basis
%  functions, such as Chebyshev or Legendre polynomials or Fourier
%  basis, for trajectory approximation. In contrast to numerical
%  integrators, these basis functions have global support which makes
%  reconstructing trajectory a computationally expensive task. However,
%  due to orthogonality of basis functions and optimal covering of
%  approximation subspace, spectral methods exhibit a faster
%  convergence rate.

%  Other benefits of using orthogonal basis functions become evident
%  when the cost function is quadratic (i.e., when linear-quadratic
%  regulation problem is considered~\cite{fernandes91variational}):
% $$
%  J(\pi) = \int^T_0 \| \pi(t) \|^2 \; {\rm d} t \; .
%  $$
%  In which case, the discretized cost functional also takes quadratic
%  form
%  $$
%  J(\tilde \pi) = \sum^N_{i = 1}  \pi^2_i \; .
%  $$
%  This direct computation of the cost functional eliminates the need
%  to use quadrature formulas and introducing additional numerical
%  errors into computations.

\paragraph*{2) Pseudospectral Methods} 
Numerical integration techniques utilize a discretization of the time interval with an interpolating function between collocation points. 
Pseudospectral approximation schemes build on this technique by additionally representing the interpolating function with a basis.
Typical basis functions interpolating between collocation points are finite subsets of the Legendre or Chebyshev polynomials.
These methods typically have improved convergence rates over basic collocation methods, which is especially true when adaptive methods for selecting collocation points and basis functions are used as in \cite{darby11hpadaptive}. 

\subsubsection*{Indirect Methods}
%Indirect methods take a different point of view on trajectory planning problem. 
Pontryagin's minimum principle \cite{pontryagin1987mathematical}, is a celebrated result from optimal control which provides optimality conditions of a solution to Problem \ref{pr:optimal_trajectory_planning}.
Indirect methods, as the name suggests, solve the problem by finding solutions satisfying these optimality conditions.
These optimality conditions are described as an augmented system of ordinary differential equations (ODEs) governing the states and a set of co-states.
However, this system of ODEs results in a two point boundary value problem and can be difficult to solve numerically.
%
%The problem is then to determine the appropriate initial and final values for the costates so that the constraints on the state trajectory are satisfied.
%
One technique is to vary the free initial conditions of the problem and integrate the system forward in search of the initial conditions which leads to the desired terminal states.
This method is known as the \emph{shooting} method, and a version of this approach has been applied to planning parking maneuvers in \cite{tassa14control}.
The advantage of indirect methods, as in the case of the shooting method, is the reduction in dimensionality of the optimization problem to the dimension of the state space.
%
%However, the lower dimensional problems are often still quite difficult to solve and formulating the optimality conditions becomes more difficult as the driving environment becomes less structured.
%

The topic of variational approaches is very extensive and hence, the above is only a brief description of select approaches. See \cite{betts98survey,polak73historical} for dedicated surveys on this topic. %f the approach.

\subsection{Graph Search Methods}

\SetKwFunction{samplepoints}{sample\textrm{-}points}
\SetKwFunction{neighbors}{neighbors}
\SetKwFunction{colfree}{col\textrm{-}free}
\SetKwFunction{path}{path}

\label{sec:state-space-discretization}

Although useful in many contexts, the applicability of variational methods is limited by their convergence to only local minima. 
In this section, we will discuss the class of methods that attempts to mitigate the problem by performing global search in the discretized version of the path space.
These so-called graph search methods discretize the configuration space $\mathcal{X}$ of the vehicle and represent it in the form of a graph and then search for a minimum cost path on such a graph. 

In this approach, the configuration space is represented as a graph $G=(V,E)$, where $V \subset \mathcal{X}$ is a discrete set of selected configurations called vertices and $E=\{(o_i,d_i,\sigma_i)\}$ is the set of edges, where $o_i \in V$ represents the origin of the edge, $d_i$ represents the destination of the edge and $\sigma_i$ represents the path segments connecting $o_i$ and $d_i$. 
It is assumed that the path segment $\sigma_i$ connects the two vertices:
$
\sigma_i(0) = o_i \textrm{ and } \sigma_i(1) = d_i. 
$
Further, it is assumed that the initial configuration $\mathbf{x}_\mathrm{init}$ is a vertex of the graph. 
The edges are constructed in such a way that the path segments associated with them lie completely in $\mathcal{X}_\mathrm{free}$ and satisfy differential constraints.
As a result, any path on the graph can be converted to a feasible path for the vehicle by concatenating the path segments associated with edges of the path through the graph.  

There is a number of strategies for constructing a graph discretizing the free configuration space of a vehicle. In the following subsections, we discuss three common strategies: Hand-crafted lane graphs, graphs derived from geometric representations and graphs constructed by either control or configuration sampling. 

\subsubsection{Lane Graph} \label{sec:road_lane_graph}
When the path planning problem involves driving on a structured road network, a sufficient graph discretization may consist of edges representing the path that the car should follow within each lane and paths that traverse intersections. 

\begin{figure}[h]
\centering
\includegraphics[width=0.4\textwidth]{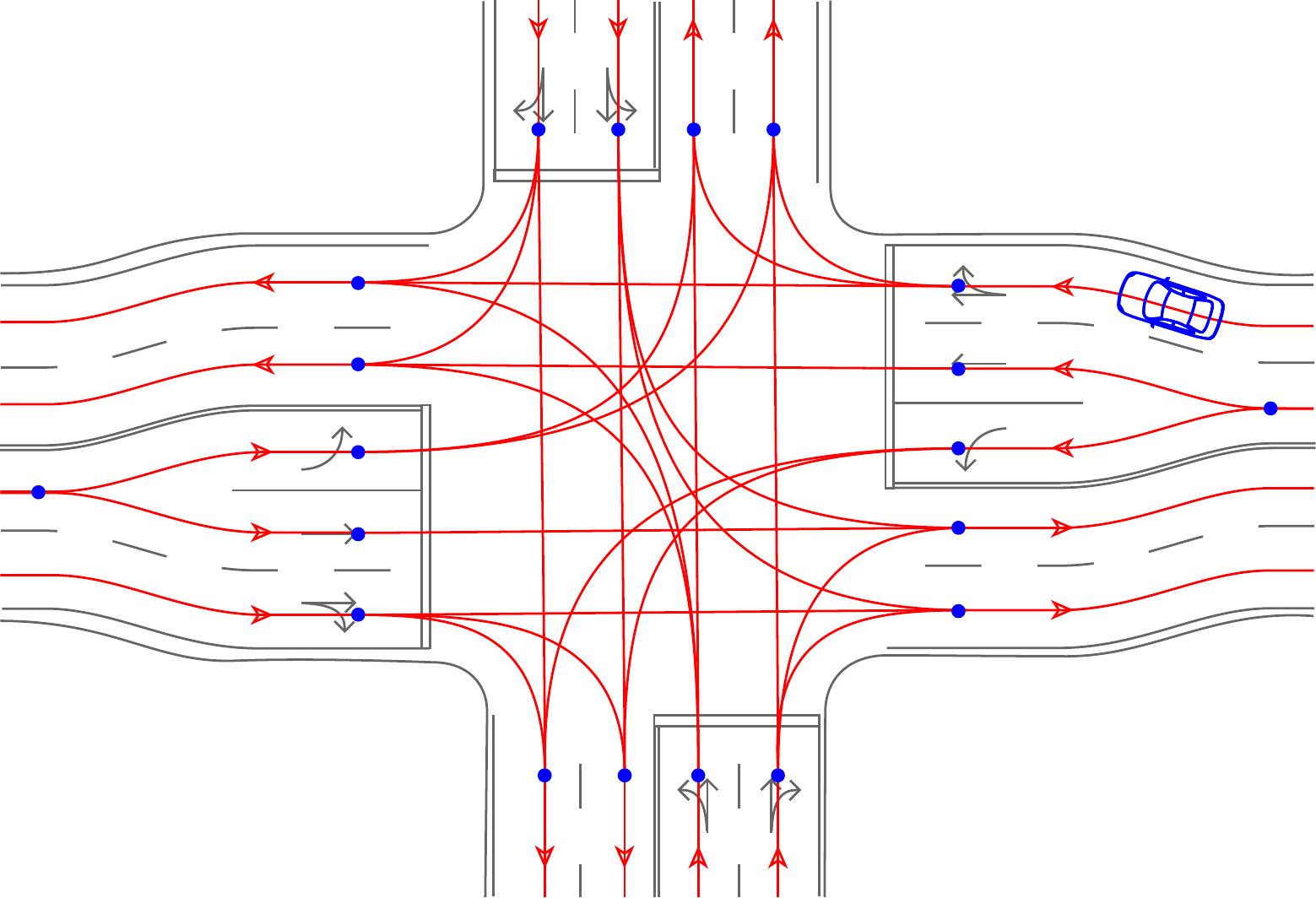}
\caption{Hand-crafted graph representing desired driving paths under normal 
circumstances.} \label{fig:roadnetwork}
\end{figure}

Road lane graphs are often partly algorithmically generated from higher-level street network maps and partly human edited. 
An example of such a graph is in Figure~\ref{fig:roadnetwork}.

Although most of the time it is sufficient for the autonomous vehicle to follow the paths encoded in the road lane graph, occasionally it must be able to navigate around obstacles that were not considered when the road network graph was designed or in environments not covered by the graph. 
Consider for example a faulty vehicle blocking the lane that the vehicle plans to traverse -- in such a situation a more general motion planning approach must be used to find a collision-free path around the detected obstacle.

The general path planning approaches can be broadly divided into two categories based on how they represent the obstacles in the environment. 
So-called geometric or combinatorial methods work with geometric representations of the obstacles, where in practice the obstacles are most commonly described using polygons or polyhedra.
On the other hand, so-called sampling-based methods abstract away from how the obstacles are internally represented and only assumes access to a function that determines if 
any given path segment is in collision with any of the obstacles.  

\subsubsection{Geometric Methods} \label{sec:geometric-models} 
In this section, we will focus on path planning methods that work with geometric representations of obstacles.
We will first concentrate on path planning 
without differential constraints because for this formulation,  efficient exact path planning algorithms exist.
Although not being 
able to enforce differential constraints is limiting for path planning for 
traditionally-steered cars because the constraint on minimum turn radius 
cannot be accounted for, these methods can be useful for obtaining the lower- 
and upper-bounds\footnote{Lower bound is the length of the path without 
curvature constraint, upper bound is the length of a path for a large robot that 
serves as an envelope within which the car can turn in any direction.} on the 
length of a curvature-constrained path and for path planning for more exotic car constructions that can turn on the spot.  

In path planning, the term roadmap is used to describe a graph discretization of $\mathcal{X}_\mathrm{free}$ that 
describes well the connectivity of the free configuration space and has the property that any point in $\mathcal{X}_\mathrm{free}$ is trivially reachable 
from some vertices of the roadmap. 
When the set  $\mathcal{X}_\mathrm{free}$ can be described geometrically using a linear or semi-algebraic model, different types of roadmaps for $\mathcal{X}_\mathrm{free}$ can be algorithmically constructed and subsequently used to obtain complete path planning algorithms. 
Most notably, for $\mathcal{X}_\mathrm{free} \subseteq \mathbb{R}^2$ and polygonal models of 
the configuration space, several efficient algorithms for constructing such roadmaps exists such as the vertical cell 
decomposition~\cite{chazelle1987approximation}, generalized Voronoi diagrams~\cite{odunlaing1985retraction,takahashi1989motion}, and visibility graphs~\cite{latombe2012robot,nilsson1969mobile}. 
For higher dimensional configuration spaces described by a general 
semi-algebraic model, the technique known as cylindrical algebraic 
decomposition can be used to construct a roadmap in the configuration 
space~\cite{lavalle2006planning,schwartz1983piano} leading to complete 
algorithms for a very general class of path planning problems. The fastest of 
this class is an algorithm developed by Canny~\cite{canny1988complexity} that has 
(single) exponential time complexity in the dimension of the configuration 
space. The result is however mostly of a theoretical nature without any known 
implementation to date.

%\subsubsection*{Differential Constraints}
Due to its relevance to path planning for car-like vehicles, a number of results also exist for the problem of path planning with a constraint on maximum curvature. Backer and Kirkpatrick~\cite{backer2007finding} provide an algorithm for constructing a path with bounded curvature that is polynomial in the number of features of the 
domain, the precision of the input and the number of segments on the simplest 
obstacle-free Dubins path connecting the specified configurations.
Since the problem of finding a \emph{shortest} path with bounded curvature 
amidst polygonal obstacles is NP-hard, it is not surprising that no exact 
polynomial solution algorithm is known. An approximation algorithm for finding 
shortest curvature-bounded path amidst polygonal obstacles has been first 
proposed by Jacobs and Canny~\cite{jacobs1993planning} and later improved by 
Wang and Agarwal~\cite{wang1996approximation} with time complexity 
$O(\frac{n^2}{\epsilon^4}\log n)$, where $n$ is the number of vertices of the 
obstacles and $\epsilon$ is the approximation factor. \fullc{For the special case of 
so-called moderate obstacles that are characterized by smooth boundary with 
curvature bounded by $\kappa$, an exact polynomial algorithm for finding a path 
with curvature bounded by at most $\kappa$ have been developed by Boissonnat 
and Lazard~\cite{boissonnat1996polynomial}.}

\subsubsection{Sampling-based Methods}
In autonomous driving, a geometric model of $\mathcal{X}_\mathrm{free}$ is usually not directly available and it would be too costly to construct from raw sensoric data. Moreover, the requirements on the resulting path are often far more complicated than a simple maximum curvature constraint. This may explain the popularity of sampling-based techniques that do not enforce a specific representation of the free configuration set and dynamic constraints. Instead of reasoning over a geometric representation, the sampling based methods explore the reachability of the free configuration space using \emph{steering} and \emph{collision checking} routines: 

The steering function \steer{$\mathbf{x}, \mathbf{y}$} returns a path segment 
starting from configuration $\mathbf{x}$ going towards configuration $\mathbf{y}$ (but not necessarily 
reaching $\mathbf{y}$) ensuring the differential constraints are satisfied, i.e., the resulting motion is feasible for the vehicle model in consideration.
The exact manner in which the steering function is implemented depends on the context in which it is used. Some typical choices encountered in the literature are: 

\begin{enumerate}
\item Random steering: The function returns a path that results from applying a  random 
control input through a forward model of the vehicle from state $\vec{x}$ for either a fixed or variable time step \cite{lavalle1998rapidly}.

\item Heuristic steering: The function returns a path that results from applying control that is heuristically constructed to guide the system from $\mathbf{x}$  towards 
$\mathbf{y}$ \cite{petti2005safe,bhatia2004incremental,glassman2010quadratic}. This includes selecting the maneuver from a pre-designed discrete set (library) of maneuvers.

\item Exact steering: The function returns a feasible path that guides the system from $\mathbf{x}$ to $\mathbf{y}$. Such a path  corresponds to a solution of a 2-point boundary value problem. For some systems and cost functionals, such a path can be obtained analytically, e.g., a straight line for holonomic systems, a Dubins curve for forward-moving unicycle \cite{dubins1957oncurves}, or a Reeds-Shepp curve for 
bi-directional unicycle \cite{reeds1990optimal}. An analytic solution also exists for differentially flat systems \cite{hwan2013optimal}, while for more complicated models, the exact steering can be obtained by solving the two-point boundary value problem. 

\item Optimal exact steering: The function returns an optimal exact steering path with respect to the given cost functional. In fact, the straight line, the Dubins curve, and the Reeds-Shepp curve from the previous point are optimal solutions assuming that the cost functional is the arc-length of the path \cite{dubins1957oncurves,reeds1990optimal}. 
\end{enumerate}

The collision checking function \colfree{$\sigma$} returns true
if path segment $\sigma$ lies entirely in $\mathcal{X}_\mathrm{free}$ and it is used to ensure that the resulting path does not collide with any of the obstacles.

Having access to steering and collision checking functions, the major challenge becomes how to construct a discretization that approximates well the connectivity of $\mathcal{X}_\mathrm{free}$ without having access to an explicit model of its geometry. 
We will now review sampling-based discretization strategies from literature.
%\subsection*{Fixed Discretization}
%In this section, we are going to overview the sampling-based algorithms for discretization of the free configuration space. 
%

A straightforward approach is to choose a set of motion primitives (fixed maneuvers) and generate the search graph by recursively applying them starting from the vehicle's initial configuration $\vec{x}_\mathrm{init}$, e.g., using the method in Algorithm~\ref{alg:recursive-maneuver-roadmap}. 
For path planning without differential constraints, the motion primitives can be simply a set of straight lines with different directions and lengths. 
For a car-like vehicle, such motion primitive might by a set of arcs representing the path the car would follow with different values of steering. 
A variety of techniques can be used for generating motion primitives for driverless vehicles.
A simple approach is to sample a number of control inputs and to simulate forwards in time using a vehicle model to obtain feasible motions.
In the interest of having continuous curvature paths, clothoid segments are also sometimes used \cite{fleury95primitives}.
The motion primitives can be also obtained by recording the motion of a vehicle driven by an expert driver~\cite{velenis07aggressive}.

\begin{figure*}[t]
    \centering
    \begin{subfigure}[t]{9cm}
        \includegraphics[height=6cm]{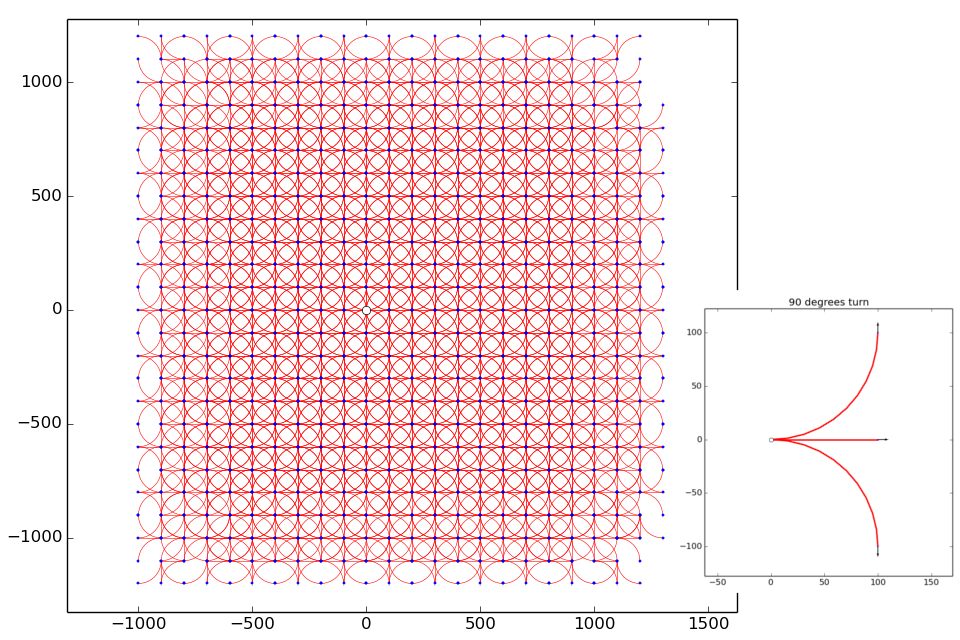}
        \caption{Lattice graph} \label{fig:lattice}
    \end{subfigure}
    \quad   
    \begin{subfigure}[t]{8cm}
        \includegraphics[height=6cm]{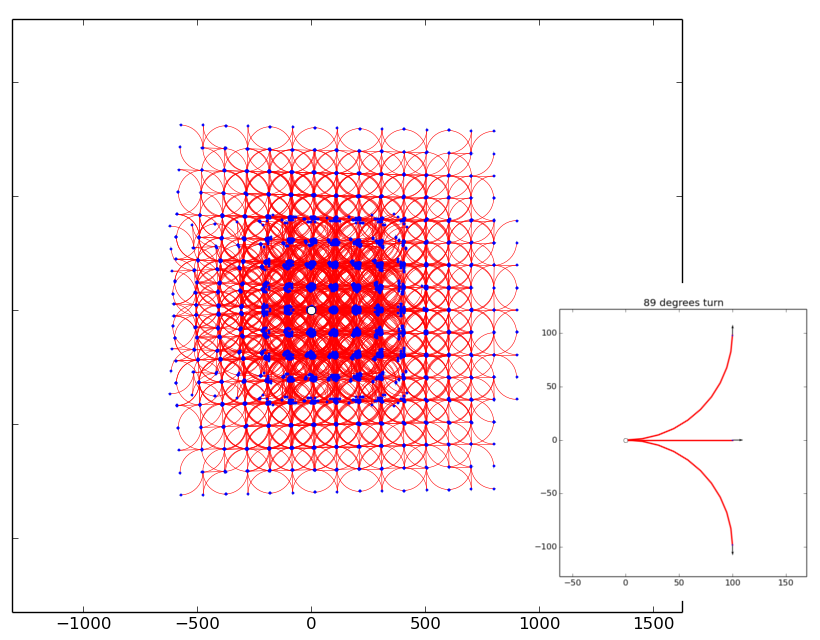}
        \caption{Non-lattice graph } \label{fig:non-lattice}
    \end{subfigure}
\caption{Lattice and non-lattice graph, both with 5000 edges. (a) The graph 
resulting from recursive application of 90\degree~left circular arc, 
90\degree~right circular arc, and a straight line. (b) The graph resulting from 
recursive application of 89\degree~left circular arc, 89\degree~right circular 
arc, and a straight line. The recursive application of those primitives does 
form a tree instead of a lattice with many branches looping in the neighborhood 
of the origin. As a consequence, the area covered by the right graph is smaller. }
\label{fig:lattice-and-non-lattice}
\end{figure*}

Observe that the recursive application of motion primitives may generate a tree graph in which in the worst-case no two edges lead to the same configuration. 
There are, however, sets of motion primitives, referred to as lattice-generating, that result in regular graphs resembling a lattice. 
See Figure~\ref{fig:lattice} for an illustration. 
The advantage of lattice generating primitives is that the vertices of the search graph cover the configuration space uniformly, while trees in general may have a high density of vertices around the root vertex.
Pivtoraiko et al. use the term "state lattice" to describe such graphs in \cite{pivtoraiko2009differentially} and point out that a set of lattice-generating motion primitives for a system in hand can be obtained by first generating regularly spaced configurations around origin and then connecting the origin to such configurations by a path that represents the solution to the two-point boundary value problem between the two configurations.

\begin{algorithm}[h]
$V \leftarrow \{\vec{x}_\mathrm{init}\}$; $E \leftarrow \emptyset$; $Q \leftarrow $ new queue($\vec{x}_\mathrm{init}$)\;
\While {$ Q \neq \emptyset $} {
  $\vec{x} \leftarrow $ pop element from $Q$\;  
  $M \leftarrow$ generate a set of path segments by applying motion primitives 
from configuration  $\vec{x}$\;
  \For{$\sigma \in M$}{
    \If{\colfree{$\sigma$}}{
      $E \leftarrow E \cup \{ (\vec{x}, \sigma(1), \sigma) \}$\;
      \If{$\sigma(1) \not\in V$} {
      add $\sigma(1)$ to $Q$\; 
      $V \leftarrow V \cup \{ \sigma(1) \}$\;
      }  
    }
  }
}
\Return (V,E)

\caption{Recursive Roadmap Construction} \label{alg:recursive-maneuver-roadmap}
\end{algorithm}

%\subsubsection*{Deterministic Sampling}

An effect that is similar to recursive application of lattice-generating motion 
primitives from the initial configuration can be achieved by generating a 
discrete set of samples covering the (free) configuration space and connecting 
them by feasible path segments obtained using an exact steering procedure.

%Such sampling-based approaches require an 
%access to a local path planner able to provide a feasible path segment 
%connecting any two configurations in the absence of obstacles. Let 
%$\steer(\mathbf{x},\mathbf{y})$ in denote such a routine, i.e. a function that 
%returns a path for a vehicle from $\mathbf{x}$ to $\mathbf{y}$ that satisfies 
%all differential constraints. 

Most sampling-based roadmap construction approaches follow the algorithmic 
scheme shown in Algorithm~\ref{alg:sampling-based-roadmap}, but differ in the 
implementation of the \samplepoints{$\mathcal{X},n$} and 
\neighbors{$\vec{x},V$} routines. The function \samplepoints{$\mathcal{X},n$} 
represents the strategy for selecting $n$ points from the configuration space 
$\mathcal{X}$, while the function \neighbors{$\vec{x},V$} represents the strategy 
for selecting a set of neighboring vertices $N \subseteq V$ for a vertex 
$\vec{x}$, which the algorithm will attempt to connect to $\vec{x}$ by a path 
segment using an exact steering function, $\steer_\mathrm{exact}(\vec{x}, \vec{y})$.

\begin{algorithm}[h]
$V \leftarrow \{\vec{x}_\mathrm{init}\} \cup \samplepoints(\mathcal{X},n)$; $E \leftarrow \emptyset$\;

\For{$\vec{x} \in V$}{
   \For{$\vec{y} \in \neighbors(\vec{x},V)$} {
       $\sigma \leftarrow \steer_\mathrm{exact}(\vec{x}, \vec{y})$\;
   
       \If{$\colfree(\sigma$)} {
          $E \leftarrow E \cup \{ (\vec{x}, \vec{y}, \sigma) \}$\;
       }
   }
}
\Return (V,E)
\caption{Sampling-based Roadmap Construction} \label{alg:sampling-based-roadmap}
\end{algorithm}

The two most common implementations of \samplepoints{$\mathcal{X},n$} function 
are 1) return $n$ points arranged in a regular grid and 2) return $n$ randomly 
sampled points from $\mathcal{X}$. 
\shortc{While random sampling has an advantage of 
being generally applicable and easy to implement, so-called Sukharev grids have 
been shown to minimize the radius of the largest empty ball with no sample point inside~\cite{lavalle2004relationship}.}
\fullc{While random sampling has an advantage of 
being generally applicable and easy to implement, so-called Sukharev grids have 
been shown to achieve optimal $L_\infty$-dispersion in unit hypercubes, i.e. 
they minimize the radius of the largest empty ball with no sample point inside. For in  
depth discussion of relative merits of random and deterministic sampling in the 
context of sampling-based path planning, we refer the reader to~\cite{lavalle2004relationship}.}
The two most commonly used strategies 
for implementing \neighbors{$\vec{x},V$} function are to take 1) the set of 
$k$-nearest neighbors to $\vec{x}$ or 2) the set of points lying within the 
ball centered at $\vec{x}$ with radius $r$. 

In particular, samples arranged deterministically in a $d$-dimensional grid with the 
neighborhood taken as $4$ or $8$ nearest neighbors in 2-D or the analogous 
pattern in higher dimensions represents a straightforward deterministic 
discretization of the free configuration space. This is in part because they arise 
naturally from widely used bitmap representations of free and occupied regions 
of robots' configuration space~\cite{lengyel1990real}. 

%\subsubsection*{Random Sampling}

Kavraki et al.~\cite{kavraki1996probabilistic} advocate the use of random sampling within the framework of Probabilistic Roadmaps (PRM) in order to  construct roadmaps in high-dimensional configuration spaces, because unlike grids, they can be naturally run in an anytime fashion. 
The batch version of PRM~\cite{kavraki1998analysis} follows the scheme in Algorithm~\ref{alg:sampling-based-roadmap} with random sampling and neighbors selected within a ball with fixed radius~$r$. 
Due to the general formulation of PRMs, they have been used for path planning for a variety of systems, including systems with differential constraints. 
However, the theoretical analyses of the algorithm have primarily been focused on the performance of the algorithm for systems without differential constraints, i.e. when a straight line is used to connect two configurations. 
Under such an assumption, PRMs have been shown in~\cite{karaman2011sampling} to be probabilistically complete and asymptotically optimal. That is, the probability that the resulting graph contains a valid solution (if it exists) converges to one with increasing size of the graph and the cost of the shortest path in the graph converges to the  optimal cost.
Karaman and Frazzoli~\cite{karaman2011sampling} proposed an adaptation of batch PRM, called PRM*, that instead only connects neighboring vertices in a ball with a logarithmically shrinking radius with increasing number of samples to maintain both asymptotic optimality and computational efficiency. 

In the same paper, the authors propose Rapidly-exploring Random Graphs (RRG*), which is an incremental discretization strategy that can be terminated at  any time while maintaining the asymptotic optimality property. 
Recently, Fast Marching Tree (FMT*)~\cite{janson2015fast} has been proposed as an asymptotically optimal alternative to PRM*. 
The algorithm combines discretization and search into one process by performing a lazy dynamic programming recursion over a set of sampled vertices that can be subsequently used to quickly determine the path from initial configuration to the goal region.

Recently, the theoretical analysis has been extended also to differentially constrained systems. 
Schmerling et al.~\cite{schmerling2015optimal} propose differential versions of PRM* and FMT* and prove asymptotic optimality of the algorithms for driftless control-affine dynamical systems, a class that includes models of non-slipping wheeled vehicles.

%\subsection*{Non-holonomic systems}
%In most cases the kinematic and/or dynamic properties of a vehicle cannot be 
%neglected and will result in differential constraints on the path or trajectory 
%for the vehicle. In this section we will focus on approaches for configuration 
%space discretization under differential constraints. 
%
%Consider the problem of planning for a car modeled as a unicycle with a 
%configuration space $\mathcal{C} \subseteq \mathbb{R}^2 \times \mathbb{S}^1$. A 
%configuration of the car is described by a triplet $(x,y,\theta)$, where $x,y$ 
%describe the position of the car and $\theta$ describe its heading. The no-slip 
%assumption and the limit on steering rate $\omega_\mathrm{max}$ yields 
%constraints 
%
%$$
%\begin{array}{lcr}
%x' &=& \cos \, \theta \\
%y' &=& \sin \, \theta \\
%\frac{\vert \theta' \vert}{\vert (x',y') \vert} &\leq& \omega_\mathrm{max}. \\
%\end{array}
%$$       
%
%

\subsubsection{Graph Search Strategies}
In the previous section, we have discussed techniques for the discretization of the free configuration space in the form of a graph. To obtain an actual optimal path in such a discretization, one must employ one of the graph search algorithms. In this section, we are going to review the graph search algorithms that are relevant for path planning.

The most widely recognized algorithm for finding shortest paths in a 
graph is probably the Dijkstra's algorithm~\cite{dijkstra1959note}. The algorithm performs 
the best first search to build a tree representing shortest paths from a given 
source vertex to all other vertices in the graph. When only a path 
to a single vertex is 
required, a heuristic can be used to guide the search process. The most 
prominent heuristic search algorithm is A* developed by Hart, Nilsson and 
Raphael~\cite{hart1968formal}. If the provided heuristic function is admissible 
(i.e., it never overestimates the cost-to-go), A* has been shown to be optimally 
efficient and is guaranteed to return an optimal solution. For many problems, a 
bounded suboptimal solution can be obtained with less computational effort 
using Weighted A*~\cite{pohl1970first}, which corresponds to simply multiplying 
the heuristic by a constant factor $\epsilon > 1$. It can be shown that the 
solution path returned by A* with such an inflated heuristics is guaranteed to 
be no worse than $(1+\epsilon)$ times the cost of an optimal path. 

Often, the shortest path from the vehicle's current configuration to the goal 
region is sought repeatedly every time the model of the world is updated using 
sensory data. Since each such update usually affects only a minor part 
of the graph, it might be wasteful to run the search every time completely from scratch. 
The family of real-time replanning search algorithms such as 
D*~\cite{stentz1994optimal}, Focussed D*~\cite{stentz1995focussed} and 
D*~Lite~\cite{koenig2005fast} has been designed to efficiently recompute the 
shortest path every time the underlying graph changes, while making use of the 
information from previous search efforts.

Anytime search algorithms attempt to provide a first suboptimal path quickly and 
continually improve the solution with more computational time. Anytime 
A*~\cite{hansen2007anytime} uses a weighted heuristic to find the first solution 
and achieves the anytime behavior by continuing the search with the cost of the 
first path as an upper bound and the admissible heuristic as a lower bound, whereas  
Anytime Repairing A*~(ARA*)~\cite{likhachev2003ara} performs a series of searches 
with inflated heuristic with decreasing weight and reuses information from 
previous iterations. On the other hand, Anytime Dynamic A*~(ADA*)~\cite{likhachev2005adastar} combines ideas behind D*~Lite and ARA* to produce an anytime search algorithm for 
real-time replanning in dynamic environments. 

\begin{full}
A clear limitation of algorithms that search for a path on a graph 
discretization of the configuration space is that the resulting optimal path on 
such graph may be significantly longer than the true shortest path in the 
configuration space. Any-angle path planning 
algorithms~\cite{daniel2010theta,nash2010lazy,yap2011block} are designed to 
operate on grids, or more generally on graphs representing cell decomposition of the free configuration space, and 
try to mitigate this shortcoming by considering "shortcuts" between the 
vertices on the graph during search. In addition, Field D* introduces 
linear-interpolation to the search procedure to produce smooth 
paths~\cite{ferguson2006using}. 
\end{full}

\subsection{Incremental Search Techniques} 
\label{sec:sampling-based-methods}

A disadvantage of the techniques that search over a fixed graph discretization is that they search only over the set of paths that can be constructed from primitives in the graph discretization. Therefore, these techniques may fail to return a feasible path or return a noticeably suboptimal one.
%esides the disadvantage of graph discretization methods that the resulting path may be significantly suboptimal, another important drawback is that these methods can even fail to find a feasible path even though such a path exists. 

The incremental \emph{feasible} motion planners strive to address this problem and provide a feasible 
path to any motion planning problem instance, if one exists, given enough computation 
time.  Typically, 
these methods incrementally build increasingly finer discretization of the 
configuration space while concurrently attempting to determine if a path from 
initial configuration to the goal region exists in the discretization at each 
step. If the instance is  ``easy'', the solution is provided quickly, 
but in general the computation time can be unbounded.
Similarly, incremental \emph{optimal} motion planning approaches on top of 
finding a feasible path fast attempt to provide a sequence of solutions of 
increasing quality that converges to an optimal path. 

The term \emph{probabilistically complete} is used in the literature to describe 
algorithms that find a solution, if one exists, with probability approaching one with increasing computation time. Note that probabilistically complete algorithm may not terminate if the solution does not exist. Similarly, the term \emph{asymptotically optimal} is used for algorithms that converge to optimal solution with probability one. 
%\subsection*{Sequential Search}

A na{\"i}ve strategy for obtaining completeness and optimality in the limit is to 
solve a sequence of path planning problems on a fixed discretization of the configuration space, 
each time with a higher resolution of the discretization. 
One disadvantage of this approach is that the path planning processes on 
individual resolution levels are independent without any information reuse. 
Moreover, it is not obvious how fast  the resolution of the 
discretization should be increased before a new graph search is initiated, i.e., if it is 
more appropriate to add a single new configuration, double the number of 
configuration, or double the number of discrete values along each configuration 
space dimension. To overcome such issues, incremental motion planning methods 
interweave incremental discretization of configuration space with search for a 
path within one integrated process.

An important class of methods for incremental path planning is based on the 
idea of incrementally growing a tree rooted at the initial configuration of the 
vehicle outwards to explore the reachable configuration space. The 
"exploratory" behavior is achieved by iteratively selecting a random vertex 
from the tree and by expanding the selected vertex by applying the steering function from it. Once the tree grows large enough to reach the goal region, the 
resulting path is recovered by tracing the links from the vertex in the goal 
region backwards to the initial configuration. The general algorithmic scheme 
of an incremental tree-based algorithm is described in 
Algorithm~\ref{alg:incremental-tree-based-algs}.

\begin{algorithm}[h]
$V \leftarrow \{\vec{x}_\mathrm{init}\} \cup \samplepoints(\mathcal{X},n)$; $E \leftarrow \emptyset$\;

\While{not interrupted}{
   $\mathbf{x}_\mathrm{selected} \leftarrow $\select{$V$}\;
   $\sigma \leftarrow $\extend{$\mathbf{x}_\mathrm{selected},V$}\;
   \If{\colfree{$\sigma$}}{
       $\mathbf{x}_\mathrm{new} = \sigma(1)$\;
       $V \leftarrow V \cup \{\mathbf{x}_\mathrm{new}\};\:\: E \leftarrow E \cup \{(\mathbf{x}_\mathrm{selected}, 
\mathbf{x}_\mathrm{new}, \sigma)\}$\;
   }
}
\Return (V,E)
\caption{Incremental Tree-based Algorithm} 
\label{alg:incremental-tree-based-algs}
\end{algorithm}

One of the first randomized tree-based incremental planners was the expansive 
spaces tree (EST) planner proposed by Hsu et al.~\cite{hsu1997path}. 
The algorithm selects a vertex for expansion, $\mathbf{x}_\mathrm{selected}$, 
randomly from $V$ with a probability that is inversely proportional to the number of 
vertices in its neighborhood, which promotes growth towards unexplored regions. 
During expansion, the algorithm samples a new vertex $\mathbf{y}$ within a 
neighborhood of a fixed radius around $\mathbf{x}_\mathrm{selected}$, and use the 
same technique for biasing the sampling procedure to select a vertex from the region that is relatively less explored. Then it returns a straight line path between 
$\mathbf{x}_\mathrm{selected}$ and $\mathbf{y}$.
A generalization of the idea for planning with kinodynamic constraints in 
dynamic environments was introduced in \cite{hsu2002randomized}, where the 
capabilities of the algorithm were demonstrated on different non-holonomic 
robotic systems and the authors use an idealized version of the algorithm to 
establish that the probability of failure to find a feasible path depends on 
the expansiveness property of the state space and decays exponentially with 
the number of samples.

Rapidly-exploring Random Trees (RRT) \cite{lavalle1998rapidly} have been 
proposed by La Valle as an efficient method for finding feasible trajectories 
for high-dimensional non-holonomic systems. The rapid exploration is achieved 
by taking a random sample $\mathbf{x}_\mathrm{rnd}$ from the free configuration 
space and extending the tree in the direction of the random sample. In RRT, the 
vertex selection function \select{V} returns the nearest neighbor to the random 
sample $\mathbf{x}_\mathrm{rnd}$ according to the given distance metric between 
the two configurations. The extension function \extend{} then generates a path 
in the configuration space by applying a control for a fixed time step that 
minimizes the distance to $\mathbf{x}_\mathrm{rnd}$. Under certain simplifying 
assumptions (random steering is used for extension), 
the RRT algorithm has been shown to be probabilistic 
complete~\cite{lavalle2001randomized}. We remark that the result on 
probabilistic completeness does not readily generalize to many practically 
implemented versions of RRT that often use heuristic steering. 
In fact, it has been recently 
shown in~\cite{kunz2015kinodynamic} that RRT using heuristic
steering with fixed time step is not probabilistically complete. 

Moreover, Karaman and Frazzoli~\cite{karaman2010optimal} demonstrated that the RRT 
converges to a suboptimal solution with probability one and designed an 
asymptotically optimal adaptation of the RRT algorithm, called RRT*. \fullc{As shown in 
Algorithm~\ref{alg:rrts}, the} \shortc{The} RRT* at every iteration considers a set of 
vertices that lie in the neighborhood of newly added vertex 
$\mathbf{x}_\mathrm{new}$ and a) connects $\mathbf{x}_\mathrm{new}$ to the 
vertex in the neighborhood that minimizes the cost of path from 
$\mathbf{x}_\mathrm{init}$ to $\mathbf{x}_\mathrm{new}$ and b) rewires any 
vertex in the neighborhood to $\mathbf{x}_\mathrm{new}$ if that results in a 
lower cost path from $\mathbf{x}_\mathrm{init}$ to that vertex. 
An important characteristic of the algorithm is that the neighborhood region is 
defined as the ball centered at $\mathbf{x}_\mathrm{new}$ with radius being 
function of the size of the tree: 
$ r = \gamma \sqrt[d]{(\log n)/n},$ 
where $n$ is the number of vertices in the tree, $d$ is the dimension of the 
configuration space, and $\gamma$ is an instance-dependent constant. It is 
shown that for such a function, the expected number of vertices in the ball is 
logarithmic in the size of the tree, which is necessary to ensure that the 
algorithm almost surely converges to an optimal path while maintaining the 
same asymptotic complexity as the suboptimal RRT. 

\begin{full}
\begin{algorithm}[htb!]
$V \leftarrow \{\vec{x}_\mathrm{init}\}$; $E \leftarrow \emptyset$\;

\While{not interrupted}{
   $\mathbf{x}_\mathrm{selected} \leftarrow $\select{$V$}\;
   $\sigma \leftarrow $\extend{$\mathbf{x}_\mathrm{selected},V$}\;
   \If{\colfree{$\sigma$}}{
       $\mathbf{x}_\mathrm{new} = \sigma(1)$ \;
       $V \leftarrow V \cup \{\mathbf{x}_\mathrm{new}\}$\;
       \vspace{5mm}
       \tcp{consider all vertices in ball of radius $r$ around 
$\mathbf{x}_\mathrm{new}$}
       $r = \gamma \sqrt[d]{\frac{\log \vert V \vert }{\vert V \vert}} $\;
       $X_\mathrm{near} \leftarrow \{ \mathbf{x} \in V \setminus
           \{\mathbf{x}_\mathrm{new}\} :
           d(\mathbf{x}_\mathrm{new}, \mathbf{x} \} < r \}$\;    
       \vspace{5mm}       
       \tcp{find best parent}
       \hangindent=.5\skiptext\hangafter=1$\mathbf{x}_\mathrm{par}$ 
$\leftarrow$ $\argmin_{\mathbf{x} \in X_\mathrm{near}}$ $c(\mathbf{x})$ + 
$c(\connect(\mathbf{x},\mathbf{x}_\mathrm{new}))$ subj. to 
\colfree{$\steer_\mathrm{exact}(\mathbf{x},\mathbf{x}_\mathrm{new})$}\;
       $\sigma'=\steer_\mathrm{exact}(\mathbf{x}_\mathrm{par},\mathbf{x}_\mathrm{new})$\;
       $E \leftarrow E \cup \{(\mathbf{x}_\mathrm{par}, 
\mathbf{x}_\mathrm{new}, \sigma')\}$\;
       \vspace{5mm}
       \tcp{rewire vertices in neighborhood}
       \For{$\mathbf{x} \in X_\mathrm{near}$}{
         $\sigma' \leftarrow \steer_\mathrm{exact}(\mathbf{x}_\mathrm{new}, \mathbf{x})$\;
         \If{$c(\mathbf{x}_\mathrm{new})$ + $c(\sigma')$ < $c(\mathbf{x})$ and 
\colfree{$\sigma'$}} {
             \hangindent=.5\skiptext\hangafter=1 $E \leftarrow $ $ (E \setminus 
\{(p(\mathbf{x}), \mathbf{x},\sigma'')\})$ $\cup$ $\{(\mathbf{x}_\mathrm{new}, 
\mathbf{x}, \sigma')\}$, where $\sigma''$ is the path from $p(\mathbf{x})$ to 
$\mathbf{x}$\;             
         }
       }       
     }
}
\Return (V,E)
\caption{RRT* Algorithm. The cost-to-come to vertex $\mathbf{x}$ is denoted as 
$c(\mathbf{x})$, the cost of path segment $\sigma$ is denoted as $c(\sigma)$ 
and the parent vertex of vertex $\mathbf{x}$ is denoted by $p(\mathbf{x})$.} 
\label{alg:rrts}
\end{algorithm}
\end{full}

Sufficient conditions for asymptotic optimality of RRT* under differential 
constraints are stated in~\cite{karaman2011sampling} and demonstrated to be 
satisfiable for Dubins vehicle and double integrator systems. In a later work, 
the authors further show in the context of small-time locally attainable 
systems that the algorithm can be adapted to maintain not only asymptotic optimality, 
but also computational efficiency~\cite{karaman2013rrtsnonholonom}. Other related works 
focus on deriving distance and steering functions for non-holonomic systems by 
locally linearizing the system dynamics~\cite{perez2012lqr} or by deriving a 
closed-form solution for systems with linear dynamics~\cite{webb2013kinodynamic}. 
On the other hand, RRT\textsuperscript{X} is an algorithm that extends RRT$^{*}$ to allow for real-time incremental replanning when the obstacle region changes, e.g., in the face of new data from sensors~\cite{otte2014rrtxwafr}.

New developments in the field of sampling-based algorithms include algorithms 
that achieve asymptotic optimality without having access to an exact steering 
procedure. In particular, Li at al.~\cite{li2015sparse} recently proposed the Stable Sparse Tree (SST) 
method for asymptotically (near-)optimal path planning, which is based on building a tree of randomly sampled controls propagated through a forward model of the dynamics of the system such that the locally suboptimal branches are pruned out to ensure that the tree remains sparse.

\subsection{Practical Deployments}
Three categories of path planning methodologies have been discussed for self-driving vehicles: variational methods, graph-searched methods and incremental tree-based methods. 
The actual field-deployed algorithms on self-driving systems come from all the categories described above. 
For example, even among the first four successful participants of DARPA Urban Challenge, the approaches used for motion planning significantly differed. 
The winner of the challenge, CMU's Boss vehicle used variational techniques for local trajectory generation in structured environments and a lattice graph in 4-dimensional configuration space (consisting of position, orientation, and velocity) together with Anytime D* to find a collision-free paths in parking lots~\cite{urmson2008autonomous}. 
The runner-up vehicle developed by Stanford's team reportedly used a search strategy coined Hybrid A* that during search, lazily constructs a tree of motion primitives by recursively applying a finite set of maneuvers. 
The search is guided by a carefully designed heuristic and the sparsity of the tree is ensured by only keeping a single node within a given region of the configuration space~\cite{dolgov2008practical}. 
Similarly, the vehicle arriving third developed by the  VictorTango team from Virginia Tech constructs a graph discretization of possible maneuvers and searches the graph with the A* algorithm~\cite{bacha2008odin}. 
Finally, the vehicle developed by MIT used a variant of RRT algorithm called closed-loop RRT with biased sampling~\cite{leonard2008perception}.

%\begin{table*}[!htbp]
%setlength{\tabcolsep}{3.5pt}
%begin{minipage}[c]{1\textwidth}%
%begin{center}
%begin{tabular}{lll}
%topline \headcol & Path tracking controllers & Trajectory tracking controllers \tabularnewline
%midline Kinematic  & \tabitem Pure pursuit \cite{wallace1985first,amidi1991integrated,coulter1992implementation}  & \tabitem Output feedback linearization \cite{andrea1995control} \tabularnewline
%odels  & \tabitem Rear wheel based feedback \cite{samson1992path}  & \tabitem Trajectory stabilization of unicycle model \cite{kanayama1990stable,jiangdagger1997tracking} \tabularnewline
%& \tabitem Front wheel based feedback \cite{ventures2006stanley,buehler20072005}  & \tabitem Unconstrained predictive control \cite{ollero1991predictive,raffo2009predictive}\tabularnewline
%% &&\tabitem Unconstrained time-varying LQR \cite{raffo2009predictive}\\
%rowcol Dynamic  & \tabitem Center of mass based predictive control   & \tabitem Nonlinear model predictive control \cite{falcone2007predictive} \tabularnewline
%\rowcol models  & \quad \cite{kim2014model}  & \tabitem Linear time-varying model predictive control \cite{falcone2007predictive,falcone2007linear,falcone2008linear}\tabularnewline
%\bottomlinec  &  & \tabularnewline
%\end{tabular}\caption{Classification of vehicle controllers based on the control problem addressed and model used.
%\label{table:control_classification}}%
%
%\par\end{center}%
%\end{minipage}
%\end{table*}

\begin{table*}[!htbp]
\setlength{\tabcolsep}{7.5pt}
\begin{minipage}[c]{1\textwidth}%
\begin{center}
\begin{tabular}{|l|c|c|c|c|c|}
\hline \cellcolor{dark_grey} Controller & \cellcolor{dark_grey} & \cellcolor{dark_grey} Model & \cellcolor{dark_grey} Stability & \cellcolor{dark_grey} Time Complexity & \cellcolor{dark_grey} Comments/Assumptions\tabularnewline 
\hline \hline
Pure Pursuit & (\ref{sub:Pure-Pursuit}) & Kinematic & \begin{tabular}{@{}l@{}} LES$^{\star}$ to \\ ref. path \end{tabular} & \begin{tabular}{@{}l@{}} $O(n)^{\ast}$ \end{tabular} & \begin{tabular}{@{}l@{}} No path  curvature \end{tabular} \tabularnewline
\hline 
\begin{tabular}{@{}l@{}} Rear wheel \\ based feedback \end{tabular} & (\ref{sub:Rear-wheel-based})& Kinematic & \begin{tabular}{@{}c@{}} LES$^{\star}$ to \\ ref. path \end{tabular} & \begin{tabular}{@{}l@{}} $O(n)^{\ast}$ \end{tabular} & \begin{tabular}{@{}c@{}} $C^{2}(\mathbb{R}^n)$ ref.  paths \end{tabular} \tabularnewline
\hline 
\begin{tabular}{@{}l@{}} Front wheel \\ based feedback \end{tabular} & (\ref{sub:Front-wheel-based}) & Kinematic & \begin{tabular}{@{}c@{}} LES$^{\star}$ to \\ ref. path \end{tabular}  & \begin{tabular}{@{}l@{}} $O(n)^{\ast}$ \end{tabular} & \begin{tabular}{@{}l@{}} $C^{1}(\mathbb{R}^n )$ ref. paths; \\ Forward driving only \end{tabular} \tabularnewline
\hline 
\begin{tabular}{@{}l@{}} Feedback  \\ linearization \end{tabular} & (\ref{output_feedback_linearization}) & \begin{tabular}{@{}c@{}} Steering rate \\ controlled kinematic \end{tabular} & \begin{tabular}{@{}c@{}} LES$^{\star}$ \\ to ref. traj. \end{tabular} & $O(1)$ & \begin{tabular}{@{}c@{}} $C^{1}(\mathbb{R}^n)$ ref. traj.;\\ Forward driving only \end{tabular} \tabularnewline
\hline 
\begin{tabular}{@{}l@{}} Control Lyapunov \\ design \end{tabular} & (\ref{control_lyapunov}) &  Kinematic & \begin{tabular}{@{}c@{}} LES$^{\star}$ to \\ ref. traj. \end{tabular} &  O(1) & \begin{tabular}{@{}c@{}} Stable for constant path \\  curvature and velocity \end{tabular} \tabularnewline
\hline 
Linear MPC & (\ref{sec:predictive}) & \begin{tabular}{@{}c@{}} $C^{1}(\mathbb{R}^n \times \mathbb{R}^m)$ \\ model$^{\sharp}$ \end{tabular} & \begin{tabular}{@{}l@{}} LES$^{\star}$ to ref. \\ or path \end{tabular} & \begin{tabular}{@{}l@{}} $O\left(\sqrt{N}\ln{\left( \frac{N}{\varepsilon} \right) } \right)^{\dagger}$ \end{tabular} & \begin{tabular}{@{}c@{}} Stability depends \\ on horizon length \end{tabular} \tabularnewline
\hline 
Nonlinear MPC & (\ref{sec:predictive}) & \begin{tabular}{@{}c@{}} $C^{1}(\mathbb{R}^n \times \mathbb{R}^m)$ \\ model$^{\sharp}$ \end{tabular} & Not guaranteed & \begin{tabular}{@{}l@{}} $O(\frac{1}{\varepsilon})\,^{\ddagger}$ \end{tabular} & \begin{tabular}{@{}c@{}}  Works well in practice \end{tabular} \tabularnewline
\hline 
\end{tabular}\caption{Overview of controllers discussed within this section. {\bf Legend:}
$\star$: local exponential stability (LES);
$\ast$: assuming \eqref{bad_equation} is evaluated by a linear search over an $n$-point discretization of the path or trajectory;
$\dagger$: assuming the use of an interior-point method to solve (\ref{eq:linear_mpc}) with a time horizon of $n$ and solution accuracy of $\varepsilon$;
$\ddagger$: based on asymptotic convergence rate to local minimum of (\ref{eq:MPC}) using steepest descent. Not guaranteed to return solution or find global minimum.;
$\sharp$: vector field over the state space $\mathbb{R}^n$ defined by each input in $\mathbb{R}^m$ is a continuously differentiable function so that the gradient of the cost or linearization about the reference is defined.
\label{control_table} } 
\par\end{center}
\end{minipage}
\end{table*}

\section{Vehicle Control} \label{section_control}

Solutions to Problem \ref{pr:optimal_path_planning} or \ref{pr:optimal_trajectory_planning} are provided by the motion planning process. 
%
%The model of vehicle mobility used by the motion planner and various manifestations of uncertainty lead to tracking error of the reference path or trajectory. 
%
The role of the feedback controller is to stabilize to the reference path or trajectory in the presence of modeling error and other forms of uncertainty.
Depending on the reference provided by the motion planner, the control objective may be \textit{path stabilization} or \textit{trajectory stabilization}. 
More formally, the path stabilization problem is stated as follows:
\begin{problem}
\label{ctrl_prb1}(Path stabilization) Given a controlled differential
equation $\dot{x}=f(x,u)$, reference path $x_{ref}:\mathbb{R}\rightarrow\mathbb{R}^n$, and velocity $v_{ref}:\mathbb{R}\rightarrow\mathbb{R}$, find a feedback law, $u(x)$, such that solutions to $\dot{x}=f(x,u(x))$
satisfy the following: $\forall\varepsilon>0$ and $t_{1}<t_{2}$,
there exists a $\delta>0$ and a differentiable $s:\mathbb{R}\rightarrow\mathbb{R}$
such that
\begin{enumerate}
\item $\quad\,\,\left\Vert x(t_{1})-x_{ref}(s(t_{1}))\right\Vert \leq\delta \\ \Rightarrow\left\Vert x(t_{2})-x_{ref}(s(t_{2}))\right\Vert \leq\varepsilon$
\item $\lim_{t\rightarrow\infty}\left\Vert x(t)-x_{ref}(s(t))\right\Vert =0$
\item $\lim_{t\rightarrow\infty}\dot{s}(t)=v_{ref}(s(t))$.
\end{enumerate}
\end{problem}
Qualitatively, these conditions are that (1) a small initial tracking error will remain small, (2) the tracking error must converge to zero, and (3) progress along the reference path tends to a nominal rate.  

\begin{figure}[!htb]
\centering{}\includegraphics[width=8cm]{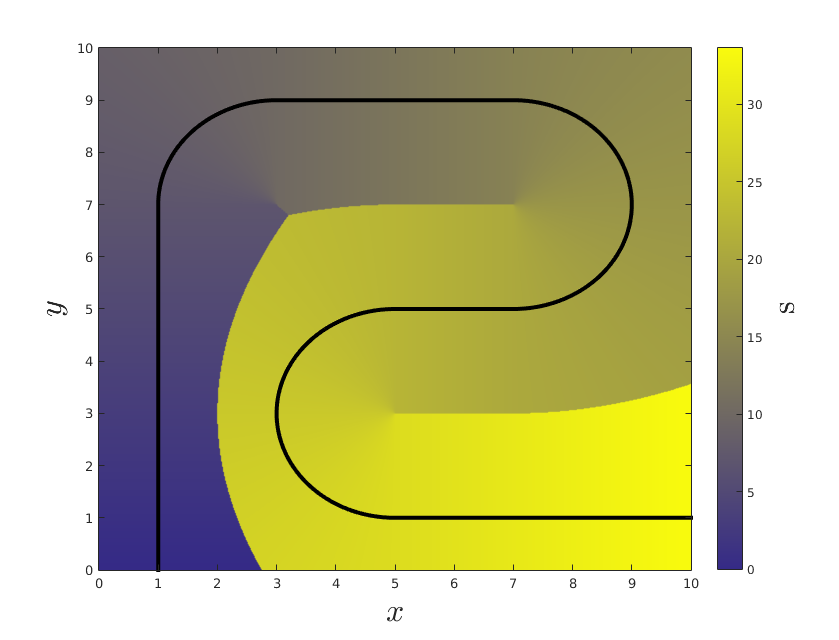}\caption{Visualization of  \eqref{eq:near_point} for a sample reference path shown in black. The color indicates the value of $s$ for each point in the plane and illustrates discontinuities in \eqref{eq:near_point}.  \label{fig:discontinuity}}
\end{figure}

Many of the proposed vehicle control laws, including several discussed in this section, use a feedback law of the form 
\begin{equation}\label{bad_equation}
u(x)=f\left(\underset{\gamma}{{\rm argmin}}\Vert x-x_{ref}(\gamma)\Vert\right),
\end{equation}
where the feedback is a function of the nearest point on the reference path. 
An important issue with controls of this form is that the closed loop vector field $f(x,u(x))$ will not be continuous. 
If the path is self intersecting or not differentiable at some point, a discontinuity in which $f(x,u(x))$ is will lie directly on the path. 
This leads to unpredictable behavior if the executed trajectory encounters the discontinuity.
This discontinuity is illustrated in Figure \ref{fig:discontinuity}.
A backstepping control design which does not use a feedback law of the form \eqref{bad_equation} is presented in \cite{hespanha2007trajectory}. 

The trajectory stabilization problem is more straightforward, but these controllers are prone to performance limitations \cite{aguiar2005path}.
\begin{problem}
\label{cntrl_prb2}(Trajectory stabilization) Given a controlled differential
equation $\dot{x}=f(x,u)$ and a reference trajectory $x_{ref}(t)$,
find $\pi(x)$ such that solutions to $\dot{x}=f(x,\pi(x))$ satisfy
the following: $\forall\varepsilon>0$ and $t_{1}<t_{2}$, there exists
a $\delta>0$ such that 
\begin{enumerate}
\item $\left\Vert x(t_{1})-x_{ref}(t_{1})\right\Vert \leq\delta\Rightarrow\left\Vert x(t_{2})-x_{ref}(t_{2})\right\Vert \leq\varepsilon$
\item $\lim_{t\rightarrow\infty}\left\Vert x(t)-x_{ref}(t)\right\Vert =0$
\end{enumerate}
\end{problem}
In many cases, analyzing the stability of trajectories can be reduced to determining the origin's stability in a time varying system. The basic form of Lyapunov's theorem is only applicable to time invariant systems. However, stability theory for time varying systems is also well established (e.g. \cite[Theorem~4.9]{khalil1996nonlinear}).

Some useful qualifiers for various types of stability include:
\begin{itemize}
\item
\textit{Uniform asymptotic stability} for a time varying system which asserts that $\delta$ in condition 1 of the above problem is independent of $t_1$.
\item
 \textit{Exponential stability} asserts that the rate of convergence is bounded above by an exponential decay. 
\end{itemize}
A delicate issue that should be noted is that controller specifications are usually expressed in terms of the asymptotic tracking error as time tends to infinity. 
In practice, reference trajectories are finite so there should also be consideration for the transient response of the system.

The remainder of this section is devoted to a survey of select control designs which are applicable to driverless cars. 
An overview of these controllers is provided in Table \ref{control_table}. 
Subsection \ref{sec:non-predictive} details a number of effective control strategies for path stabilization of the kinematic model, and subsection \ref{output_feedback_linearization} discusses trajectory stabilization techniques. 
Predictive control strategies, discussed in subsection \ref{sec:predictive}, are effective for more complex vehicle models and can be applied to path and trajectory stabilization.

\subsection{Path Stabilization for the Kinematic Model}\label{sec:non-predictive}
\subsubsection{Pure Pursuit\label{sub:Pure-Pursuit}}

\begin{figure}[!htb]
\centering{}\includegraphics[width=8cm]{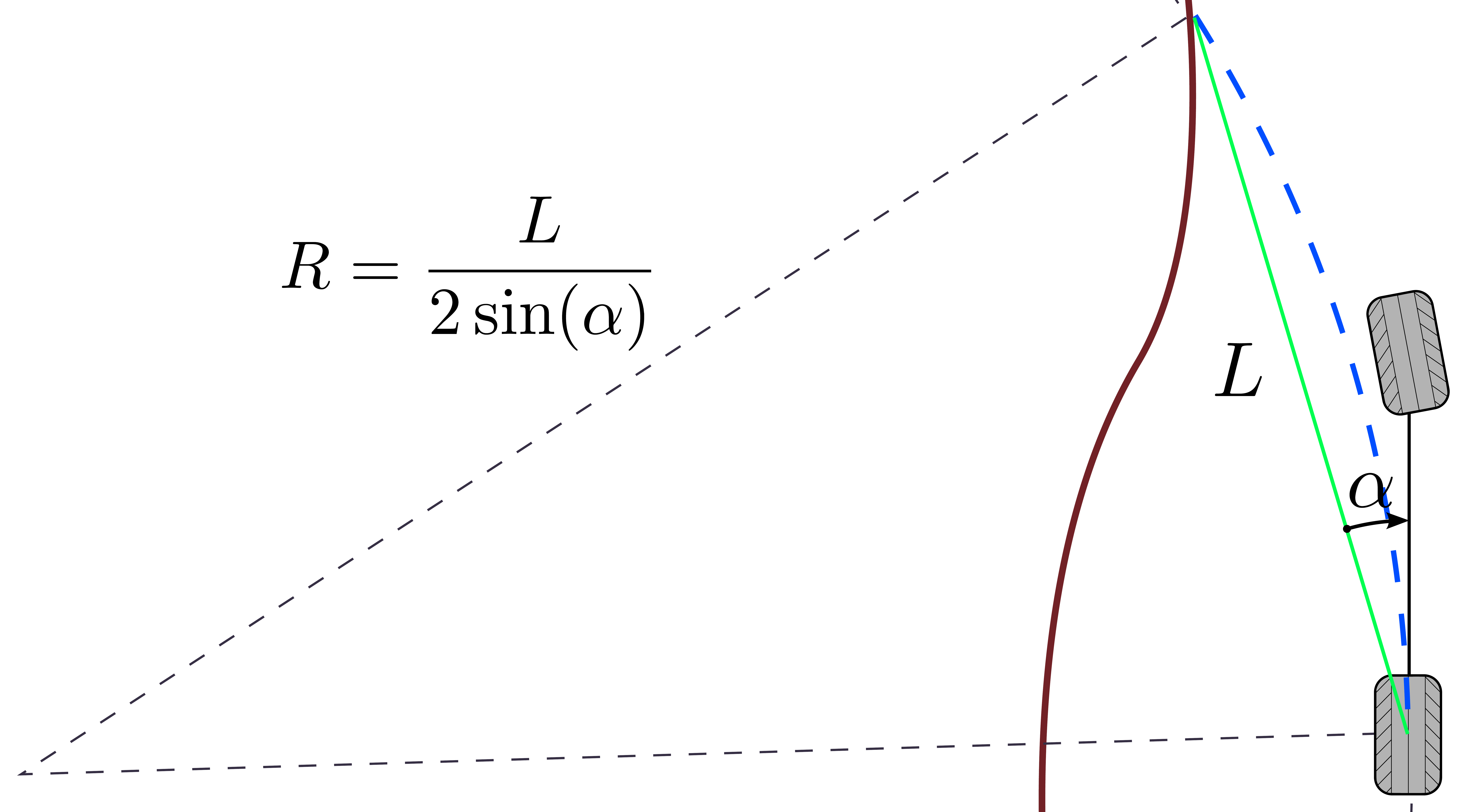}\caption{Geometry of the pure pursuit controller. A circle (blue) is fit between
rear wheel position and the reference path (brown) such that
the chord length (green) is the look ahead distance $L$ and the circle
is tangent to the current heading direction.\label{fig:Geometry-of-the}}
\end{figure}

Among the earliest proposed path tracking strategies is pure pursuit. 
The first discussion appeared in \cite{wallace1985first}, and was elaborated upon in \cite{amidi1991integrated,coulter1992implementation}.
This strategy and its variations (e.g. \cite{rankin1996autonomous,wit2004autonomous}) have proven to be an indispensable tool for vehicle control owing to its simple implementation and satisfactory performance. 
Numerous publications including two vehicles in the DARPA Grand Challenge \cite{buehler20072005} and three vehicles in the DARPA Urban challenge \cite{buehler2009darpa} reported using the pure pursuit controller.
The control law is based on fitting a semi-circle through the vehicle's current configuration to a point on the reference path ahead of the
vehicle by a distance $L$ called the lookahead distance. 
Figure \ref{fig:Geometry-of-the} illustrates the geometry. 
The circle is defined as passing through the position of the car and the point on the path ahead of the car by one lookahead distance with the circle tangent to the car's heading. 
The curvature of the circle is given by 
\begin{equation}
\kappa=\frac{2\sin(\alpha)}{L}.\label{eq:pursuit_curvature}
\end{equation}
For a vehicle speed $v_{r}$, the commanded heading rate is 
\begin{equation}
\omega=\frac{2v_{r}\sin(\alpha)}{L}.\label{eq:pure_pursuit_control}
\end{equation}
In the original publication of this controller \cite{wallace1985first}, the angle $\alpha$ is computed directly from camera output data.
However, $\alpha$ can be expressed in terms of the inertial coordinate system to define a state feedback control. 
Consider the configuration $(x_{r},y_{r},\theta)^{T}$ and the points on the path, $(x_{ref}(s),y_{ref}(s))$, such that $\Vert(x_{ref}(s),y_{ref}(s))-(x_r,y_r)\Vert=L$. 
Since there is generally more than one such point on the reference, take the one with the greatest value of the parameter $s$ to uniquely define a control. 
Then $\alpha$ is given by 
\begin{equation}
\alpha=\arctan\left(\frac{y_{ref}-y_{r}}{x_{ref}-x_{r}}\right)-\theta.\label{eq:pursuit_feedback}
\end{equation}
Assuming that the path has no curvature and the vehicle speed is constant (potentially negative), the pure pursuit controller solves Problem \ref{ctrl_prb1}. 
For a fixed nonzero curvature, pure pursuit has a small steady state tracking error. 
In the case where the vehicle's distance to the path is greater than $L$, the controller output is not defined. 
Another consideration is that changes in reference path curvature can lead to the car deviating from the reference trajectory. 
This may be acceptable for driving along a road, but can be problematic for tracking parking maneuvers. 
Lastly, the heading rate command $\omega$ becomes increasingly sensitive to the feedback angle $\alpha$ as the vehicle speed increases. 
A common fix for this issue is to scale $L$ with the vehicle speed. 

\subsubsection{\label{sub:Rear-wheel-based}Rear wheel position based feedback}

\begin{figure}[!htb]
\centering{}\includegraphics[width=5cm]{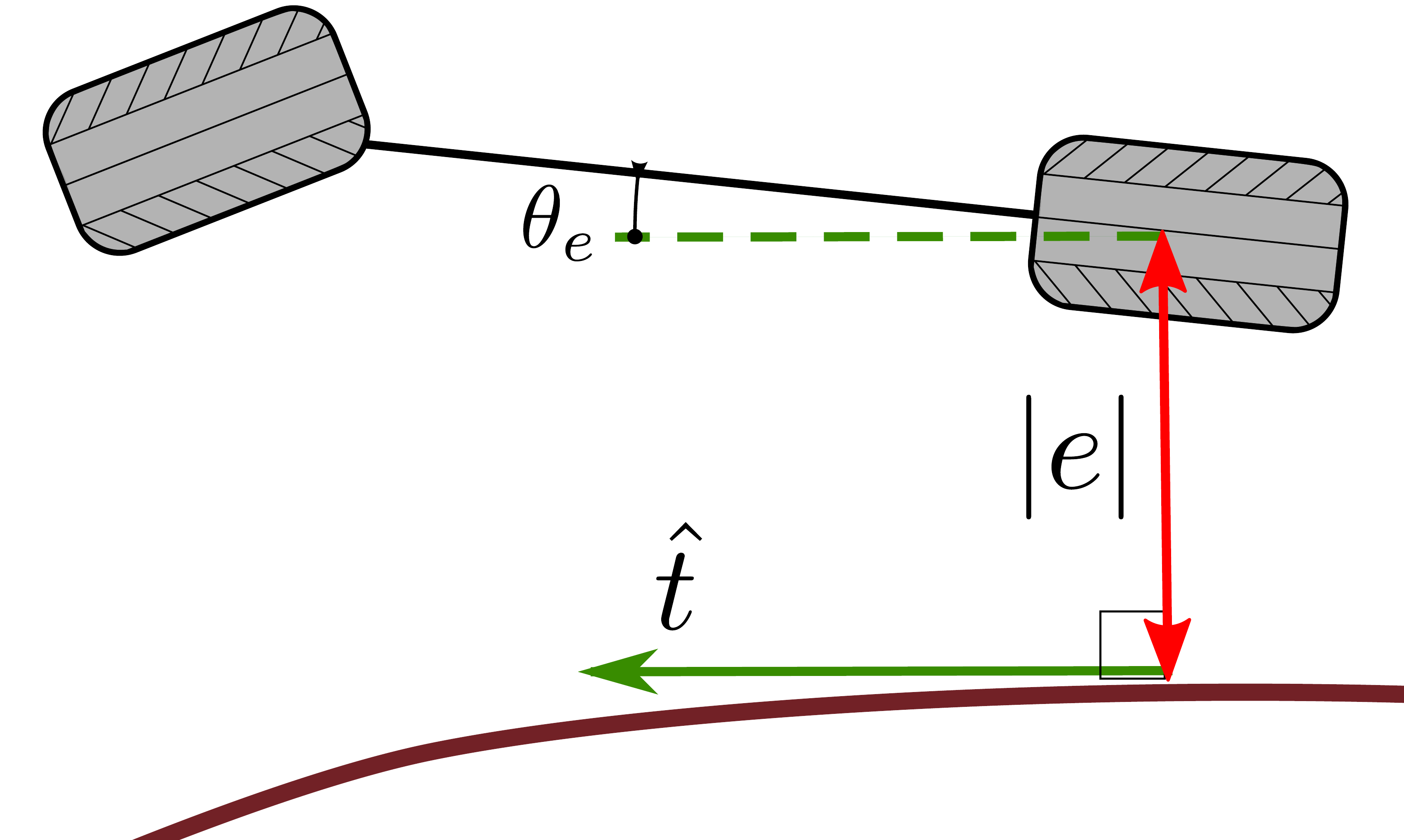}\caption{Feedback variables for the rear wheel based feedback control. $\theta_{e}$
is the difference between the tangent at the nearest point on the
path to the rear wheel and the car heading. 
The magnitude of the scalar value $e$ is illustrated in red. 
As illustrated $e>0$, and for the case where the car is to the left of the path, $e<0$. \label{fig:rear_wheel_control}}
\end{figure}

The next approach uses the rear wheel position as an output to stabilize a nominal rear wheel path \cite{samson1992path}.
The controller assigns 
\begin{equation}
s(t)=\underset{\gamma}{{\rm argmin}}\Vert(x_{r}(t),y_{r}(t))-(x_{ref}(\gamma),y_{ref}(\gamma))\Vert.\label{eq:near_point}
\end{equation}
Detailed assumptions on the reference path and a finite domain containing the reference path where \eqref{eq:near_point} is a continuous function are described in \cite{samson1992path}. The unit tangent to the path at $s(t)$ is given
by 
\begin{equation}
\hat{t}=\frac{\left(\left.\frac{\partial x_{ref}}{\partial s}\right|_{s(t)},\left.\frac{\partial y_{ref}}{\partial s}\right|_{s(t)}\right)}{\left\Vert \left(\frac{\partial x_{ref}(s(t))}{\partial s},\frac{\partial y_{ref}(s(t))}{\partial s}\right)\right\Vert },\label{eq:unit_tangent}
\end{equation}
and the tracking error vector is 
\begin{equation}
d(t):=\left(x_{r}(t),y_{r}(t)\right)-\left(x_{ref}(s(t)),y_{ref}(s(t))\right).\label{eq:error_vector}
\end{equation}
These values are used to compute a transverse error coordinate from the path $e$ which is a cross product between the two vectors 
\begin{equation}
e=d_{x}\hat{t}_{y}-d_{y}\hat{t}_{x}\label{eq:error_coordinate}
\end{equation}
with the subscript denoting the component indices of the vector. 
The control uses the angle $\theta_{e}$ between the vehicle's heading vector and the tangent vector to the path. 
\begin{equation}
\theta_{e}(t)=\theta-\arctan_{2}\left(\frac{\partial y_{ref}(s(t))}{\partial s},\frac{\partial x_{ref}(s(t))}{\partial s}\right).\label{eq:heading_error}
\end{equation}
The geometry is illustrated in Figure \ref{fig:rear_wheel_control}.
A change of coordinates to $(s,e,\theta_{e})$ yields 
\begin{equation}
\begin{array}{ccc}
\dot{s} & = & \frac{v_{r}\cos(\theta_{e})}{1-\kappa(s)e},\\
\dot{e} & = & v_{r}\sin(\theta_{e}),\\
\dot{\theta}_{e} & = & \omega-\frac{v_{r}\kappa(s)\cos(\theta_{e})}{1-\kappa(s)e},
\end{array}\label{eq:curvilinear_coord}
\end{equation}
where $\kappa(s)$ denotes the curvature of the path at $s$. 
The following heading rate command provides local asymptotic convergence to twice continuously differentiable paths: 
\begin{equation}
\omega=\frac{v_{r}\kappa(s)\cos(\theta_{e})}{1-\kappa(s)e}-g_{1}(e,\theta_{e},t)\theta_{e}-k_{2}v_{r}\frac{\sin(\theta_{e})}{\theta_{e}}e,\label{eq:rear_wheel_feedback}
\end{equation}
with $g_{1}(e,\theta_{e},t)>0$, $k_{2}>0$, and $v_{r}\neq0$ which
is verified with the Lyapunov function $V(e,\theta_{e})=e^{2}+\theta_{e}^{2}/k_{2}$
in \cite{samson1992path} using the coordinate system \eqref{eq:curvilinear_coord}.
The requirement that the path be twice differentiable comes from the appearance of the curvature in the feedback law.
An advantage of this control law is that stability is unaffected by the sign of $v_{r}$ making it suitable for reverse driving.

Setting $g_{1}(v_{r},\theta_{e},t)=k_{\theta}|v_{r}|$ for $k_{e}>0$ leads to local exponential convergence with a rate independent of the vehicle speed so long as $v_{r}\neq0$. 
The control law in this case is 
\begin{equation}
\omega=\frac{v_{r}\kappa(s)\cos(\theta_{e})}{1-\kappa(s)e}-\left(k_{\theta}|v_{r}|\right)\theta_{e}-\left(k_{e}v_{r}\frac{\sin(\theta_{e})}{\theta_{e}}\right)e.\label{eq:rear_wheel_feedback_exponential_stable}
\end{equation}
%This resolves, to some extent, the issue that arises with the pure
%pursuit control where the heading rate command for a particular configuration
%is proportional to velocity. Further, unlike pure pursuit, the car
%converges to paths with varying curvature.

\subsubsection{Front wheel position based feedback\label{sub:Front-wheel-based}}

This approach was proposed and used in Stanford University's entry to the 2005 DARPA Grand Challenge \cite{ventures2006stanley},\cite{buehler20072005}.
The approach is to take the front wheel position as the regulated variable. 
%
%The basic form of the controller can be derived from intuitive geometric considerations. 
%
The control uses the variables $s(t),$ $e(t),$ and $\theta_{e}(t)$ as in the previous subsections, with the modification that $e(t)$ is computed with the front wheel position as opposed to the rear wheel position. 
Taking the time derivative of the transverse error reveals 
\begin{equation}
\dot{e}=v_{f}\sin\left(\theta_{e}+\delta\right)\label{eq:error_dot}
\end{equation}
The error rate in \eqref{eq:error_dot} can be directly controlled by the steering angle for error rates with magnitude less than $v_{f}$. 
Solving for the steering angle such that $\dot{e}=-ke$ drives $e(t)$ to zero exponentially fast. 
\begin{equation}
\begin{array}{crcl}
 & v_{f}\sin\left(\delta+\theta_{e}\right) & = & -ke\\
\Rightarrow & \delta & = & \arcsin(-ke/v_{f})-\theta_{e}.
\end{array}\label{eq:stanley_control}
\end{equation}
The term $\theta_{e}$ in this case is not interpreted as heading
error since it will be nonzero even with perfect tracking. It is more
appropriately interpreted as a combination of a feed-forward term of
the nominal steering angle to trace out the reference path and a heading error term. %combined with the heading error.

\begin{figure}[h]
\centering{}\includegraphics[width=7cm]{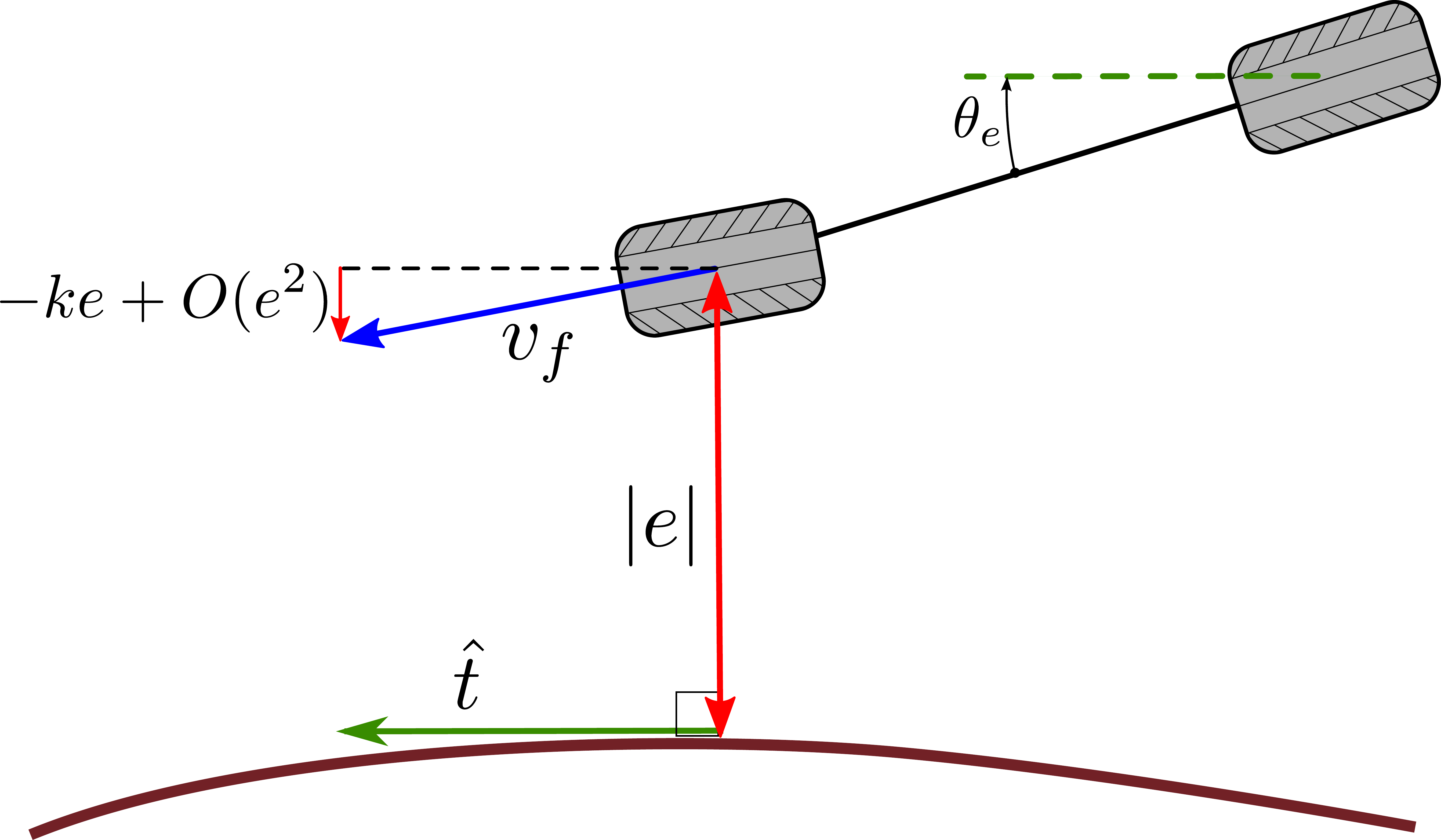}\caption{Front wheel output based control. The control strategy is to point
the front wheel towards the path so that the component of the front
wheel's velocity normal to the path is proportional to the distance
to the path. This is achieved locally and yields local exponential
convergence.\label{fig:Front-wheel-output}}
\end{figure}

The drawback to this control law is that it is not defined when $|ke/v_{f}|> 1$. 
The exponential convergence over a finite domain can be relaxed to local exponential convergence with the feedback law 
\begin{equation}
\delta=\arctan(-ke/v_{f})-\theta_{e},\label{eq:stanley_control_law}
\end{equation}
which, to first order in $e$, is identical to the previous equation.
This is illustrated in Figure \ref{fig:Front-wheel-output}.

Like the control law in \eqref{eq:stanley_control} this controller locally exponentially stabilizes the car to paths with varying curvature with the condition that the path is continuously differentiable. 
The condition on the path arises from the definition of $\theta_{e}$ in the feedback policy.
A drawback to this controller is that it is not stable in reverse making it unsuitable for parking.

\begin{figure}[!htb]
\centering{}\includegraphics[width=\columnwidth]{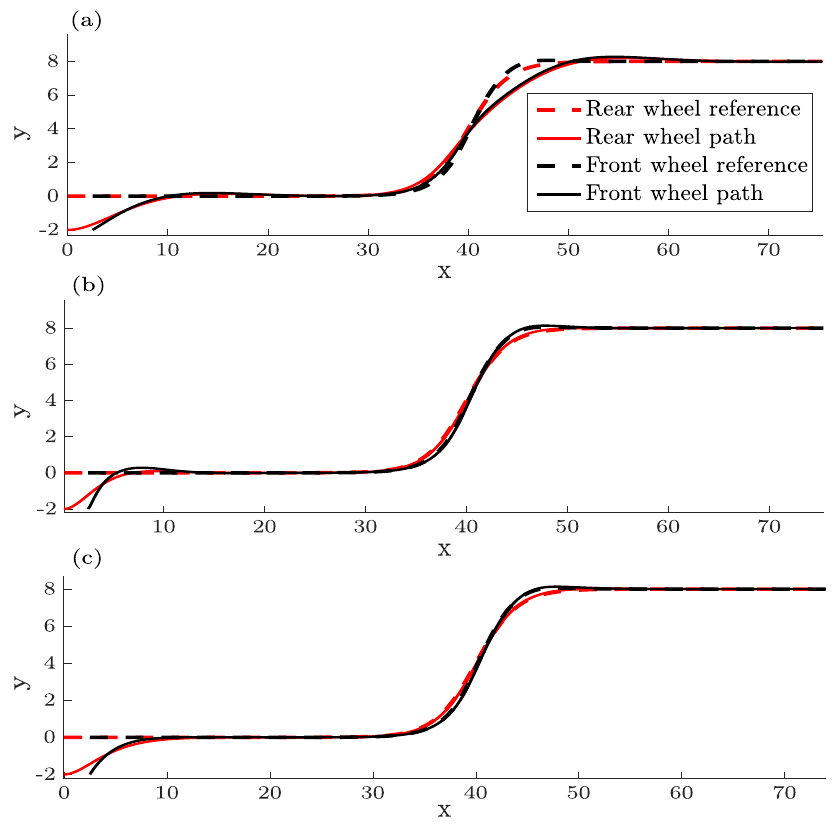}\caption{Tracking performance comparison for the three path stabilizing control
laws discussed in this section. 
(a) Pure pursuit deviates from reference when curvature is nonzero.
(b) The rear wheel output based controller drives the rear wheel to the rear wheel reference path. Overshoot is a result of the second order response of the system.\label{fig:comparison-of-pure}
(c) The front wheel output based controller drives the front wheel to the reference path with a first order response and tracks the path through the maneuver.}
\end{figure}

\paragraph*{Comparison of path tracking controllers for kinematic models}
The advantages of controllers based on the kinematic model with the no-slip constraint on the wheels is that they have low computational requirements, are readily implemented, and have good performance at moderate speeds. 
Figure \ref{fig:comparison-of-pure} provides a qualitative comparison of the path stabilizing controllers of this sections based on, and simulated with,  \eqref{eq:rear_wheel_dynamics} for a lane change maneuver. 
In the simulation of the front wheel output based controller, the rear wheel reference path is replaced by the front wheel reference path satisfying 
\begin{equation}
\begin{array}{c}
x_{ref}(s)\mapsto x_{ref}(s)+l\cos(\theta),\\
y_{ref}(s)\mapsto y_{ref}(s)+l\sin(\theta).
\end{array}
\end{equation}

The parameters of the simulation are summarized in Table \ref{tab:sim}.

In reference to Figure \ref{fig:comparison-of-pure}, the pure pursuit control tracks the reference path during periods with no curvature. 
In the region where the path has high curvature, the pure pursuit control causes the system to deviate from the reference path.
In contrast, the latter two controllers converge to the path and track it through the high curvature regions. 

In both controllers using \eqref{eq:near_point} in the feedback policy, local exponential stability can only be proven if there is a neighborhood of the path where \eqref{eq:near_point} is continuous. 
Intuitively, this means the path cannot cross over itself and must be differentiable.

\begin{table}
\centering{} \setlength{\tabcolsep}{2.25pt}
%\begin{tabular}{|>{\raggedright}p{3.6cm}|>{\centering}p{3cm}|>{\centering}p{2.25cm}|>{\centering}p{2.1cm}|>{\centering}p{3.3cm}|>{\centering}p{1.25cm}|}\rowcolor{dark_grey}
\begin{tabular}{@{}|c|c|c|c|@{}}
\hline 
\multicolumn{4}{|c|}{{ \cellcolor{dark_grey} Example Parameters}}\tabularnewline 
\hline \hline
Reference path & \begin{tabular}{@{}c@{}}Wheel-\\base\end{tabular} & \begin{tabular}{@{}c@{}}Steering\\ limit\end{tabular} & \begin{tabular}{@{}c@{}}Initial\\ Configuration\end{tabular}
\tabularnewline \hline %\rowcolor{white}
\begin{tabular}{@{}c@{}}
$\left(x_{ref}(s),y_{ref}(s)\right)=$\\
$(s,4\cdot\tanh\left(\frac{s-40}{4}\right)
$\vspace*{0.05cm}\end{tabular}
&
$L=5$
&
$|\delta|\leq\pi/4$ &  \begin{tabular}{@{}c@{}}$\left(x_{r}(0),y_{r}(0),\theta(0)\right)$\\$=\left(0,-2,0\right)$\end{tabular}
\tabularnewline
\hline 
\end{tabular}

\vskip0.075cm
\setlength{\tabcolsep}{4.5pt}
\begin{tabular}{@{}|c|c|c|@{}}
\hline %\rowcolor{dark_grey}
\multicolumn{3}{|c|}{{ \cellcolor{dark_grey} Controller Parameters}}\tabularnewline 
\hline \hline
Pure pursuit & Rear wheel feedback & \begin{tabular}{@{}c@{}}Front wheel\\ feedback\end{tabular}
\tabularnewline \hline %\cellcolor{white}
$L=5,\,v_r=1$ 
& 
$\begin{array}{c}
k_{e}=0.25,\,k_{\theta}=0.75,\\v_{r}=1
\end{array}$ & $k=0.5,\,v_r=1$
\tabularnewline
\hline 
\end{tabular}

%\scriptsize{
%\begin{tabular}{@{}|c|@{}}
%\hline
%\rowcolor{dark_grey}Example Parameters
%\tabularnewline
%\hline \hline
%\begin{tabular}{@{}c|c|c|c@{}}
% Reference path & \begin{tabular}{@{}c@{}}Wheel-\\base\end{tabular} & \begin{tabular}{@{}c@{}}Steering\\ limit\end{tabular} & \begin{tabular}{@{}c@{}}Initial\\ Configuration\end{tabular}
%\tabularnewline \hline \rowcolor{white}
%\begin{tabular}{@{}c@{}}
%$\left(x_{ref}(s),y_{ref}(s)\right)=$\\
%$(s,4\cdot\tanh\left(\frac{s-40}{4}\right)
%$\end{tabular}
%&
%$L=5$
%&
%$|\delta|\leq\pi/4$ &  \begin{tabular}{@{}c@{}}$\left(x_{r}(0),y_{r}(0),\theta(0)\right)$\\$=\left(0,-2,0\right)$\end{tabular}
%\end{tabular}
%\tabularnewline
%\hline\hline
%\rowcolor{dark_grey} Controller Parameters
%\tabularnewline 
%\hline\hline 
%\begin{tabular}{@{}c|c|c@{}} 
%Pure pursuit & Rear wheel feedback & \begin{tabular}{@{}c@{}}Front wheel\\ feedback\end{tabular}
%\tabularnewline \hline \rowcolor{white}
%$L=5,\,v_r=1$ 
%& 
%$\begin{array}{c}
%k_{e}=0.25,\,k_{\theta}=0.75,\\v_{r}=1
%\end{array}$ & $k=0.5,\,v_r=1$
%\end{tabular}
%\tabularnewline
%\hline
%\end{tabular}
%}
\vspace*{2mm}\caption{Simulation and controller parameters used to generate Figure \ref{fig:comparison-of-pure}. \label{tab:sim}}
\end{table}

\subsection{Trajectory Tracking Control for the Kinematic Model}

\subsubsection{Control Lyapunov based design\label{control_lyapunov}} 
A control design based on a control Lyapunov function is described in \cite{kanayama1990stable}. 
The approach is to define the tracking error in a coordinate frame fixed to the car.
The configuration error can be expressed by a change of basis from the inertial coordinate frame using the reference trajectory,
and velocity, $\left(x_{ref},y_{ref},\theta_{ref},v_{ref},\omega_{ref}\right)$,
\[
\begin{pmatrix}x_{e}\\
y_{e}\\
\theta_{e}
\end{pmatrix}=\begin{pmatrix}\cos(\theta) & \sin(\theta) & 0\\
-\sin(\theta) & \cos(\theta) & 0\\
0 & 0 & 1
\end{pmatrix}\begin{pmatrix}x_{ref}-x_{r}\\
y_{ref}-y_{r}\\
\theta_{ref}-\theta
\end{pmatrix}.
\]
The evolution of the configuration error is then 
\[
\begin{array}{@{\hspace{1pt}}r@{\hspace{1pt}}c@{\hspace{1pt}}l@{\hspace{1pt}}}
\dot{x}_{e} & = & \omega y_{e}-v_{r}+v_{ref}\cos(\theta_{e}),\\
\dot{y}_{e} & = & -\omega x_{e}+v_{ref}\sin(\theta_{e}),\\
\dot{\theta}_{e} & = & \omega_{ref}-\omega.
\end{array}
\]
With the control assignment, 
\[
\begin{array}{@{\hspace{1pt}}r@{\hspace{1pt}}c@{\hspace{1pt}}l@{\hspace{1pt}}}
v_{r} & = & v_{ref}\cos(\theta_{e})+k_{1}x_{e},\\
\omega & = & \omega_{ref}+v_{ref}\left(k_{2}y_{e}+k_{3}\sin(\theta_{e})\right).
\end{array}
\]
the closed loop error dynamics become 
\[
\begin{array}{@{\hspace{1pt}}r@{\hspace{1pt}}c@{\hspace{1pt}}l@{\hspace{1pt}}}
\dot{x}_{e} & = & (\omega_{ref}+v_{ref}\left(k_{2}y_{e}+k_{3}\sin(\theta_{e})\right))y_{e}-k_{1}x_{e},\\
\dot{y}_{e} & = & -\left(\omega_{ref}+v_{ref}\left(k_{2}y_{e}+k_{3}\sin(\theta_{e})\right)\right)x_{e}+v_{ref}\sin\left(\theta_{e}\right),\\
\dot{\theta}_{e} & = & \omega_{ref}-\omega.
\end{array}
\]
Stability is verified for $k_{1,2,3}>0$, $\dot{\omega}_{ref}=0$,
and $\dot{v}_{ref}=0$ by the Lyapunov function 
\[
V=\frac{1}{2}\left(x_{e}^{2}+y_{e}^{2}\right)+\frac{\left(1-\cos(\theta_{e})\right)}{k_{2}},
\]
with negative semi-definite time derivative, 
\[
\dot{V}=-k_{1}x_{e}^{2}-\frac{v_{ref}k_{3}\sin^{2}(\theta_{e})}{k_{2}}.
\]
A local analysis shows that the control law provides local exponential
stability. 
However, for the system to be time invariant, $\omega_{ref}$ and $v_{ref}$ are required to
be constant.

A related controller is proposed in \cite{jiangdagger1997tracking}
which utilizes a backstepping design to achieve uniform local exponential
stability for a finite domain with time varying references.

\subsubsection{Output feedback linearization\label{output_feedback_linearization}}

For higher vehicle speeds, it is appropriate to constrain the steering angle to have continuous motion as in \eqref{eq:Steering_rate_control}. 
With the added state, it becomes more difficult to design a controller from simple geometric considerations.
A good option in this case is to output-linearize the system. 

This is not easily accomplished using the front or rear wheel positions. 
An output which simplifies the feedback linearization is proposed in \cite{andrea1995control}, where a point ahead of the vehicle by any distance $d\neq0$, aligned with the steering angle is selected.

Let $x_{p}=x_{f}+d\cos(\theta+\delta)$ and $y_{p}=y_{f}+d\sin(\theta+\delta)$ be the output of the system. Taking the derivative of these outputs and substituting the dynamics of \ref{eq:Steering_rate_control} yields 

\begin{equation}
\begin{array}{@{\hspace{1pt}}l@{\hspace{1pt}}}
\left(\begin{array}{@{\hspace{1pt}}c@{\hspace{1pt}}}
\dot{x}_{p}\\
\dot{y}_{p}
\end{array}\right)=\\
\ \underset{A(\theta,\delta)}{\underbrace{\left(\begin{array}{@{\hspace{1pt}}c@{\hspace{3pt}}c@{\hspace{1pt}}}
\cos(\theta+\delta)-\frac{d}{l}\sin(\theta+\delta)\sin(\delta) & -d\sin(\theta+\delta)\\
\sin(\theta+\delta)+\frac{d}{l}\cos(\theta+\delta)\sin(\delta) & d\cos(\theta+\delta)
\end{array}\right)}}\left(\begin{array}{@{\hspace{1pt}}c@{\hspace{1pt}}}
v_{f}\\
v_{\delta}
\end{array}\right).
\end{array}\label{eq:fb_linearization}
\end{equation}

Then, defining the right
hand side of  \eqref{eq:fb_linearization} as auxiliary control variables  $u_{x}$ and $u_{y}$ yields 
\begin{equation}
\left(\begin{array}{@{\hspace{1pt}}l@{\hspace{1pt}}}
\dot{x}_{p}\\
\dot{y}_{p}
\end{array}\right)=\left(\begin{array}{@{\hspace{1pt}}c@{\hspace{1pt}}}
u_{x}\\
u_{y}
\end{array}\right),
\end{equation}
which makes control straightforward. From $u_{x}$ and $u_{y}$,
the original controls $v_{f}$ and $v_{\delta}$ are recovered by
using the inverse of the matrix in \eqref{eq:fb_linearization}, provided
below: 

%\begin{flushleft}
%\[
\begin{align}
\nonumber &[A(\theta,\delta)]^{-1}=\\
%\]
%\par\end{flushleft}
%{\smaller{}
&\footnotesize{\left(\hspace{-0.1cm}\begin{array}{@{\hspace{1pt}}c@{\hspace{5pt}}c@{\hspace{1pt}}}
 \cos(\theta+\delta) & \sin(\theta+\delta)\\
 -\frac{1}{d}\sin(\theta\hspace{-0.05cm}+\hspace{-0.05cm}\delta)\hspace{-0.05cm}-\hspace{-0.05cm}\frac{1}{l}\cos(\theta\hspace{-0.05cm}+\hspace{-0.05cm}\delta)\sin(\delta) &  \frac{1}{d}\cos(\theta\hspace{-0.05cm}+\hspace{-0.05cm}\delta)\hspace{-0.05cm}-\hspace{-0.05cm}\frac{1}{l}\sin(\theta\hspace{-0.05cm}+\hspace{-0.05cm}\delta)\sin(\delta)
\end{array}\hspace{-0.1cm}\right).}
%}
\end{align}

%{\tiny{}
%\[
%\left(\begin{array}{cc}
%\frac{l\cos(\delta+\theta)}{l\cos(2(\delta+\theta))-d\sin(\delta)\sin(2(\delta+\theta))} & \frac{l\sin(\delta+\theta)}{d\sin(\delta)\sin(2(\delta+\theta))-l\cos(2(\delta+\theta))}\\
%\frac{d\cos(\delta+\theta)\sin(\delta)+l\sin(\delta+\theta)}{d(d\sin(\delta)\sin(2(\delta+\theta))-l\cos(2(\delta+\theta)))} & \frac{d\sin(\delta)\sin(\delta+\theta)-l\cos(\delta+\theta)}{d(d\sin(\delta)\sin(2(\delta+\theta))-l\cos(2(\delta+\theta)))}
%\end{array}\right).
%\]
%}
From the input-output linear system, local trajectory stabilization
can be accomplished with the controls 
\begin{equation}
\begin{array}{c}
u_{x}=\dot{x}_{p,ref}+k_{x}(x_{p,ref}-x_{p}),\\
u_{y}=\dot{y}_{p,ref}+k_{y}(y_{p,ref}-y_{p}).
\end{array}
\end{equation}

To avoid confusion, note that in this case, the output position $(x_{p},y_{p})$
and controlled speed $v_{f}$ are not collocated as in the previously
discussed controllers.

%\paragraph{Unconstrained predictive control approaches}

%%%% MPC %%%%%
\subsection{Predictive Control Approaches} \label{sec:predictive}

The simple control laws discussed above are suitable for moderate driving conditions. 
\begin{full}
However, slippery roads or emergency maneuvers may require a more accurate model, such as the one introduced in Section \ref{inertia}.
\end{full}
\begin{short}
However, a higher fidelity model may be required to plan and execute aggressive or emergency maneuvers.
\end{short} 
The added detail of more sophisticated models complicates the control design making it difficult to construct controllers from intuition and geometry of the configuration space. 

Model predictive control~\cite{garcia1989model} is a general control design methodology which can be very effective for this problem.
Conceptually, the approach is to solve the motion planning problem over a short time horizon, take a short interval of the resulting open loop control, and apply it to the system. 
While executing, the motion planning problem is re-solved to find an appropriate control for the next time interval.
Advances in computing hardware as well as mathematical programming algorithms have made predictive control feasible for real-time use in driverless vehicles.
MPC is a major field of research on its own and this section is only intended to provide a brief description of the technique and to survey results on its application to driverless vehicle control.

Since model predictive control is a very general control technique, the model takes the form of a general continuous time control system with control, $u(t)\in\mathbb{R}^{m}$, and state, $x(t)\in\mathbb{R}^{n}$,
\begin{equation}
\dot{x}=f(x,u,t)\label{eq:system_dynamics}
\end{equation}
A feasible reference trajectory $x_{ref}(t)$, and for some motion planners $u_{ref}(t)$, are provided satisfying \eqref{eq:system_dynamics}. 
The system is then discretized by an appropriate choice of numerical approximation so that \eqref{eq:system_dynamics} is given at discrete time instances by
\begin{equation}
x_{k+1}\mathbf{=}F_{k}\left(x_{k},u_{k}\right),\quad k\in\mathbb{N},
\end{equation}
\begin{full}
One of the simplest discretization schemes is Euler's method with a zero order hold on control: 
\begin{equation}
\begin{array}{l}
F_{k}\left(x(k\cdot\Delta t),u(k\cdot\Delta t)\right)=\\
x(k\cdot\Delta t)+\Delta t\cdot f\left(x(k\cdot\Delta t),u(k\cdot\Delta t),t_{k}\right)
\end{array},\quad k\in\mathbb{N}.\label{eq:euler_approx}
\end{equation}
The state and control are discretized by their approximation at times $t_{k}=k\cdot\Delta t$. 
Solutions to the discretized system are approximate and will not match the continuous time equation exactly. 
Similarly, the reference trajectory and control sampled at the discrete times $t_{k}$ will not satisfy the discrete time equation. 
For example, the mismatch between solutions to \eqref{eq:euler_approx} and \eqref{eq:system_dynamics}
will be $O(\Delta t)$ and the reference trajectory sampled at time
$t_{k}$ will result in
\begin{equation}
F_{k}\left(x_{ref}(t_{k}),u_{ref}(t_{k})\right)-x_{ref}(t_{k+1})=O(\Delta t^{2}).
\end{equation}
\end{full}
To avoid over-complicating the following discussion, we assume the discretization is exact for the remainder of the section. 
The control law typically takes the form 
\[
u_{k}(x_{meas.})=\hspace{-0.1cm}\underset{\begin{array}{@{\hspace{1pt}}c@{\hspace{1pt}}}
x_{n}\in\mathcal{X}_{n},\\u_{n}\in\mathcal{U}_{n}%,\\
%\Delta u_{n}\in\Delta\mathcal{U}_{n}
\end{array}}{{\rm argmin}}\hspace{-0.1cm}\left\{ h(x_{N}-x_{ref,N},u_{N}-u_{ref,N})\begin{array}{c}
\\
\\
\end{array}\right.
\]
\begin{equation}
\left.+\sum_{n=k}^{k+N-1}g_{n}(x_{n}-x_{ref,n},u_{n}-u_{ref,n})\right\} \label{eq:MPC}
\end{equation}
\[
\begin{array}{l}
{\rm subject\, to}\\
x_{k}=x_{meas.}\\
x_{n+1}=F(x_{n},u_{n}),\, \\n\in\{k,...,k+N-1\}.
\end{array}
\]
The function $g_{n}$ penalizes deviations from the reference trajectory and control at each time step, while the function $h$ is a terminal penalty at the end of the time horizon. 
The set $\mathcal{X}_{n}$ is the set of allowable states which can restrict undesirable positions or velocities, e.g., excessive tire slip or obstacles.  
The set $\mathcal{U}_{n}$ encodes limits on the magnitude of the input signals.
Important considerations are whether the solutions to the right hand side of \eqref{eq:MPC} exist, and when they do, the stability and robustness of the closed loop system.
These issues are investigated in the predictive control literature \cite{camacho2013model, mayne2000constrained}.
% as well as the complexity of computing the solution. 

To implement an MPC on a driverless car,  \eqref{eq:MPC} must be solved several times per second which is a major obstacle to its use.
In the special case that $h$ and $g_{n}$ are quadratic, $\mathcal{U}_{n}$ and $\mathcal{X}_{n}$ are polyhedral, and $F$ is linear, the problem becomes a quadratic program.
Unlike a general nonlinear programming formulation, interior point algorithms are available for solving quadratic programs in polynomial time. To leverage this, the complex vehicle model is often linearized to obtain an approximate linear model.
 Linearization approaches typically differ in the reference about which the linearization is computed---current operating point \cite{falcone2007predictive,falcone2007linear,falcone2008linear}, reference path \cite{kim2014model} or more generally, about a reference trajectory, which results in the following approximate linear model:

%{[}Linearize about operating point, about reference path and about
%reference trajectory{]}

\begingroup
\renewcommand*{\arraystretch}{1.5}
\begin{equation}
\begin{array}{l}
x_{ref,k+1}+\xi_{k+1} =F_{k}(x_{ref,k},u_{ref,k})\\ \qquad+\underset{A_{k}}{\underbrace{\nabla_{x}F_{k}(x_{ref,k},u_{ref,k})}}\xi_{k}
+\underset{B_{k}}{\underbrace{\nabla_{u}F_{k}(x_{ref,k},u_{ref,k})}}\eta_{k}\\ \qquad +O(\Vert\xi_k\Vert^{2})+O(\Vert\eta_k\Vert^{2}),
\end{array}
\end{equation}
\endgroup
where $\xi:=x-x_{ref}$ and $\eta:=u-u_{ref}$ are the deviations
of the state and control from the reference trajectory. This first
order expansion of the perturbation dynamics yields a linear time
varying (LTV) system, 
\begin{equation}
\xi_{k+1}=A_{k}\xi_{k}+B_{k}\eta_{k}.\label{eq:linearization}
\end{equation}
Then using a quadratic objective, and expressing the polyhedral constraints
algebraically, we obtain
\[
u_{k}(x_{meas.})\hspace{-0.05cm}=\hspace{-0.05cm}\underset{\xi_{k},\eta_{k}}{{\rm argmin}}\left\{ \xi_{N}^{T}H\xi_{N}\hspace{-0.05cm}+\hspace{-0.1cm}\sum_{n=k}^{k+N-1}\xi_{k}^{T}Q_{k}\xi_{k}\hspace{-0.05cm}+\hspace{-0.05cm}\eta_{k}^{T}R_{k}\eta_{k}\right\} 
\] \vspace{-0.3cm}
\begin{equation}\label{eq:linear_mpc}
\begin{array}{l}
{\rm subject\, to}\\
\xi_{k}=x_{meas.}-x_{ref}\\
C_{n}\xi_{n}\leq0,\quad D_{n}\eta_{n}\leq0\\%,\quad D_{n}^{\Delta}\Delta\eta_{n}\leq0,\\
\xi_{k+1}=A_{k}\xi_{k}+B_{n}\eta_{n},\, n\in\{k,...,k+N-1\},
\end{array}
\end{equation}
where $R_{k}$ and $Q_{k}$ are positive semi-definite. 
If the states and inputs are unconstrained, i.e., $\mathcal{U}_{n}=\mathbb{R}^{m}$, $\mathcal{X}_{n}=\mathbb{R}^{n}$, a semi-closed form solution can be obtained by dynamic programming requiring only the calculation of an $N$ step matrix recursion~\cite{kirk2012optimal}. 
Similar closed form recursive solutions have also been explored when vehicle models are represented by controlled auto-regressive integrated moving average (CARIMA) with no state and input constraints \cite{ollero1991predictive,raffo2009predictive}.

A further variation in the model predictive control approach is to
replace state constraints (e.g., obstacles) and input constraints
by penalty functions in the performance functional. %[Penalty functions (potential field based) instead of as state constraints] 
Such a predictive control approach based on a dynamic model with nonlinear
tire behavior is presented in \cite{yoon2009model}. In addition to
penalizing control effort and deviation from a reference path to a
goal which does not consider obstacles, the performance functional
penalizes input constraint violations and collisions with obstacles
over the finite control horizon. In this sense it is similar to potential
field based motion planning, but is demonstrated to have improved
performance.

The following are some variations of the model predictive control
framework that are found in the literature of car controllers:

\subsubsection{Unconstrained MPC with Kinematic Models}

The earliest predictive controller in \cite{ollero1991predictive}
falls under this category, in which the model predictive control framework
is applied without input or state constraints using a CARIMA model.
The resulting semi-closed form solution has minimal computational requirements and has also been adopted in \cite{raffo2009predictive}.
Moreover, the time-varying linear quadratic programming approach with
no input or state constraints was considered in \cite{raffo2009predictive}
using a linearized kinematic model.
% Thus, the CARIMA model approach was favored over this approach in \cite{raffo2009predictive}. %In comparison, the work in \cite{raffo2009predictive} focuses on a simplified model with a linear tire behavior and without input and state constraints in order to implement a predictive controller using modest computation hardware. [closed-form solutions]

\subsubsection{Path Tracking Controllers}

In \cite{kim2014model}, a predictive control is investigated using a center of mass based linear dynamic model (assuming constant velocity) for path tracking and an approximate steering model. 
The resulting integrated model is validated with a detailed automatic steering model and a 27 degree-of-freedom CarSim vehicle model.
\begin{full}
\subsubsection{Trajectory Tracking Controllers}
A predictive controller using a tire model similar to Section \ref{section_models} was investigated in \cite{falcone2007predictive}.
\end{full}
\begin{short}
\subsubsection{Trajectory Tracking Controllers}
A predictive controller using a dynamic model with nonlinear tire behavior was investigated in \cite{falcone2007predictive}.
\end{short}
The full nonlinear predictive control strategy was carried out in simulation and shown to stabilize a simulated emergency maneuver in icy conditions with a control frequency of 20 Hz.
However, with a control horizon of just two time steps the computation time was three times the sample time of the controller making experimental validation impossible. 
A linearization based approach was also investigated in \cite{falcone2007predictive,falcone2007linear,falcone2008linear} based on a single linearization about the state of the vehicle at the current time step. 
The reduced complexity of solving the quadratic program resulted in acceptable computation time, and successful experimental results are reported for driving in icy conditions at speeds up to 21m/s. 
The promising simulation and experimental results of this approach are improved upon in \cite{falcone2008linear} by providing conditions for the uniform local-asymptotic stability of the time varying system.

\subsection{Linear Parameter Varying Controllers} \label{sec:lpv}

Many controller design techniques are available for linear systems making a linear model desirable. 
However, the broad range of operating points encountered under normal driving conditions make it difficult to rely on a model linearized about a single operating point. 
To illustrate this, consider the lateral error dynamics in  \eqref{eq:curvilinear_coord}. 
If the tracking error is assumed to remain small a linearization of the dynamics around the operating point $\theta_{e}=0$ and $e=0$ yields 
\begin{equation}
\begin{array}{rcl}
\left(\begin{array}{c}
\dot{e}\\
\dot{\theta}_{e}
\end{array}\right) & = & \left(\begin{array}{cc}
0 & v_{r}\\
0 & 0
\end{array}\right)\left(\begin{array}{c}
e\\
\theta_{e}
\end{array}\right)\\
 & + & \left(\begin{array}{c}
0\\
1
\end{array}\right)\omega+\left(\begin{array}{c}
0\\
-v_{r}\kappa(s)
\end{array}\right).
\end{array}
\end{equation}
Introducing a new control variable incorporating a feed-forward $u=\omega+v_{r}\kappa(s)$
simplifies the discussion. 
The dynamics are now 
\begin{equation}
\left(\begin{array}{c}
\dot{e}\\
\dot{\theta}_{e}
\end{array}\right)=\left(\begin{array}{cc}
0 & v_{r}\\
0 & 0
\end{array}\right)\left(\begin{array}{c}
e\\
\theta_{e}
\end{array}\right)+\left(\begin{array}{c}
0\\
1
\end{array}\right)u.
\end{equation}
Observe that the model is indeed linear, but the forward speed $v_{r}$ appears in the linear model. 
A simple proportional plus derivative control with gains $k_{p}$ and $k_{d}$ will stabilize the lateral dynamics but the poles of the closed loop system are given by $\left(-k_{d}\pm\sqrt{k_{d}^{2}-4k_{p}v_{r}}\right)/2.$
At higher speeds the poles move into the complex plane leading to an oscillatory response. 
In contrast, a small $k_{p}$ gain leads to a poor response at low speed. 
A very intuitive and widely used remedy to this challenge is gain scheduling. 
In this example, parameterizing $k_{p}$ as a function of $v_{r}$ fixes the poles to a single value for each speed. 
This technique falls into the category of control design for linear parameter varying (LPV) models \cite{sename2013robust}.
Gain scheduling is a classical approach to this type of controller design. 
Tools from robust control and convex optimization are readily applied to address more complex models. 
LPV control designs for lateral control are presented in \cite{gaspar2012lpv, huang2005ltv, baslamisli2009gain}. 
LPV models are used in \cite{besselmann2009autonomous, besselmann2007hybrid} together with predictive control approaches for path and trajectory stabilization. 
At a lower level of automation, LPV control techniques have been proposed for integrated system control. 
In these designs, several subsystems are combined under a single controller to achieve improved handling performance. 
LPV control strategies for actuating active and semi-active suspension systems are developed in \cite{gaspar2005design, poussot2008new}, while integrated suspension and braking control systems are developed in \cite{doumiati2013integrated, gaspar2007toward}. 

\section{Conclusions} \label{section_conclusion}
The past three decades have seen increasingly rapid progress in driverless vehicle technology. 
In addition to the advances in computing and perception hardware, this rapid progress has been enabled by major theoretical progress in the computational aspects of mobile robot motion planning and feedback control theory. 
Research efforts have undoubtedly been spurred by the improved utilization and safety of road networks that driverless vehicles would provide.  
Driverless vehicles are complex systems which have been decomposed into a hierarchy of decision making problems, where the solution of one problem is the input to the next.
The breakdown into individual decision making problems has enabled the use of well developed methods and technologies from a variety of research areas. 
The task is then to integrate these methods so that their interactions are semantically valid, and the combined system is computationally efficient.  
A more efficient motion planning algorithm may only be compatible with a computationally intensive feedback controller such as model predictive control. 
Conversely, a simple control law may require less computation to execute, but is also less robust and requires using a more detailed model for motion planning. 

This paper has provided a survey of the various aspects of driverless vehicle decision making problems with a focus on motion planning and feedback control. 
The survey of performance and computational requirements of various motion planning and control techniques serves as a reference for assessing compatibility and computational tradeoffs between various choices for system level design.

\begin{full}
\balance
\end{full}

%\Urlmuskip=0mu plus 50mu
\bibliographystyle{ieeetr}
\bibliography{references} 

%\vspace*{0.05cm}
\begin{long}
%\linebreak
\begin{minipage}{0.475\textwidth} \footnotesize \vspace*{0.05cm}
\begin{wrapfigure}{l}{0.275\textwidth}
\vspace{-0.425cm}
\includegraphics[width=1in,height=1.25in,clip,keepaspectratio]{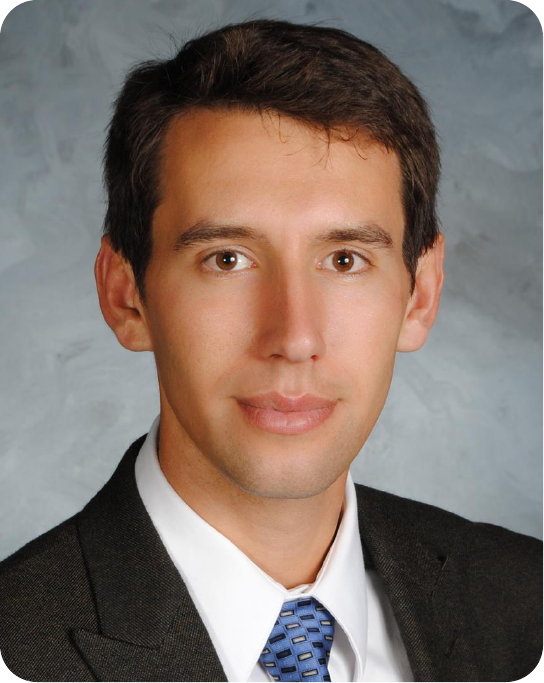}\vspace{-0.285cm}
\end{wrapfigure} \textbf{Brian Paden} received B.S. and M.S. degrees in Mechanical Engineering in 2011 and  2013 respectively from UC Santa Barbara. Concurrently from 2011 to 2013 he was an Engineer at LaunchPoint Technologies developing electronically controlled automotive valve-train systems. His research interests are in the areas of control theory and motion planning with a focus on applications for autonomous vehicles. His current affiliation is with the Laboratory for Information and Decision systems and he is a Doctoral Candidate in the department of Mechanical Engineering at MIT. \vspace*{0.3cm}
\end{minipage}
%\end{IEEEbiography}
\begin{minipage}{0.475\textwidth} \footnotesize
\begin{wrapfigure}{l}{0.275\textwidth}
\vspace{-0.425cm}
\includegraphics[width=1in,height=1.25in,clip,keepaspectratio]{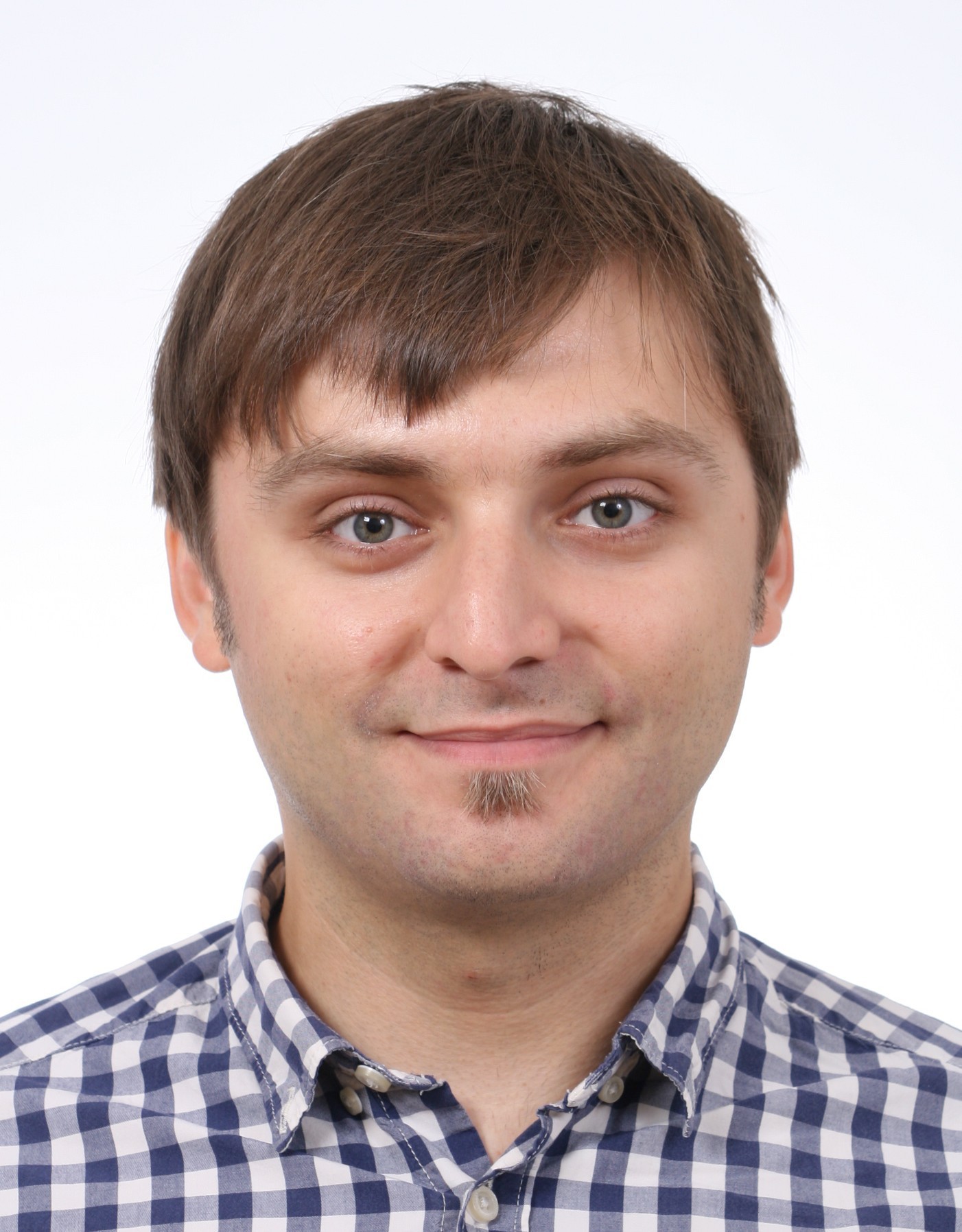}\vspace{-0.285cm}
\end{wrapfigure} \textbf{Michal Čáp} received his Bc. degree in Information Technology from Brno University of Technology, the Czech Republic and MSc degree in Agent Technologies from Utrecht University, the Netherlands. He is currently pursuing PhD degree in Artificial Intelligence at CTU in Prague, Czech Republic. At the moment, Michal Cap holds the position of Fulbright-funded visiting researcher at Massachusetts Institute of Technology, USA. His research interests include motion planning, multi-robot trajectory coordination and autonomous transportation systems in general. \vspace*{0.3cm}
\end{minipage}
\begin{minipage}{0.475\textwidth} \footnotesize
\begin{wrapfigure}{l}{0.275\textwidth}
\vspace{-0.425cm}
\includegraphics[width=1in,height=1.25in,clip,keepaspectratio]{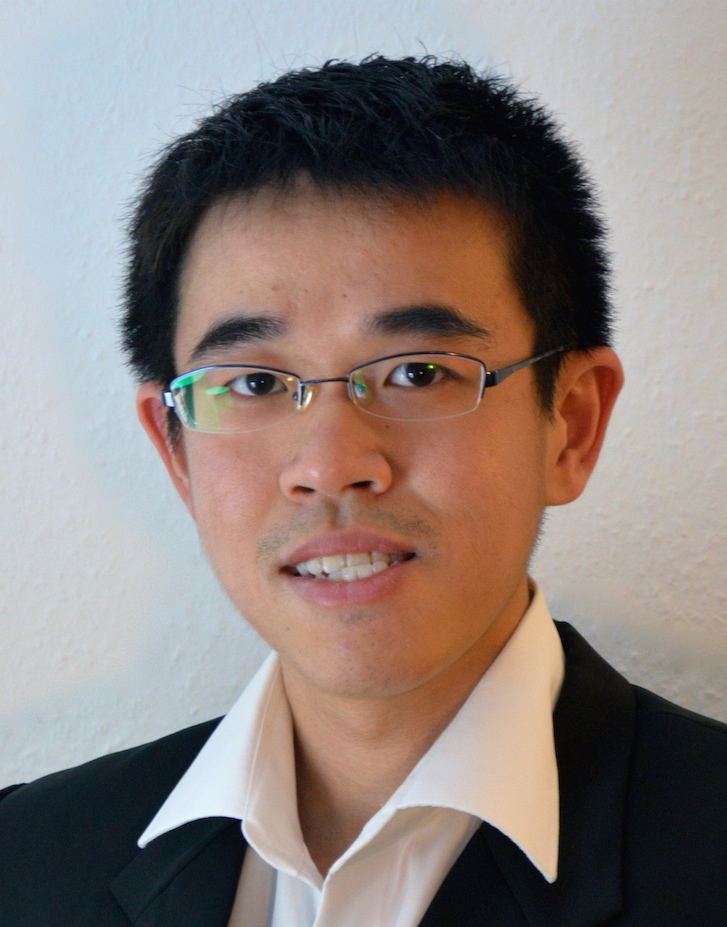}\vspace{-0.285cm}
\end{wrapfigure} \textbf{Sze Zheng Yong}
is currently a postdoctoral associate in the Laboratory for Information and Decision Systems  at Massachusetts Institute of Technology. He obtained his Dipl.-Ing.(FH) degree in automotive engineering with a specialization in mechatronics and control systems  from the Esslingen University of Applied Sciences, Germany in 2008 and his S.M. and Ph.D. degrees in mechanical engineering from MIT in 2010 and 2016, respectively. His research interests lie in the broad area of control and estimation of hidden mode hybrid systems, with applications to intention-aware autonomous systems and resilient cyber-physical systems. \vspace*{0.5cm}
\end{minipage}
\begin{minipage}{0.475\textwidth} \footnotesize
\begin{wrapfigure}{l}{0.275\textwidth}
\vspace{-0.425cm}
\includegraphics[width=1in,height=1.25in,clip,keepaspectratio]{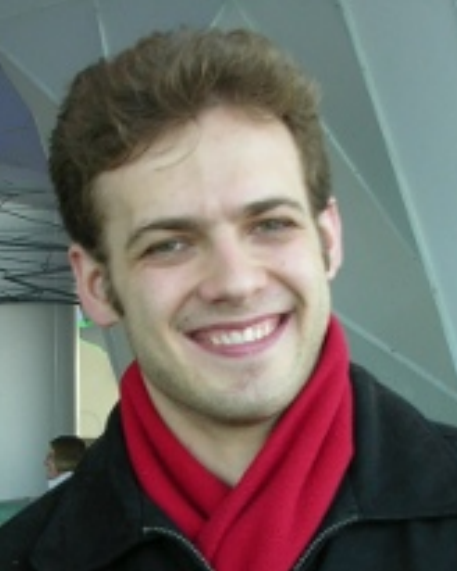}\vspace{-0.285cm}
\end{wrapfigure} \textbf{Dmitry Yershov} was born in Kharkov, Ukraine, in 1983. He received his
B.S. and M.S. degree in applied mathematics from Kharkov National
University in 2004 and 2005, respectively. In 2013, he graduated with
a Ph.D. degree from the the Department of Computer Science at the
University of Illinois at Urbana-Champaign. He joined the Laboratory
for Information and Decision Systems at the Massachusetts Institute of
Technology in years 2013--2015 where he was a postdoctoral
associate. Currently, he is with nuTonomy inc. His research is focused
primarily on the feedback planning problem in robotics, which extends
to planning in information spaces and planning under uncertainties. \vspace*{0.5cm}
\end{minipage}
\begin{minipage}{0.475\textwidth} \footnotesize
\begin{wrapfigure}{l}{0.275\textwidth}
\vspace{-0.425cm}
\includegraphics[width=1in,height=1.25in,clip,keepaspectratio]{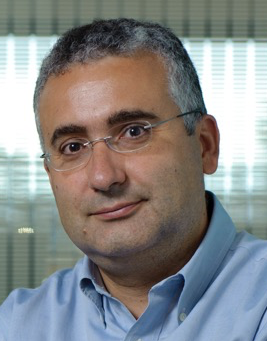}\vspace{-0.285cm}
\end{wrapfigure} \textbf{Emilio Frazzoli}
is a Professor of Aeronautics and Astronautics with the Laboratory for Information and Decision Systems at the Massachusetts Institute of Technology. He received a Laurea degree in Aerospace Engineering from the University of Rome, ``Sapienza" , Italy, in 1994, and a Ph. D. degree from the Department of Aeronautics and Astronautics of the Massachusetts Institute of Technology, in 2001. Before returning to MIT in 2006, he held faculty positions at the University of Illinois, Urbana-Champaign, and at the University of California, Los Angeles. He was the recipient of a NSF CAREER award in 2002, and of the IEEE George S. Axelby award in 2015. His current research interests focus primarily on autonomous vehicles, mobile robotics, and transportation systems.
\end{minipage}
\end{long}

\begin{short}
\begin{IEEEbiography}[{\includegraphics[width=1in,height=1.25in,clip,keepaspectratio]{Figures/bio_photos/brian_bio}}]{Brian Paden} received B.S. and M.S. degrees in Mechanical Engineering in 2011 and  2013 respectively from UC Santa Barbara. Concurrently from 2011 to 2013 he was an Engineer at LaunchPoint Technologies developing electronically controlled automotive valve-train systems. His research interests are in the areas of control theory and motion planning with a focus on applications for autonomous vehicles. His current affiliation is with the Laboratory for Information and Decision systems and he is a Doctoral Candidate in the department of Mechanical Engineering at MIT.
\end{IEEEbiography}

\begin{IEEEbiography}[{\includegraphics[width=1in,height=1.25in,clip,keepaspectratio]{Figures/bio_photos/cap.jpg}}]{Michal Čáp} received his Bc. degree in Information Technology from Brno University of Technology, the Czech Republic and MSc degree in Agent Technologies from Utrecht University, the Netherlands. He is currently pursuing PhD degree in Artificial Intelligence at CTU in Prague, Czech Republic. At the moment, Michal Cap holds the position of Fulbright-funded visiting researcher at Massachusetts Institute of Technology, USA. His research interests include motion planning, multi-robot trajectory coordination and autonomous transportation systems in general. 
\end{IEEEbiography}

\begin{IEEEbiography}[{\includegraphics[width=1in,height=1.25in,clip,keepaspectratio]{Figures/bio_photos/Yong.jpg}}]{Sze Zheng Yong}
is currently a postdoctoral associate in the Laboratory for Information and Decision Systems  at Massachusetts Institute of Technology. He obtained his Dipl.-Ing.(FH) degree in automotive engineering with a specialization in mechatronics and control systems  from the Esslingen University of Applied Sciences, Germany in 2008 and his S.M. and Ph.D. degrees in mechanical engineering from MIT in 2010 and 2016, respectively. His research interests lie in the broad area of control and estimation of hidden mode hybrid systems, with applications to intention-aware autonomous systems and resilient cyber-physical systems.
\end{IEEEbiography}

\begin{IEEEbiography}[{\includegraphics[width=1in,height=1.25in,clip,keepaspectratio]{Figures/bio_photos/dmitry_bio}}]{Dmitry Yershov}
Dmitry Yershov was born in Kharkov, Ukraine, in 1983. He received his
B.S. and M.S. degree in applied mathematics from Kharkov National
University in 2004 and 2005, respectively. In 2013, he graduated with
a Ph.D. degree from the the Department of Computer Science at the
University of Illinois at Urbana-Champaign. He joined the Laboratory
for Information and Decision Systems at the Massachusetts Institute of
Technology in years 2013--2015 where he was a postdoctoral
associate. Currently, he is with nuTonomy inc. His research is focused
primarily on the feedback planning problem in robotics, which extends
to planning in information spaces and planning under uncertainties.
\end{IEEEbiography}

\begin{IEEEbiography}[{\includegraphics[width=1in,height=1.25in,clip,keepaspectratio]{Figures/bio_photos/emilio-ilp}}]{Emilio Frazzoli}
is a Professor of Aeronautics and Astronautics with the Laboratory for Information and Decision Systems at the Massachusetts Institute of Technology. He received a Laurea degree in Aerospace Engineering from the University of Rome, ``Sapienza" , Italy, in 1994, and a Ph. D. degree from the Department of Aeronautics and Astronautics of the Massachusetts Institute of Technology, in 2001. Before returning to MIT in 2006, he held faculty positions at the University of Illinois, Urbana-Champaign, and at the University of California, Los Angeles. He was the recipient of a NSF CAREER award in 2002, and of the IEEE George S. Axelby award in 2015. His current research interests focus primarily on autonomous vehicles, mobile robotics, and transportation systems.
\end{IEEEbiography}
\vspace{8.0cm}
\end{short}
\end{document}